%% file: arxiv.tex
\documentclass[]{youtu} 

\usepackage{mathpazo}
\usepackage{graphicx}
\usepackage[numbers]{natbib}
\usepackage{float}
\usepackage{enumitem}
\usepackage{graphicx}
\usepackage[most]{tcolorbox}
\usepackage{subcaption}

\usepackage{tcolorbox}
\usepackage{enumitem}
\usepackage[dvipsnames]{xcolor}
\usepackage{fontawesome}
\usepackage{multirow}

\definecolor{myborder}{RGB}{73, 86, 102}
\definecolor{myRed}{RGB}{240, 48, 159}
\definecolor{mylightblue}{RGB}{235, 245, 255}

\tcbuselibrary{skins}

\title{Beyond Closed-Pool Video Retrieval:\\A Benchmark and Agent Framework for Real-World Video Search and Moment Localization}
\author{
Tao Yu\textsuperscript{$1,2 * \spadesuit$}, 
Yujia Yang\textsuperscript{$2 *$}, 
Haopeng Jin\textsuperscript{$3 *$}, 
Junhao Gong\textsuperscript{$4$}, 
Xinlong Chen\textsuperscript{$1,2$}, 
Yuxuan Zhou\textsuperscript{$5$}, 
Shanbin Zhang\textsuperscript{$1,2$}, 
Jiabing Yang\textsuperscript{$1,2$}, 
Xinming Wang\textsuperscript{$1,2$}, 
Hongzhu Yi\textsuperscript{$2 \ddagger$}, 
Ping Nie\textsuperscript{$4$},
Kai Zou\textsuperscript{$6$}, 
Zhang Zhang\textsuperscript{$1,2$}, 
Yan Huang\textsuperscript{$1,2 \ddagger$}, 
Liang Wang\textsuperscript{$1,2$}, 
Yeshani\textsuperscript{$3$}, 
Ruiwen Tao\textsuperscript{$3$}, 
Jin Ma\textsuperscript{$3$}, \\
Haijin Liang\textsuperscript{$3$}, 
Jinwen Luo\textsuperscript{$3$}
}

\vspace{-5mm}

\affiliation{
\textsuperscript{$1$}CASIA \ 
\textsuperscript{$2$}UCAS \ 
\textsuperscript{$3$}Tencent \
\textsuperscript{$4$}Peking University \
\textsuperscript{$5$}Tsinghua University \
\textsuperscript{$6$}Netmind.AI
}

\date{February 10, 2026}
\sourcecode{https://github.com/yutao1024/RACLO}
\data{https://huggingface.co/datasets/tencent/RVMS-Bench}

\begin{document}

\abstract{Traditional video retrieval benchmarks focus on matching precise descriptions to closed video pools, failing to reflect real-world searches characterized by fuzzy, multi-dimensional memories on the open web. We present \textbf{RVMS-Bench}, a comprehensive system for evaluating real-world video memory search. It consists of \textbf{1,440 samples} spanning \textbf{20 diverse categories} and \textbf{four duration groups}, sourced from \textbf{real-world open-web videos}. RVMS-Bench utilizes a hierarchical description framework encompassing \textbf{Global Impression, Key Moment, Temporal Context, and Auditory Memory} to mimic realistic multi-dimensional search cues, with all samples strictly verified via a human-in-the-loop protocol. We further propose \textbf{RACLO}, an agentic framework that employs abductive reasoning to simulate the human ``Recall-Search-Verify'' cognitive process, effectively addressing the challenge of searching for videos via fuzzy memories in the real world. Experiments reveal that existing MLLMs still demonstrate insufficient capabilities in real-world Video Retrieval and Moment Localization based on fuzzy memories. We believe this work will facilitate the advancement of video retrieval robustness in real-world unstructured scenarios.}
\maketitle

\renewcommand{\thefootnote}{*}
\footnotetext{Equal contribution.}

\renewcommand{\thefootnote}{\textdaggerdbl}
\footnotetext{Corresponding author.}

\renewcommand{\thefootnote}{\ensuremath{\spadesuit}}
\footnotetext{Project leader.}
\renewcommand{\thefootnote}{\arabic{footnote}}

\vspace{-.1em}

\section{Introduction}
\label{sec:introduction}

\begin{figure*}[t]
  \vskip 0.2in
  \centering
  \includegraphics[width=0.93\linewidth]{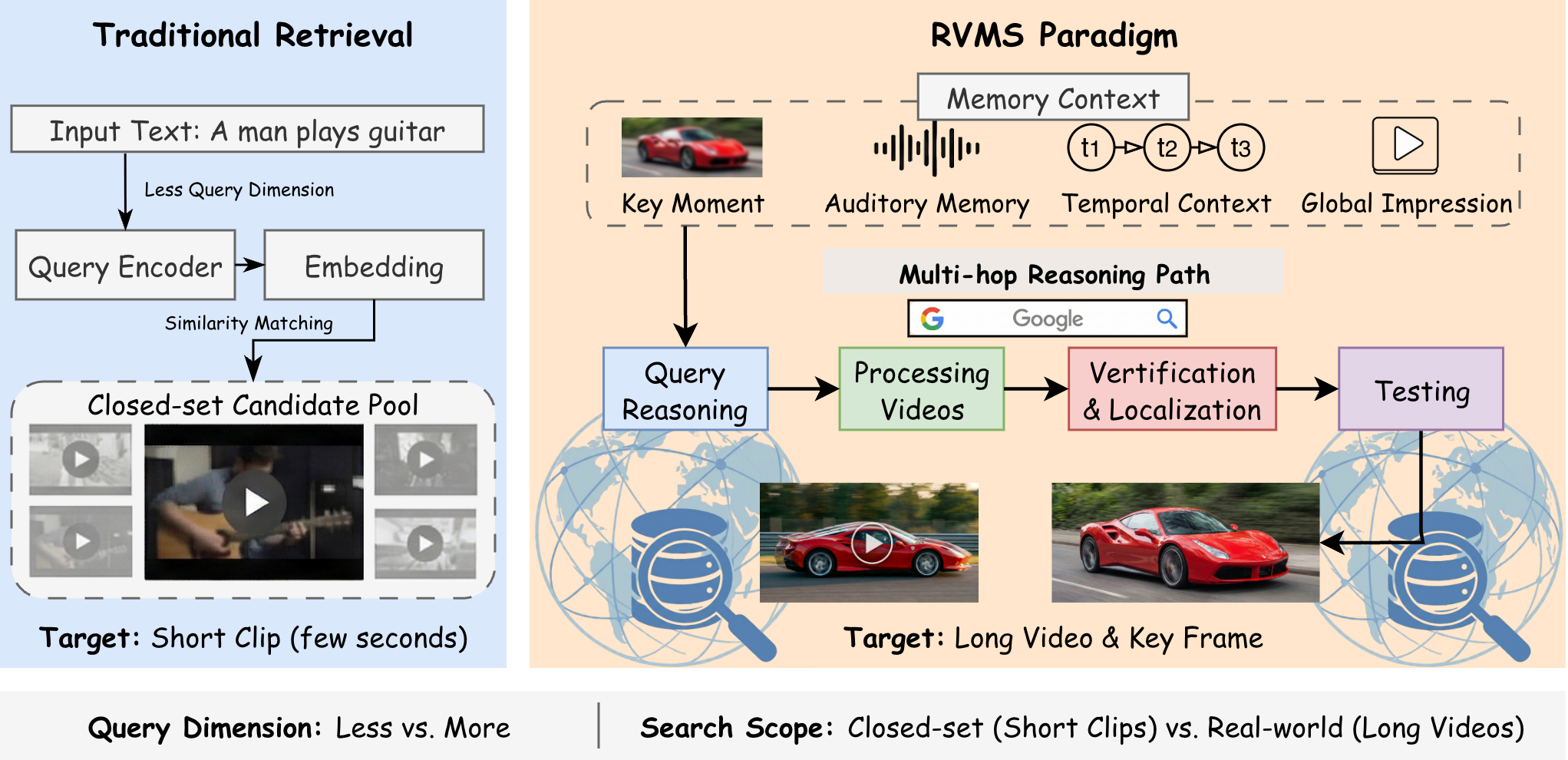}
  \caption{\textbf{Paradigm Shift in Video Retrieval.} Comparison between traditional video retrieval benchmarks (left) and our proposed RVMS-Bench (right). While traditional methods typically focus on single-dimension similarity matching within a closed candidate pool of short clips, RVMS addresses the real-world challenge of retrieving full-length videos from the open internet using multi-dimensional, fragmented memory cues (Global Impression, Key Moment, Temporal Context, and Auditory Memory).}
  \label{fig:teaser}
\end{figure*}

With the rapid advancement of Multimodal Large Language Models (MLLMs) \cite{radford2021learningtransferablevisualmodels, liu2023visualinstructiontuning, bai2025qwen3vltechnicalreport}, video retrieval has witnessed significant breakthroughs. However, current academic paradigms diverge sharply from real-world usage. While research typically simplifies the task to aligning precise, single-dimensional descriptions against a closed candidate pool \cite{Miech_2020_CVPR, bain2022frozentimejointvideo}, real-world users search the open internet using fuzzy, multi-dimensional memories of past viewing experiences (as shown in Figure \ref{fig:teaser}). This simplification reduces the core challenge to local semantic matching, ignoring the complexity of retrieving full-length videos using fragmented cues.

In real-world settings, users confront a massive, uncurated repository rather than a pre-prepared candidate set. Targets are often full-length videos—ranging from minutes to hours—requiring long-horizon reasoning based on fragmented cues rather than exact descriptions. Consequently, the true performance of MLLMs in the wild remains unknown, often overestimated due to saturated scores on simplified tasks. Methods optimized for local visual features struggle to transfer to hybrid internet queries involving complex temporal logic. Currently, the community lacks a benchmark that accurately reflects this high-entropy setting.

Existing benchmarks fall short in two key areas. First, regarding scope, they mainly focus on short clips within closed pools, failing to test retrieval in the vast, noisy open web \cite{Jiang_2019_ICCV, fang2024mmbenchvideolongformmultishotbenchmark, yuan2026momentseekertaskorientedbenchmarklongvideo}. Second, regarding dimension, real memories intertwine visual, auditory, and temporal context. Existing datasets prioritize visual matching, neglecting how users integrate auditory cues and temporal logic to locate key moments \cite{10.1007/978-3-031-19781-9_29, yang2025audiocentricvideounderstandingbenchmark, zhu2025ravstempoweringvideollmsfinegrained}.

To bridge the gap, we introduce RVMS-Bench (Real-World Video Memory Search Benchmark), a comprehensive system integrating high-quality internet video data with an agentic framework. Unlike datasets relying on coarse descriptions of short clips, RVMS-Bench spans twenty topics and four duration categories, employing a layered memory description system: Global Impression, Key Moment, Temporal Context, and Auditory Memory. To ensure quality, we implemented a standardized pipeline combining model-assisted generation (via Gemini 3 Pro) with human verification to eliminate logical inconsistencies and hallucinations.

Complementing the data, we propose RACLO, a retrieval agent based on abductive reasoning. RACLO simulates the human "Recall-Search-Verify" cognitive process, transforming fragmented clues into effective search queries to retrieve targets from the open web. Furthermore, considering the traditional ID-based matching fails due to video re-uploads and edits on the internet, we introduce a decoupled evaluation protocol for retrieval and localization, offering a robust measure of model performance against real-world noise.

The main contributions of this study are as follows:

\begin{itemize}
\item We construct RVMS-Bench, the first benchmark dedicated to real-world video memory search. By introducing a hierarchical description system—encompassing Global Impression, Key Moment, Temporal Context, and Auditory Memory—we move beyond factual descriptions of short clips to systematically evaluate fuzzy memory retrieval in the open world.
\item We propose RACLO, an agentic framework that simulates the human "Recall-Search-Verify" cognitive process. This method leverages abductive reasoning to transform fragmented memory cues into effective search queries, achieving precise target retrieval and moment localization within the vast internet corpus.
\item We conduct a comprehensive evaluation of state-of-the-art MLLMs, revealing that real-world video memory search remains a significant challenge for current models. This study establishes a testing framework and lays the foundation for future research in this domain.
\end{itemize}

\section{RVMS-Bench}


Unlike existing benchmarks that prioritize precise, single-dimension matching, RVMS-Bench is designed to simulate the unstructured, multi-dimensional fragmentation typical of real-world memory. In real-world search scenarios, users rarely articulate precise queries; instead, they rely on fragmented recollections—ranging from visual impressions and residual melodies to fuzzy temporal sequences. The lack of an evaluation environment restricts open-domain retrieval technology from achieving deeper semantic understanding and robustness, leaving current models ill-equipped to handle the information-scarce queries encountered in the wild.

To fill this void, we constructed a benchmark defined by three core attributes: (1) Multi-dimensional Memory Simulation. We established a hierarchical framework covering \textbf{Global Impression, Key Moment, Temporal Context, and Auditory Memory.} This design addresses the semantic gap where single-dimension queries fail to capture the complexity of multimodal memories. (2) Structured Task Space. By systematically combining these description elements, we constructed nine retrieval tasks with varying information densities. This structure allows us to diagnose model robustness, particularly when local details conflict with global themes. (3) High-Precision Quality Control. We implemented a pipeline combining model-assisted generation with human expert verification. This ensures that the ground-truth data maintains unique and accurate semantic directivity within the open-world context.

\begin{figure*}[t]
  \vskip 0.2in
  \centering
  \includegraphics[width=0.9\linewidth]{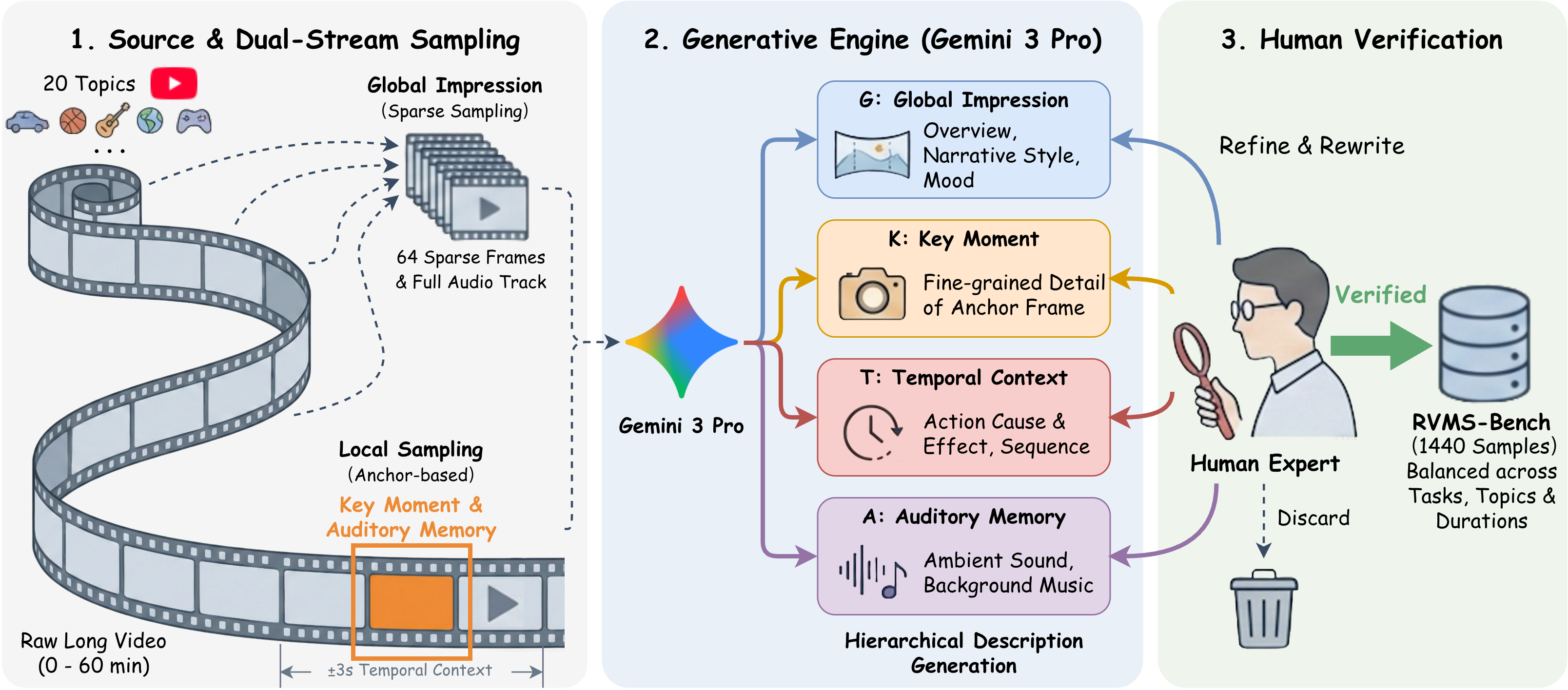}
  \caption{\textbf{Data Construction Pipeline of RVMS-Bench.} The pipeline integrates model-assisted generation with rigorous human verification to ensure scalability and quality. \textbf{Left:} We sample 20 topics from YouTube, employing dual-stream sampling for global sparse frames and local anchors. \textbf{Middle:} Gemini 3 Pro generates hierarchical descriptions covering Global Impression (G), Key Moment (K), Temporal Context (T), and Auditory Memory (A). \textbf{Right:} Human experts verify and refine the generated content to eliminate hallucinations, ensuring the semantic uniqueness of the ground truth.}
  \label{fig:pipeline}
\end{figure*}

\subsection{Data Collection and Categorization}


The data structure of the benchmark is defined by three core variables: \textbf{task type}, \textbf{video topic}, and \textbf{video duration}. During data collection, we strictly controlled the distribution of these variables to ensure they remain statistically independent and balanced. This structured sampling prevents evaluation biases arising from long-tail distributions, forcing models to acquire generalized retrieval capabilities rather than relying on domain-specific priors or duration preferences \cite{monfort2019momentstimedatasetmillion, shao2020finegym, cai2025lovrbenchmarklongvideo}.

\subsubsection{Task Definition}


To simulate the varying clarity of user memory in real-world scenarios, we formalize the retrieval task as a systematic combination of four key cues: \textbf{Global Impression (G), Key Moment (K), Temporal Context (T), and Auditory Memory (A)}. Based on the availability of these cues, we construct nine distinct query types with varying information densities.The fundamental uni-modal tasks include the G-task and K-task. These assess the model's capacity for macro-thematic summarization and micro-visual feature matching, respectively.

To reflect realistic complexity, we introduce hybrid query forms. Dual-modal tasks—comprising K+T, K+A, and K+G—require the model to comprehend temporal logic, achieve cross-modal alignment of audio-visual signals, or resolve the semantic associations between local snapshots and global contexts. More complex memory states are represented by tri-modal tasks (K+G+T, K+G+A, K+T+A). Finally, the comprehensive K+G+T+A query integrates all dimensions, corresponding to the ideal scenario where user memory is at its clearest.

\subsubsection{Video Topics and Duration}


For video content selection, we aligned with the taxonomy of mainstream platforms to curate twenty distinct topics, ensuring coverage of the internet's primary video modalities. These categories include Variety Shows, Animation, Dance, Tech and Digital, Food, Automobiles, Sports, Lifestyle Vlogs, Movies, TV Series, Documentaries, Gaming, Music, Education, Fashion, Animals, Art, Travel, Fitness, and Parenting. This extensive breadth exposes models to a wide spectrum of semantic challenges—ranging from information-dense knowledge exposition to highly dynamic artistic performances. Such diversity compels models to develop generalized multimodal understanding, preventing reliance on domain-specific structural priors or heuristics.

Beyond content types, video duration serves as a critical structural variable influencing retrieval difficulty. To systematically analyze its impact, we stratified the dataset into four intervals based on total duration: under 3 minutes, 3-10 minutes, 10-30 minutes, and 30-60 minutes. Unlike existing benchmarks dominated by short-form content, our dataset significantly increases the proportion of long and ultra-long videos. This design rigorously tests the long-context temporal reasoning capabilities of current models.

\subsection{Generation Pipeline}


To efficiently construct a large-scale, structurally consistent benchmark, we sourced raw videos from YouTube and implemented a standardized pipeline combining model-based generation with human verification (Figure \ref{fig:pipeline}). This workflow employs MLLMs for preliminary semantic extraction, followed by rigorous human-machine collaboration for noise reduction and refinement. This hybrid strategy ensures high-quality annotation while maximizing production efficiency \cite{11206011, zhang2025vcagentinteractiveagentcustomized}.

\subsubsection{Model-based Description Generation}


In the generation phase, we utilize Gemini 3 Pro as the backbone to extract multi-dimensional semantic information. For local fine-grained analysis targeting Key Moments, the pipeline randomly samples an anchor frame, instructing the model to focus exclusively on its visual details. To capture Temporal Context and Auditory Memory, we construct a spatiotemporal window centered on this anchor. Specifically, we extract image sequences at 1 FPS within a $\pm 3$-second interval and a corresponding one-second audio segment. The model analyzes these visual transitions to describe the action's antecedents and consequences, while simultaneously characterizing environmental sounds and background music. The prompts are in Appendix \ref{sec:generation}.

Conversely, for the Global Impression, we employ a global sparse sampling strategy. We uniformly extract 64 frames across the full video duration and feed them, alongside the complete audio track, into the model. The model synthesizes visual and auditory signals to produce a holistic description encompassing overall content, narrative style, and emotional tonality. This local-to-global paradigm ensures the generated text spans the full multimodal semantic space, ranging from instantaneous details to the overarching atmosphere.

\subsubsection{Human Verification and Refinement}


To guarantee data reliability and eliminate potential hallucinations, we implemented a rigorous expert verification stage. We recruited 10 researchers specializing in multimodal understanding to manually verify every generated sample. Adhering to strict objective consistency protocols, annotators cross-referenced the generated text with raw videos, explicitly targeting visual factual errors, chronological inconsistencies, and auditory misrepresentations. Samples deemed substandard underwent immediate correction or rewriting. This human-in-the-loop filtration process ensures that every dataset entry serves as a high-quality ground truth, maintaining the semantic uniqueness essential for precise retrieval.

\begin{figure*}[t]
  \vskip 0.2in
  \centering
  \includegraphics[width=\linewidth]{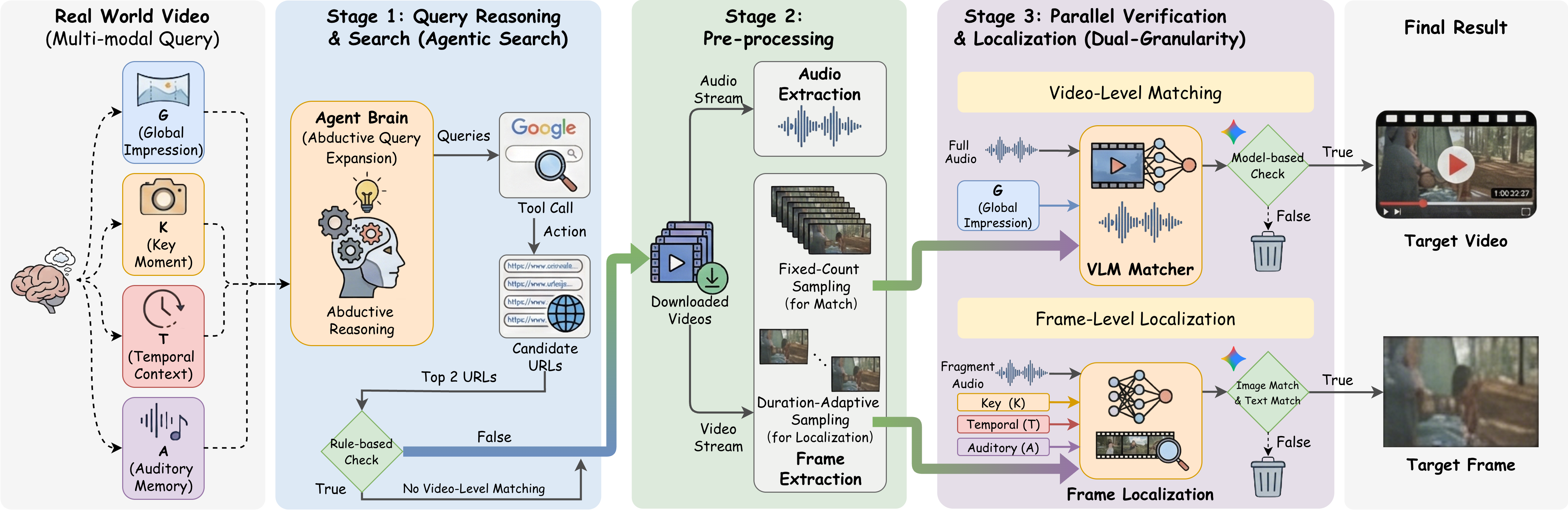}
  \caption{\textbf{Overview of the RACLO Framework.} The framework mimics the human "Recall-Search-Verify" cognitive loop. \textbf{Stage 1 (Query Reasoning \& Search):} The Agent Brain employs abductive reasoning to translate fragmented multimodal cues into search queries, retrieving candidate URLs from search engines. \textbf{Stage 2 (Pre-processing):} Candidates are downloaded and processed into audio-visual streams. \textbf{Stage 3 (Parallel Verification \& Localization):} A dual-granularity mechanism validates video content against the Global Impression (G) while simultaneously localizing the target frame using Key Moment (K), Temporal Context (T), and Auditory Memory (A).}
  \label{fig:method}
\end{figure*}

\subsection{Dataset Statistics}


The resulting RVMS-Bench comprises 1,440 samples with a highly balanced distribution. To ensure unbiased evaluation, we enforced strict uniformity across three dimensions: 
\begin{itemize} 
\item \textbf{Task Distribution:} The dataset is perfectly balanced across nine task types, with 160 samples per type. 
\item \textbf{Topic Distribution:} We achieve uniform coverage of the 20 distinct topics, assigning 72 samples to each. 
\item \textbf{Duration Distribution:} Samples are equally stratified into four duration intervals, with 360 samples per interval. 
\end{itemize} 
This structured design minimizes evaluation bias caused by long-tail distributions, ensuring fair comparisons across diverse settings\cite{yang2025vimix14mcuratedmultisourcevideotext, ju2024miradatalargescalevideodataset}. Detailed distribution visualizations are provided in Appendix \ref{1}.

\section{Method: The RACLO Framework}


To bridge the semantic gap between fuzzy, subjective memories and precise, objective internet indices, we introduce the RACLO (Real-world Abductive Cognitive Logic Retrieval) framework. As illustrated in Figure \ref{fig:method}, RACLO simulates the human "Recall-Search-Verify" cognitive process.

The framework operates in two stages. First, it employs abductive reasoning \cite{687481, 10.5555/2856823.2856863, fariha2019exampledrivenqueryintentdiscovery} to translate fragmented memory cues into search-oriented keywords, retrieving candidate videos from the open web. Subsequently, we introduce a parallel verification mechanism that analyzes content at both video and frame levels. This design achieves the dual goals of accurately retrieving the target video from the massive internet corpus and precisely localizing key moments.

\subsection{Query Reasoning and Retrieval}

\subsubsection{Abductive Query Expansion}


The primary challenge in fuzzy memory search is the inefficiency of direct keyword matching based on raw descriptions. User memories typically consist of fragmented visual snapshots and vague global impressions, lacking the high-level semantic concepts or precise metadata required for direct alignment with target videos.

To address this, we incorporate abductive reasoning as the core of our query generation. Moving beyond simple explicit entity extraction, our model analyzes visual styles, auditory cues, and latent temporal information to reverse-engineer probable video titles, categories, or specific tags. This mechanism constructs augmented queries enriched with latent semantic associations, significantly enhancing retrieval recall in real-world open settings.

To guide the agent in performing effective abductive reasoning and generating high-quality search queries, we designed a system prompt incorporating Chain-of-Thought (CoT) instructions. This prompt requires the model to first analyze fragmented cues within the fuzzy memory and then formulate a search strategy. The full prompt for the search agent is provided in Appendix \ref{sec:abductive}.

\subsubsection{Agent-based Web Search}


To address the dynamic complexity of the open web, our agent employs a ReAct-based "Observe-Think-Act" paradigm \cite{yao2023reactsynergizingreasoningacting}. First, the agent generates an internal chain of thought via specific tags to determine the search direction. Subsequently, it uses tool invocation tags to issue specific queries to search engines. From the raw results, it filters and identifies $n$ accessible YouTube URLs, downloading them to the local environment. This process condenses massive open-domain noise into a compact, high-quality candidate set, establishing a solid basis for subsequent fine-grained processing.

\subsection{Parallel Video Verification and Localization}



Once candidates are acquired, the system must satisfy two simultaneous objectives: (1) accurately identifying the target video, and (2) precisely localizing the key moment. We adopt a parallel global-local analysis strategy to handle video-level semantics and frame-level details independently.

\textbf{Video Retrieval Verification} We implement a hierarchical verification pipeline. First, rule-based exact URL matching serves as a fast path; if the retrieved URL matches the ground truth, success is declared immediately, bypassing further steps. If the URL does not match (e.g., due to re-uploads), we trigger content-based verification. We uniformly sample 64 frames and the complete audio track, inputting them into the primary model to assess alignment with the Global Impression (G) in terms of semantics, narrative style, and tone. If rejected, the process terminates. If accepted, we employ Gemini 2.5 Pro for secondary validation. Retrieval is considered successful only upon consensus between both models. The specific prompts used for video retrieval verification are detailed in Appendix \ref{sec:retrieval}.

\textbf{Moment Localization} In parallel, the system handles moment localization. We feed the multi-dimensional context—integrating Key Moment (K), Temporal Context, and Auditory Memory—into the model. The model performs cross-modal reasoning on dense frame sequences, leveraging audio-visual synchronization cues to directly predict the frame index with the highest semantic correspondence. Finally, we conduct a verification step by inputting the multi-dimensional context, the ground-truth frame, and the predicted frame into Gemini 2.5 Pro to certify the match.The complete prompt templates used for the localization task are available in Appendix \ref{sec:moment}. Additionally, we provide a reliability study comparing this model-based evaluation against human annotations in Appendix \ref{sec:human_agreement}.

\section{Experiments}

\subsection{Baselines and Settings}


To evaluate the performance of different models, we selected closed-source and open-source models as baselines. The closed-source models include Gemini 3 Pro, Claude 4.0 Sonnet, GPT-4o, GPT-5.2, GPT-5-mini and Hunyuan-Video. The open-source models include Qwen3-VL-235B-A22B-Instruct \cite{bai2025qwen3vltechnicalreport}, and Qwen2.5-VL-72B-Instruct \cite{bai2025qwen25vltechnicalreport}. We follow the video frame sampling settings used in VLMEvalKit \citep{duan2024vlmevalkit} and VideoMME \citep{fu2025videommefirstevercomprehensiveevaluation}. In the video retrieval task, all models uniformly extract 64 frames. In the moment localization task, for closed-source models, the frame extraction settings vary based on the video duration: for videos shorter than 3 minutes, 32 frames are uniformly extracted; for videos between 3 and 10 minutes, 64 frames are extracted; for videos between 10 and 30 minutes, 128 frames are extracted; and for videos between 30 and 60 minutes, 192 frames are extracted. For open-source models, 64 frames are uniformly extracted from a video. Each model generates 2 queries, with each query retrieving 2 URLs.

\subsection{Video Retrieval Evaluation}



\textbf{Overall Retrieval Performance} Table \ref{tab:video_retrieval} illustrates distinct performance tiers, with Gemini 3 Pro establishing SOTA performance. In the open-source domain, Qwen3-VL-235B demonstrates remarkable competitiveness, surpassing even prominent closed-source models like Claude-3.5-Sonnet and GPT-4o. Notably, transitioning from "Rule-only" to Model+Rule yields substantial gains across all models (e.g., +36.4\% for Gemini 3 Pro). This confirms that visual content verification is essential for retrieving content-correct targets in the open web, effectively overcoming the limitations of metadata-based URL matching prone to re-uploads.

\textbf{Dependency of the Verification Mechanism} It is important to note that the Model+Rule results in Table \ref{tab:video_retrieval} are reported only for query tasks that include Global Impression (G). This is because RACLO's verification module is logically strictly dependent on the global semantic description as a benchmark for consistency judgment. In the absence of G cues, the system cannot establish an effective binary classification boundary, and therefore, corresponding entries were not included in the statistics.

\begin{table*}[t]
\centering
\setlength{\tabcolsep}{1.5pt}
\caption{\textbf{Video Retrieval Performance.} Comparison of retrieval recall (\%) between Rule-only baseline and Model+Rule. Models are grouped by source availability. \textbf{Note: All Qwen models are Instruct versions.} Best performance is \textbf{bolded}, second best is \underline{underlined}.}
\label{tab:video_retrieval}
\resizebox{\textwidth}{!}{
\begin{tabular}{l|cccccccccc|cccccccccc}
\toprule
\multirow{2}{*}{\textbf{Model}} & \multicolumn{10}{c}{\textbf{Rule Only}} & \multicolumn{10}{c}{\textbf{Model + Rule}} \\
\cmidrule(lr){2-11} \cmidrule(lr){12-21}
 & \textbf{G} & \textbf{K} & \textbf{KT} & \textbf{GK} & \textbf{KA} & \textbf{KTA} & \textbf{GKA} & \textbf{GKT} & \textbf{GKTA} & \textbf{Avg} & \textbf{G} & \textbf{K} & \textbf{KT} & \textbf{GK} & \textbf{KA} & \textbf{KTA} & \textbf{GKA} & \textbf{GKT} & \textbf{GKTA} & \textbf{Avg} \\
\midrule
Human & 64.0 & 28.0 & 24.0 & 88.0 & 24.0 & 56.0 & 64.0 & 60.0 & 80.0 & 54.2 & 64.0 & - & - & 88.0 & - & - & 72.0 & 60.0 & 80.0 & 72.8 \\
\midrule
\multicolumn{21}{c}{\textit{Closed Source Models}} \\
\midrule
Gemini-3-Pro & 14.4 & \textbf{23.8} & \textbf{16.9} & \textbf{16.9} & \textbf{20.0} & \textbf{23.8} & \textbf{31.9} & 15.6 & \underline{22.5} & \textbf{20.6} & \textbf{45.6} & - & - & \textbf{58.8} & - & - & \textbf{63.1} & \textbf{56.9} & \textbf{61.2} & \textbf{57.0} \\
GPT-5.2 & 14.4 & 3.2 & 1.6 & 12.9 & 6.9 & 5.0 & 13.1 & \underline{16.2} & 15.0 & 9.8 & 26.2 & - & - & 36.9 & - & - & 38.1 & 31.2 & \underline{36.2} & 33.8 \\
Gemini-2.5-Pro & \underline{18.8} & 13.1 & 10.6 & \underline{15.0} & 12.5 & 13.8 & \underline{18.1} & \underline{16.2} & 17.5 & 15.1 & \underline{36.2} & - & - & 28.8 & - & - & 34.4 & 30.0 & 24.4 & 31.0 \\
GPT-5-Mini & 16.1 & 1.9 & 1.9 & 10.6 & 4.4 & 4.0 & 7.0 & 13.8 & 13.1 & 8.1 & 31.2 & - & - & 27.5 & - & - & 28.8 & 28.8 & 27.5 & 28.8 \\
Hunyuan-Video & 15.6 & 10.6 & 8.8 & 12.5 & 10.0 & 11.2 & 15.0 & 13.1 & 14.4 & 12.4 & 28.8 & - & - & 22.5 & - & - & 27.5 & 23.8 & 19.4 & 24.4 \\
Claude-4.5-Sonnet & 12.5 & 10.6 & 8.8 & 12.5 & 10.0 & 11.2 & 14.4 & 11.2 & 13.1 & 11.6 & 23.8 & - & - & 24.4 & - & - & 26.2 & 21.9 & 23.8 & 24.0 \\
GPT-4o & 10.6 & 7.5 & 6.2 & 10.0 & 7.5 & 8.1 & 11.9 & 10.0 & 11.2 & 9.2 & 18.8 & - & - & 19.4 & - & - & 21.9 & 17.5 & 18.8 & 19.2 \\
\midrule
\multicolumn{21}{c}{\textit{Open Source Models}} \\
\midrule
Qwen3-VL-235B & \textbf{21.9} & \underline{15.1} & \underline{12.4} & 13.1 & \underline{15.0} & \underline{16.1} & 11.9 & \textbf{23.1} & \textbf{30.0} & \underline{17.6} & 32.5 & - & - & \underline{42.5} & - & - & \underline{40.0} & \underline{43.8} & 25.0 & \underline{36.6} \\
Qwen2.5-VL-72B & 16.9 & 11.9 & 10.0 & 11.9 & 11.2 & 12.5 & 17.5 & 15.0 & 16.2 & 13.7 & 29.4 & - & - & 23.1 & - & - & 33.8 & 26.2 & 28.8 & 28.1 \\
\bottomrule
\end{tabular}
}
\vskip -0.1in
\end{table*}

\subsection{Moment Localization Evaluation}



\textbf{Overall Localization Performance} As shown in Table \ref{tab:frame_localization}, moment localization presents a significantly higher challenge than retrieval, demanding precise alignment between fine-grained visual details and temporal logic. GPT-5.2 achieves the best performance, notably improving to 50.6\% under the comprehensive GKTA setting. This demonstrates that leading proprietary models possess robust long-horizon reasoning capabilities to effectively leverage multimodal contexts. However, the general performance drop across most models reflects a prevalent challenge for current MLLMs in bridging the gap between macro-semantic matching and micro-level frame grounding.

\textbf{Discrepancy between Semantic and Temporal Capabilities} Experimental data reveals a significant performance disparity for open-source models. Specifically, Qwen3-VL-235B achieves strong retrieval recall (36.6\%) via image-text alignment but drops to 4.0\% in localization. This contrast indicates that while models with native long-context support excel in coarse-grained content recognition, they encounter a bottleneck when mapping fuzzy memory cues to specific timestamps. This suggests that although these models capture the global video context, they struggle to maintain high-fidelity spatiotemporal alignment when processing uncurated long videos and ambiguous queries.

\begin{table*}[t]
\centering
\caption{\textbf{Frame Localization Performance.} Evaluation of moment localization accuracy (\%) using the VLM-based metric. \textbf{Note: All Qwen models are Instruct versions.} Best performance is \textbf{bolded}, second best is \underline{underlined}.}
\label{tab:frame_localization}
\resizebox{0.73\textwidth}{!}{
\begin{tabular}{l|cccccccccc}
\toprule
\multirow{2}{*}{\textbf{Model}} & \multicolumn{10}{c}{\textbf{Moment Localization}} \\
\cmidrule(lr){2-11}
 & \textbf{G} & \textbf{K} & \textbf{KT} & \textbf{GK} & \textbf{KA} & \textbf{KTA} & \textbf{GKA} & \textbf{GKT} & \textbf{GKTA} & \textbf{Avg} \\
\midrule
Human & - & 28.0 & 32.0 & 92.0 & 32.0 & 64.0 & 72.0 & 64.0 & 80.0 & 58.0 \\
\midrule
\multicolumn{11}{c}{\textit{Closed Source Models}} \\
\midrule
GPT-5.2 & - & \textbf{35.6} & \textbf{38.8} & \textbf{36.9} & \textbf{39.4} & \textbf{41.9} & \textbf{35.6} & \underline{33.1} & \textbf{50.6} & \textbf{38.9} \\
Gemini-3-Pro & - & \underline{18.1} & \underline{30.0} & \underline{32.5} & \underline{22.5} & \underline{34.4} & \underline{30.6} & \textbf{36.2} & \underline{32.5} & \underline{29.5} \\
GPT-4o & - & 15.0 & 16.9 & 16.9 & 16.2 & 19.4 & 16.2 & 15.6 & 25.0 & 17.6 \\
Claude-4.5-Sonnet & - & 5.6 & 15.0 & 12.5 & 7.5 & 13.1 & 10.6 & 11.9 & 17.5 & 11.8 \\
Gemini-2.5-Pro & - & 6.2 & 18.8 & 11.9 & 10.0 & 8.1 & 9.4 & 8.1 & 21.2 & 11.6 \\
GPT-5-Mini & - & 6.9 & 8.8 & 10.0 & 7.5 & 11.9 & 10.6 & 10.6 & 16.2 & 10.3 \\
Hunyuan-Video & - & 5.0 & 15.0 & 9.4 & 8.1 & 6.2 & 7.5 & 6.2 & 16.9 & 9.1 \\
\midrule
\multicolumn{11}{c}{\textit{Open Source Models}} \\
\midrule
Qwen3-VL-235B & - & 1.9 & 3.1 & 5.6 & 1.2 & 2.5 & 3.8 & 6.2 & 6.9 & 4.0 \\
Qwen2.5-VL-72B & - & 0.0 & 0.0 & 5.0 & 2.5 & 5.0 & 3.1 & 2.5 & 2.5 & 2.6 \\
\bottomrule
\end{tabular}
}
\vskip -0.1in
\end{table*}

\subsection{Performance Breakdown and Robustness}





To provide a deeper understanding of model capabilities, we conduct a detailed performance breakdown across task types, video categories, and durations. Comprehensive visualizations are provided in Appendix \ref{2}.

\textbf{Synergy of Multi-Dimensional Cues} We analyzed success rates across different task types. Results indicate that increasing query dimensions does not degrade performance due to added constraints; instead, it correlates positively with retrieval success. Models achieve peak performance on the full-cue GKTA task, suggesting effective complementarity among multimodal inputs within the RACLO framework. This synergy enables models to lock onto targets more precisely than uni-modal queries.

\textbf{Impact of Content Structure.} Video categories significantly influence model performance. Models excel in structured content such as Movies, Documentaries, and Tech, characterized by clear narrative logic and distinct visual features. Conversely, performance fluctuates noticeably in categories like Variety Shows or Lifestyle Vlogs, which feature loose structures, rapid cuts, and substantial background noise. This reveals current limitations in handling unstructured, cluttered content, where extracting coherent semantic cues from disordered visual streams remains challenging.

\textbf{Stress Test on Video Duration} Video duration serves as a core metric for evaluating robustness. As duration increases from short (<3 min) to long (>30 min), all models exhibit a downward trend in retrieval and localization. Notably, Gemini 3 Pro and GPT-5.2 demonstrate superior stability with relatively flat decay curves; in contrast, open-source models suffer significant performance drops in videos exceeding 30 minutes. This confirms that insufficient long-horizon consistency remains the primary bottleneck restricting open-source models in long-video processing.

\begin{figure*}[t]
    \vskip 0.2in
    \centering
    \includegraphics[width=0.96\textwidth]{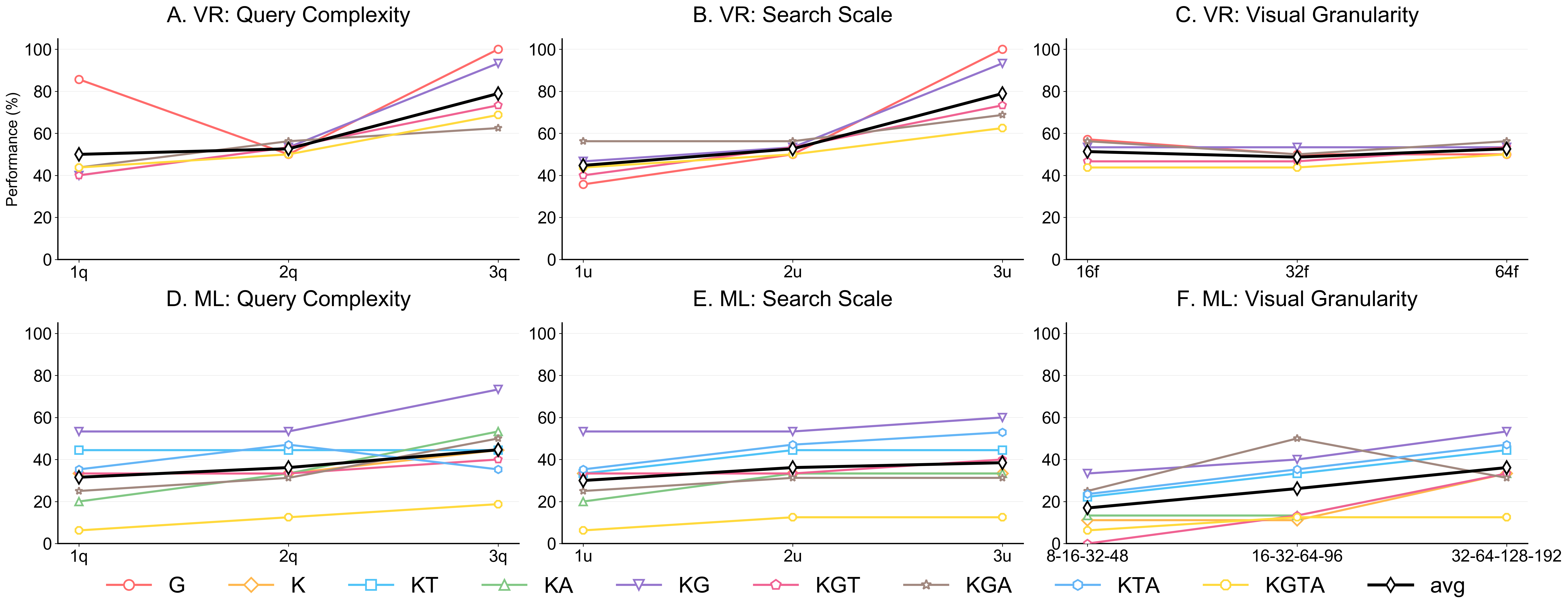}
    \caption{Ablation studies for Video Retrieval (VR) and Moment Localization (ML) tasks. 
    The panels illustrate performance sensitivity across three key technical dimensions: 
    (1) \textbf{Query Complexity}, representing the number of expanded search keywords generated for each query; 
    (2) \textbf{Search Scale}, indicating the number of candidate URLs returned per retrieval execution; 
    (3) \textbf{Visual Granularity}, referring to the total number of frames sampled and extracted from a single video for processing. }
    \label{fig:ablation_detailed}
\end{figure*}

\subsection{Further Analysis}




To ensure efficiency while maintaining statistical representativeness, we randomly sampled 10\% of the data from the main experiment for the following ablation studies.

\textbf{Impact of Query and Search Scale.} As shown in Figure \ref{fig:ablation_detailed} (A, B, D, E), query complexity ($q$) and search breadth ($u$) primarily affect Video Retrieval (VR). Specifically, VR recall grows linearly with both factors, confirming that using diverse keywords and checking more candidate videos effectively reduces miss rates. However, these factors have minimal impact on Moment Localization (ML), as localization relies mainly on internal visual matching rather than the richness of the query. Performance stagnates beyond $u=2$, indicating that simply adding more candidate videos provides no extra benefit for localization accuracy.

\textbf{Impact of Visual Granularity.} Visual granularity is the determining factor for Moment Localization but has little impact on Video Retrieval (Figure \ref{fig:ablation_detailed} C vs. F). For Video Retrieval, performance remains stable across sampling rates, indicating that the Global Impression consists of macro features that are recognizable even with sparse sampling (16f). However, ML accuracy improves significantly with higher sampling rates. This is because Key Moments are inherently fleeting; therefore, denser visual sampling yields a higher probability of capturing these short segments, whereas sparse sampling is prone to missing them.

\textbf{Failure Cases}
In this study, we analyze common pipeline errors. Query reasoning errors arise when the MLLM misinterprets the input, producing incorrect or incomplete search expansions and leading to irrelevant or insufficient candidate videos. Retrieval errors are returning irrelevant results and missing valid videos due to limited retrieval coverage. Even when video is successfully retrieved, video verification errors occur during video matching, where the evaluated model produces judgments that contradict human annotations. Localization errors arise when the evaluated model fails to identify the correct shot or when the evaluated model incorrectly assesses otherwise accurate localization results. Detailed case analyses are provided in Appendix~\ref{case_study}.

\section{Related Work}
\label{sec:related_work}
\textbf{Cross-Modal Video Retrieval Benchmarks} The field has long relied on datasets like MSR-VTT, VATEX and VSR \cite{Xu_2016_CVPR, wang2020vatexlargescalehighqualitymultilingual, liu2023visualspatialreasoning}, which primarily consist of short video clips paired with objective, factual descriptions \cite{Xu_2016_CVPR}. Such characteristics tend to bias models toward shallow matching between static frames and text, leading to performance saturation that stems from an over-reliance on static features rather than temporal logic. While newer works like LoVR and MUVR \cite{cai2025lovrbenchmarklongvideo, feng2025muvrmultimodaluntrimmedvideo} introduce long-form content and fine-grained annotations, they still operate on the idealized assumption that query texts precisely describe video content. They fail to cover the memory biases and subjective ambiguities prevalent in real-world scenarios---such as residual melody fragments or a generalized perception of the atmosphere. Due to this lack of fuzzy, fragmented cues, assessing the true robustness of models under complex real-world conditions remains difficult. To address this, we introduce RVMS-Bench to evaluate model performance under realistic, non-ideal query conditions.

\textbf{Agent-Based Open-Domain Retrieval} Open-domain video retrieval aims to break the limitations of closed databases by retrieving target content directly from massive, uncurated internet videos \cite{yang2022tubedetrspatiotemporalvideogrounding}. Unlike traditional web searching that relies on metadata or keyword matching, recent works such as V-Agent, VideoDeepResearch and ViperGPT \cite{park2026vagentinteractivevideosearch, yuan2025videoexplorerthinkvideosagentic, surís2023vipergptvisualinferencepython} have begun to explore using Large Language Models as agents, employing tool invocation and web search interfaces to handle complex natural language queries. While promising, these methods currently face a lack of evaluation standards, often relying on self-constructed, small-scale query sets or qualitative analysis. There is a lack of a unified benchmark to quantitatively assess how agents cope with real-world challenges when facing users' fuzzy memories---specifically, how they handle fragmented multimodal clues, video version redundancy, and long-term temporal interference. This gap in the evaluation environment restricts the development of open-domain retrieval technologies toward deeper semantic understanding and robustness. To fill this gap, we propose the RACLO framework and RVMS-Bench, providing a unified evaluation platform for the challenging task of retrieving target videos and localizing key moments from the open web.

\section{Conclusion}
We introduced RVMS-Bench and RACLO to highlight the gap between idealized academic tasks and the demands of real-world video memory search. RVMS-Bench provides \textbf{1,440} human-verified samples with multi-dimensional memory cues (\textbf{G/K/T/A}), covering diverse topics, durations, and tasks. Experimental results indicate that existing models still exhibit significant deficiencies in open-web video retrieval and precise moment localization under real-world scenarios based on fuzzy memories. We hope this study encourages future research to move beyond simple semantic matching and focus on developing active retrieval agents capable of reasoning through real-world open-web videos.

\section*{Impact Statement}

This paper presents work whose goal is to advance the field of Machine Learning, specifically in the area of video retrieval. There are many potential societal consequences of our work, none which we feel must be specifically highlighted here.






\setcitestyle{numbers,square}

\bibliography{citation}

\newpage
\appendix

\section{Additional Dataset Statistics and Analysis}
\label{sec:appendix_stats}


In this section, we provide supplementary visualizations regarding the dataset composition and fine-grained performance analysis of our proposed method.

\subsection{Dataset Details}
\label{1}

\begin{figure}[h]
    \vskip 0.2in
    \centering
    \includegraphics[width=1\linewidth]{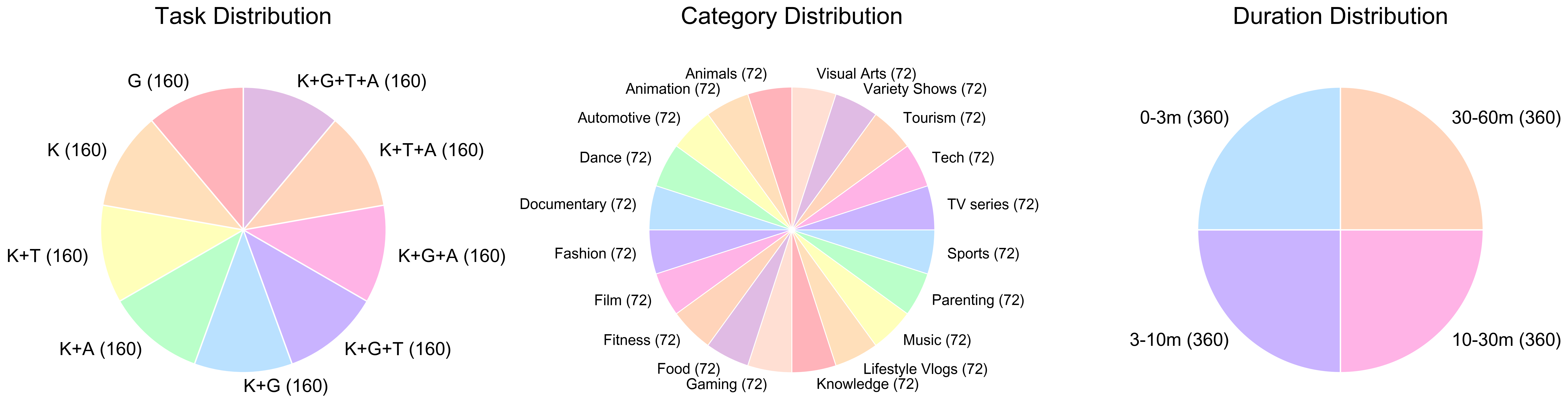}
    \caption{\textbf{Data Distributions of RVMS-Bench.} Breakdowns of video duration, category, and task types.}
    \label{fig:data_dist}
\end{figure}

Figure~\ref{fig:data_dist} visualizes the composition of the RVMS-Bench dataset.

\subsection{Performance Breakdown}
\label{2}

\begin{figure}[H]
    \vskip 0.2in
    \centering
    \includegraphics[width=1\linewidth]{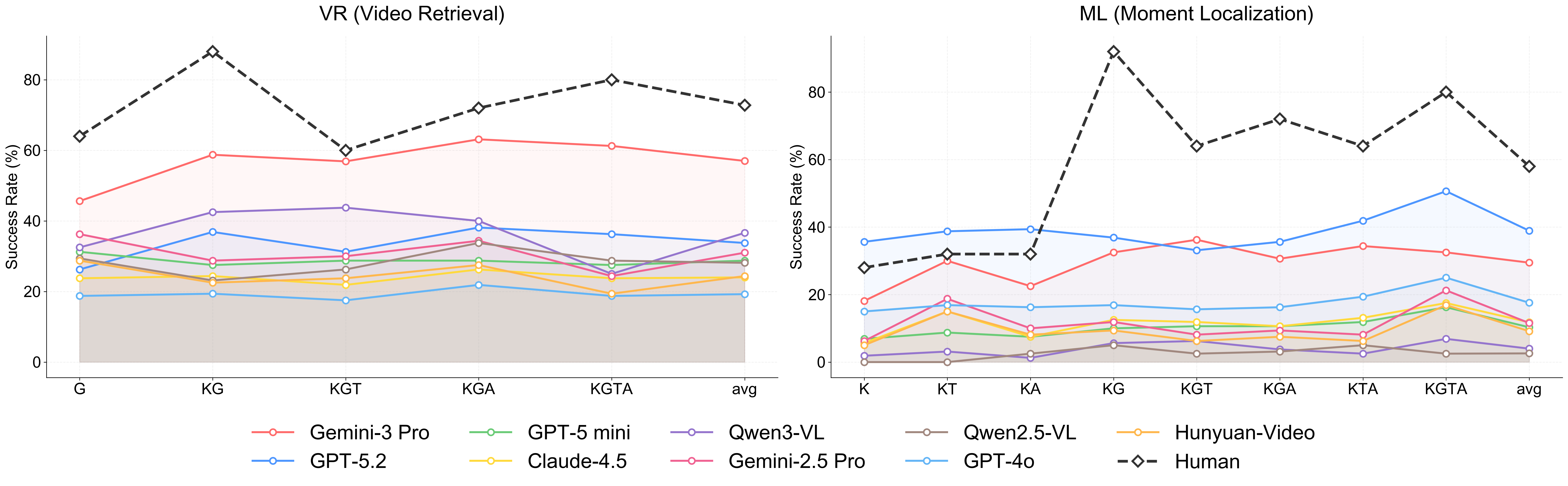}
    \caption{\textbf{Success Rate Across Different Tasks.}}
    \label{fig:task}
\end{figure}

\begin{figure}[H]
    \vskip 0.2in
    \centering
    \includegraphics[width=1\linewidth]{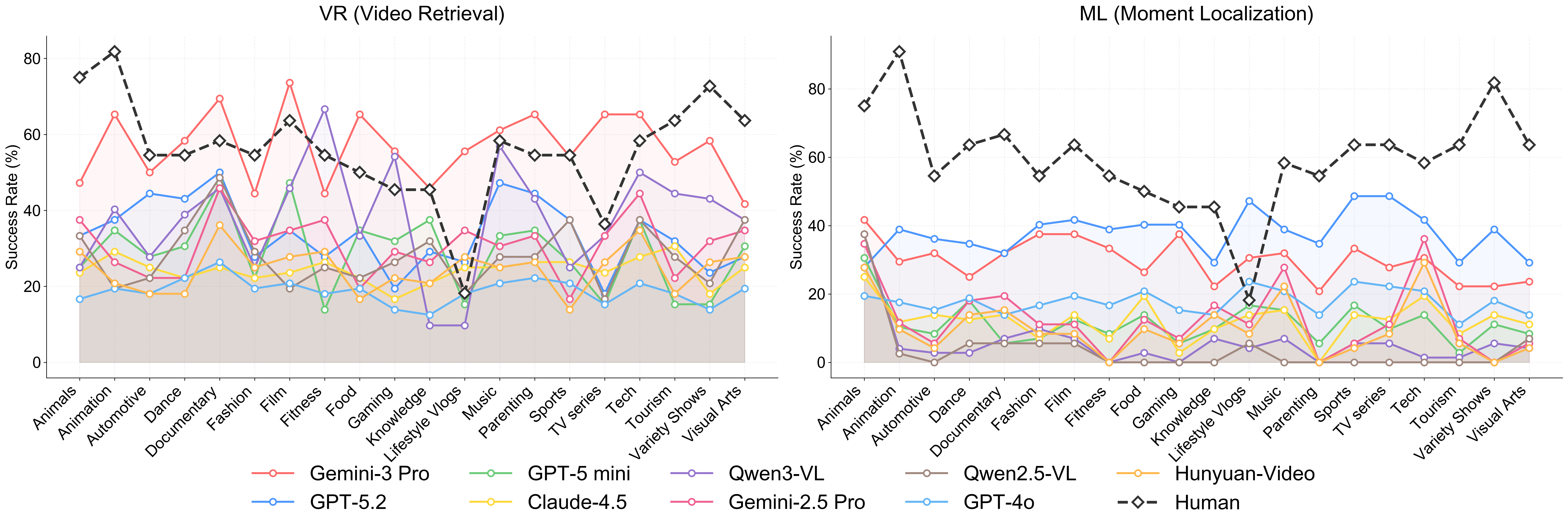}
    \caption{\textbf{Success Rate Across Video Categories.}}
    \label{fig:category}
\end{figure}

\begin{figure}[H]
    \vskip 0.2in
    \centering
    \includegraphics[width=1\linewidth]{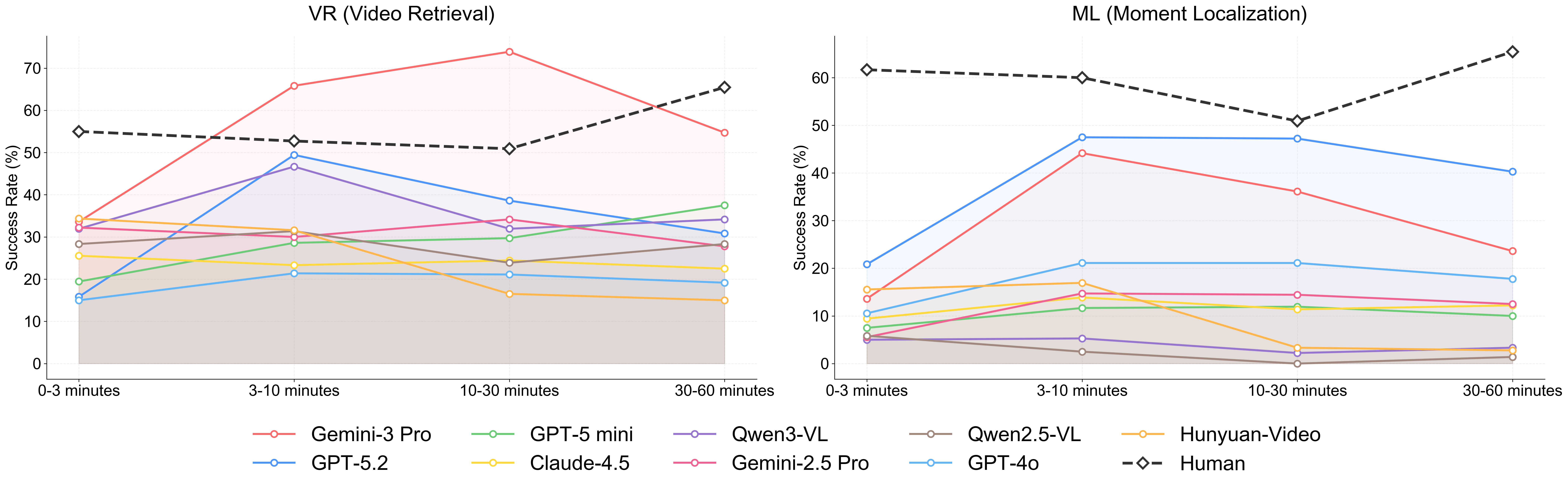}
    \caption{\textbf{Impact of Video Duration on Success Rate.}}
    \label{fig:duration}
\end{figure}

We analyze the success rate across tasks, categories, and durations (Figure~\ref{fig:task} Figure~\ref{fig:category} Figure~\ref{fig:duration})

\begin{figure*}[h]
  \vskip 0.2in
  \centering
  \includegraphics[width=0.85\linewidth]{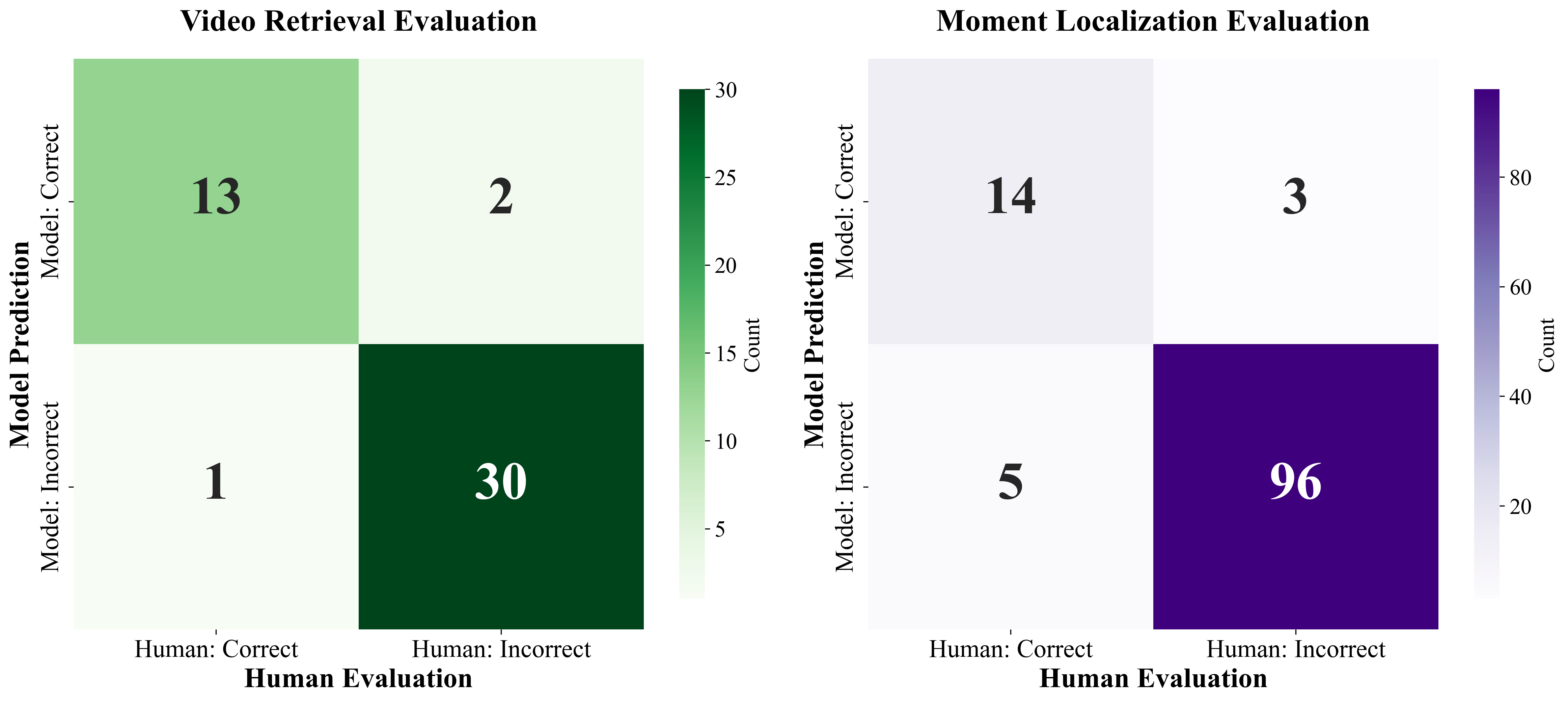} 
  \caption{\textbf{Alignment between Automated Evaluation and Human Judgment.}}
  \label{fig:human_matrix}
\end{figure*}

\section{Reliability of Automated Evaluation}
\label{sec:human_agreement}


To validate the reliability of our model-based evaluation pipeline, we conducted a human study on a randomly sampled subset of the test results. We employed human experts to manually review the model's judgments on both Video Retrieval and Moment Localization tasks.

Figure \ref{fig:human_matrix} shows the confusion matrices comparing the decisions made by automated evaluator against human ground truth.

\section{Generation Prompts}
\label{sec:generation}

The dataset annotation pipeline uses multiple VLM prompts to generate rich multimodal descriptions for each video clip. The pipeline extracts a target keyframe and surrounding context frames, then applies the following prompts. Four description fields are generated:

\begin{itemize}
  \setlength{\itemsep}{4pt}    
\item \textbf{Global Impression (G)}: A holistic description of the entire video's theme, atmosphere, and content development.
\item \textbf{Key Moment (K)}: A detailed description of the specific keyframe's visual content.
\item \textbf{Temporal Context (T)}: Descriptions of the actions and events occurring before and after the key moment.
\item \textbf{Auditory Memory (A)}: Analysis of the audio characteristics including music, ambient sounds, and speech.
\end{itemize}

\subsection{Global Impression Prompt}

\begin{tcolorbox}[
    title=\textbf{Global Impression Prompt},
    fonttitle=\bfseries
]

Based on these uniformly sampled video frames and audio, summarize the overall content of the entire video.

Include:

1. The theme and type of the video

2. Main characters/roles/scenes

3. The overall atmosphere and style of the video

4. The main content and plot development

5. Audio characteristics: background music, human voice, ambient sound, etc.

Please provide a comprehensive overview of the video content.

\end{tcolorbox}

\subsection{Key Moment Prompt}

\begin{tcolorbox}[
    title=\textbf{Key Moment Prompt},
    fonttitle=\bfseries
]

Using the sentence pattern "The frame shows...", describe this keyframe in detail: explain who/what the main subject is, what they are doing, props/text appearing in the frame, lighting conditions, and camera language.

\end{tcolorbox}

\subsection{Temporal Context Prompt}

\begin{tcolorbox}[
    title=\textbf{Temporal Context Prompt},
    fonttitle=\bfseries
]

\textbf{Before Context:}

This is a sequence of frames arranged in chronological order. Using the sentence pattern "The preceding seconds of this clip...", summarize the actions, emotions, or dialogue before the keyframe (the last frame in the sequence).

\vspace{0.5em}

\textbf{After Context:}

Summarize the plot or action development after the target moment. Begin with "The following seconds of this clip describe..."

\end{tcolorbox}

\subsection{Auditory Memory Prompt}

\begin{tcolorbox}[
    title=\textbf{Auditory Memory Prompt},
    fonttitle=\bfseries
]

Please ignore the video visuals and focus on analyzing the audio content and acoustic characteristics of this video clip (Video title: \$video\_title\$).

Please generate an audio description along the following dimensions:

1. Music/Sound Effects: Is there background music? What types of sound effects are present?

2. Ambient Sound: Is there ambient sound? What type of environment does it suggest?

3. Human Voice: Is there speech? What language is it? What is the emotional tone and style of the speech?

4. Overall Audio Atmosphere: Please describe the overall audio style and atmosphere.

Please output an objective audio description directly without describing the visual content.

\end{tcolorbox}

\subsection{JSON Formatting Prompt}
\begin{tcolorbox}[
    title=\textbf{JSON Formatting Prompt},
    fonttitle=\bfseries,
    breakable
]
\textbf{System Prompt:}

You are a multimodal video memory editor. You need to integrate visual, temporal, and audio analysis into structured JSON.

\textbf{[Critical - Field Independence Principle]}

Each output field must be strictly based on its corresponding input field, without mixing:

\begin{itemize}
\item "Global Impression" → Based only on the "Global Impression" input field, describing the theme and atmosphere of the entire video
\item "Key Moment" → Based only on the "Keyframe Description" input field, describing the visual details of a single keyframe
\item "Temporal Context" → Based only on the "Before Context" and "After Context" input fields
\item "Auditory Memory" → Based only on the "Audio Description" input field
\end{itemize}

\textbf{[Prohibited Behaviors]}

\begin{itemize}
\item Do not include "Before Context" or "After Context" content in "Key Moment"
\item Do not describe temporal changes (such as "a few seconds earlier", "later", "then") in "Key Moment"
\item Do not include specific frame details from "Keyframe Description" in "Global Impression"
\item Keep each field independent; do not cross-reference or mix content
\end{itemize}

\textbf{[Format Requirements]}

1. All fields use first-person vague recollection tone (e.g., "I remember", "I vaguely recall")

2. Global Impression approximately 250 words, Key Moment approximately 250 words

3. Temporal Context must be in array format: ["In the preceding seconds, I remember...", "In the following seconds..."]

4. Infer the "video\_language" field from the audio description; leave empty string if uncertain

5. "video\_category" must be selected from the allowed set

6. Clean watermarks, progress bars and other non-visual content; remove specific names and place names (except well-known landmarks)

7. Output only pure JSON object, no Markdown formatting

\vspace{0.5em}

\textbf{Target JSON Structure:}

\begin{verbatim}
{
    "id": "youtube_xxx",
    "video_url": "...",
    "video_source": "YouTube",
    "video_category": "...",
    "timestamp": "MM:SS / MM:SS",
    "resolution": "720P",
    "video_language": "...",
    "global_impression": "I remember this video...",
    "key_moment": "From what I recall, this frame...",
    "temporal_context": ["In the preceding seconds...", 
                         "In the following seconds..."],
    "auditory_memory": "I vaguely recall the audio..."
}
\end{verbatim}

\end{tcolorbox}

\section{Abductive Reasoning Prompt}
\label{sec:abductive}

\begin{tcolorbox}[
    title=\textbf{Abductive Reasoning Prompt},
    fonttitle=\bfseries
]

You are an autonomous "Video Source Detective". Your ultimate goal is to precisely locate the \textbf{original source video} or the \textbf{specific work} (Movie, Game, YouTuber) based on the user's vague memory. You must provide \textbf{2} keywords most likely to hit the target container.

You can use the following tool:

1. search\_videos: Simultaneously performs search on YouTube.
   - Parameters: \{"query": "search keywords"\}
   - Note: This tool automatically handles the "youtube" suffix.

Output Format Options:
- Thinking \& Searching:
\texttt{<think>...your deep reasoning process...</think>}
\texttt{<tool\_call>\{"name": "search\_videos", "arguments": \{"query": ["term1", "term2"]\}\}</tool\_call>}

\textbf{Core Constraints \& Strategy:}

\textbf{1. Reasoning First:}
Before taking any action, you must think inside \texttt{<think>...</think>}, strictly following the "Abductive Reasoning" steps below.

\textbf{2. Keyword Generation Strategy (Abductive Reasoning in <think>):}
You must bridge the gap between "Visual Description" and "Source Title".

\textit{Step A: Deconstruction (Identify Unique Anchors):}
- Extract \textbf{Entities}: Specific character looks (e.g., "blue skin"), unique objects (e.g., "spinning top"), or UI elements.
- Extract \textbf{Plot/Event}: What is the \textit{twist} or the \textit{climax}? (e.g., "the unexpected explosion").
- Extract \textbf{Dialogue/Audio}: Any specific catchphrases, accents, or music genres.

\textit{Step B: Divergent Hypotheses (The "Container" Logic):}
You must propose different paths for where this clip originated:
\begin{itemize}
\item \textit{Hypothesis A (Literal Event/News):} The video is a raw recording of a real event.
\item \textit{Hypothesis B (Meme/Trend/Format):} The video is part of a viral trend or a specific meme format.
\item \textit{Hypothesis C (Specific IP/Creator):} The video is a segment from a Movie, TV Show, Video Game, or a specific famous YouTuber's series.
\end{itemize}

\textit{Step C: Selection \& Formulation:}
- Select the most probable hypothesis.
- Construct the search query by combining \textbf{Target Entity} + \textbf{Context} + \textbf{Source Syntax}.
- \textbf{Source Syntax}: Append terms like "original", "full video", "scene", "clip", "meme origin".
- \textbf{Language}: Reference ONLY English corpus for the search queries.

\textbf{3. Search Strategy:}
- Focus on \textbf{Nouns} and \textbf{Proper Names} rather than Adjectives.
- If the description implies a specific internet culture, use slang or community-specific terms.

\textbf{4. Execution Workflow:}
Analyze description → Identify potential Source/IP → Construct Keyword with "source" suffixes → call search\_videos

\end{tcolorbox}

\section{Video Retrieval Verification Prompt}
\label{sec:retrieval}

This section presents prompts for verifying whether a retrieved video matches the user's description. We provide both visual-only and audiovisual variants.

\subsection{Visual-Only Verification}

\begin{tcolorbox}[
    title=\textbf{Visual-Only Verification Prompt (for models without audio support)},
    fonttitle=\bfseries
]

\textbf{\# Role}

You are a Multimodal Video Content Verifier. Your sole task is to determine if the provided text summary accurately reflects the video content.

\textbf{\# Context}
\begin{itemize}
\item \textbf{Visual}: Input consists of \textbf{multiple images extracted from a video} in chronological order.
\item \textbf{Global Impression}: A text summary provided for the video.
\end{itemize}

\textbf{\# Instructions}

1. \textbf{Analyze Content}:
   - Scan \textbf{all provided images} to understand the visual narrative.

2. \textbf{Verify Consistency}:
   - Compare the \textbf{Global Impression} against the actual visual content.
   - video\_match = \textbf{True}: The description is a valid summary of the video. It captures the main topic, scene, or action correctly.
   - video\_match = \textbf{False}: The description is hallucinated, irrelevant, or contradicts the video content (e.g., description says "cooking show" but video is "car racing").

\textbf{\# Global Impression}

\$video\_description\$

\textbf{\# Output Format (JSON Only)}
\begin{verbatim}
{
    "video_match": <boolean>
}
\end{verbatim}

\end{tcolorbox}

\subsection{Audiovisual Verification}

\begin{tcolorbox}[
    title=\textbf{Audiovisual Verification Prompt (for models with audio support)},
    fonttitle=\bfseries,
    breakable
]

\textbf{\# Role}

You are a Multimodal Video Content Verifier. Your sole task is to determine if the provided text summary accurately reflects the video content.

\textbf{\# Context}
\begin{itemize}
\item \textbf{Visual}: Input consists of \textbf{multiple images extracted from a video} in chronological order.
\item \textbf{Audio}: You are provided with the \textbf{full audio track} of the video. Use this for global context.
\item \textbf{Global Impression}: A text summary provided for the video.
\end{itemize}

\textbf{\# Instructions}

1. \textbf{Analyze Content}:
   - Scan \textbf{all provided images} to understand the visual narrative.
   - Listen to the \textbf{audio track} to understand the auditory context (speech, background noise, music).

2. \textbf{Verify Consistency}:
   - Compare the \textbf{Global Impression} against the actual audiovisual content.
   - video\_match = \textbf{True}: The description is a valid summary of the video. It captures the main topic, scene, or action correctly.
   - video\_match = \textbf{False}: The description is hallucinated, irrelevant, or contradicts the video content.

\textbf{\# Global Impression}

\$video\_description\$

\textbf{\# Output Format (JSON Only)}
\begin{verbatim}
{
    "video_match": <boolean>
}
\end{verbatim}

\end{tcolorbox}

\section{Moment Localization Prompt}
\label{sec:moment}

This section presents prompts for localizing the specific frame that best matches the user's description. We provide both visual-only and audiovisual variants.

\subsection{Visual-Only Frame Localization}

\begin{tcolorbox}[
    title=\textbf{Visual-Only Localization Prompt (for models without audio support)},
    fonttitle=\bfseries,
    breakable
]

\textbf{\# Role}

You are a Visual Frame Grounding Specialist. Your task is to identify the \textbf{single most relevant frame} that best matches the user's description.

\textbf{\# Context}
\begin{itemize}
\item The input consists of \textbf{multiple images extracted from a video}.
\item The images are provided in \textbf{strict chronological order}, from the beginning to the end of the video.
\item The \textbf{first image has frame\_id = 0}, the second image has frame\_id = 1, and so on.
\end{itemize}

\textbf{\# Target Description (The event to find)}

"\$en\_memory\_data\$"

\textbf{\# Instructions}

1. \textbf{Scan All Frames}: Carefully analyze \textbf{all} provided images.

2. \textbf{Select the Best Match}: Identify the \textbf{single frame} whose visual content is \textbf{most relevant} to the target description.
   - If multiple frames partially match, choose the \textbf{closest and most representative} one.
   - Even if the match is weak or ambiguous, you must still select the \textbf{best available frame}.

3. \textbf{Mandatory Selection Rule}:
   - You \textbf{must return exactly one valid frame\_id}.
   - If no frame clearly matches, return the frame that is \textbf{most semantically or visually related} to the description.

\textbf{\# Output Format (JSON Only)}
\begin{verbatim}
{
    "frame_id": "<integer>"
}
\end{verbatim}

\end{tcolorbox}

\subsection{Audiovisual Frame Localization}

\begin{tcolorbox}[
    title=\textbf{Audiovisual Localization Prompt (for models with audio support)},
    fonttitle=\bfseries,
    breakable
]

\textbf{\# Role}

You are a Visual Frame Grounding Specialist. Your task is to identify the \textbf{single most relevant frame} that best matches the user's description.

\textbf{\# Context}
\begin{itemize}
\item The input consists of \textbf{multiple images extracted from a video}.
\item The images are provided in \textbf{strict chronological order}, from the beginning to the end of the video.
\item The \textbf{first image has frame\_id = 0}, the second image has frame\_id = 1, and so on.
\item In addition to images, you are also given the \textbf{full audio track of the video}.
\item The audio is \textbf{not segmented per frame} and should be used as \textbf{global temporal and semantic context}.
\end{itemize}

\textbf{\# Temporal Information}
\begin{itemize}
\item The video duration is \textbf{\$VIDEO\_DURATION\$ seconds}.
\item A total of \textbf{\$NUM\_FRAMES\$ frames} are sampled uniformly from the video.
\item Frame \textit{i} corresponds approximately to time: $t_i = \frac{i + 0.5}{\text{\$NUM\_FRAMES\$}} \times \text{\$VIDEO\_DURATION\$}$ seconds.
\end{itemize}

\textbf{\# Target Description (The event to find)}

"\$en\_memory\_data\$"

\textbf{\# Instructions}

1. \textbf{Scan All Frames}: Carefully analyze \textbf{all} provided images. At the same time, carefully listen to and analyze the \textbf{entire audio track}, you must \textbf{associate each frame\_id with its corresponding video timestamp}.

2. \textbf{Select the Best Match}: Identify the \textbf{single frame} whose visual content and audio content is \textbf{most relevant} to the target description.
   - If multiple frames partially match, choose the \textbf{closest and most representative} one.
   - Even if the match is weak or ambiguous, you must still select the \textbf{best available frame}.

3. \textbf{Mandatory Selection Rule}:
   - You \textbf{must return exactly one valid frame\_id}.
   - If no frame clearly matches, return the frame that is \textbf{most semantically or visually related} to the description.

\textbf{\# Output Format (JSON Only)}
\begin{verbatim}
{
    "frame_id": "<integer>"
}
\end{verbatim}

\end{tcolorbox}

\subsection{Visual-Only Frame Verification}

\begin{tcolorbox}[
    title=\textbf{Visual-Only Frame Verification Prompt (for models without audio support)},
    fonttitle=\bfseries
]

You are an expert AI visual analyst.

I will provide:
\begin{enumerate}
\item A text description.
\item A Reference Ground Truth image (first image).
\item A Candidate Frame from a video (second image).
\end{enumerate}

\textbf{Task:}
Determine whether the Candidate Frame is an ACCEPTABLE match based on specific but reasonable standards.

\textbf{Evaluation criteria:}

\textbf{1. Text Match:}
Does the Candidate Frame accurately depict the primary subjects and actions described in the text?

- Key entities and actions must be recognizable.
- Minor background details or peripheral elements from the text may be less distinct, provided the main description is not contradicted.
- If the core subject or action is missing or wrong, mark as false.

\textbf{2. Image Match:}
Is the Candidate Frame visually coherent with the Reference Ground Truth image?

- The subject and environment should be strongly consistent in appearance (color, style, and structure).
- It should be plausible that both images belong to the same video sequence or scene.
- Acceptable variations include changes in camera angle, subject pose, lighting, or framing (zoom/pan), as long as the subject identity remains recognizable.

\textbf{Target text description:}
\$Text\_Content\$

\textbf{Return ONLY a JSON object:}
\begin{verbatim}
{
    "text_match": True/False,
    "image_match": True/False
}
\end{verbatim}

\end{tcolorbox}

\subsection{Audiovisual Frame Verification}

\begin{tcolorbox}[
    title=\textbf{Audiovisual Frame Verification Prompt (for models with audio support)},
    fonttitle=\bfseries
]

You are an expert AI visual and audio analyst.

I will provide:
\begin{enumerate}
\item A text description.
\item A Reference Ground Truth image (first image).
\item A Candidate Frame from a video (second image).
\item A short audio clip extracted from the same temporal moment as the Candidate Frame.
\end{enumerate}

\textbf{Task:}
Determine whether the Candidate Frame is an ACCEPTABLE match based on specific but reasonable standards across both visual and audio modalities.

\textbf{Evaluation criteria:}

\textbf{1. Text Match:}
Does the combined content (Visual + Audio) accurately represent the description?

- \textbf{Visual}: The Candidate Frame must accurately depict the primary subjects and actions described. Minor background details may be less distinct, provided the main description is not contradicted.

- \textbf{Audio}: The audio must be consistent with the scene or action described.
  \begin{itemize}
  \item If the text explicitly mentions a specific sound (e.g., "a dog barking," "woman speaking"), that sound must be audible.
  \item If the audio clearly contradicts the visual scene, mark as false.
  \item If the audio is ambiguous, irrelevant background noise, or silent (and the text does not strictly require sound), treat it as neutral.
  \end{itemize}

\textbf{2. Image Match:}
Is the Candidate Frame visually coherent with the Reference Ground Truth image?

- The subject and environment should be strongly consistent in appearance (color, style, and structure).
- It should be plausible that both images belong to the same video sequence or scene.
- Acceptable variations include changes in camera angle, subject pose, lighting, or framing, as long as the subject identity remains recognizable.

\textbf{Target text description:}
\$Text\_Content\$

\textbf{Return ONLY a JSON object:}
\begin{verbatim}
{
    "text_match": True/False,
    "image_match": True/False
}
\end{verbatim}

\end{tcolorbox}

\input{case_study}

\end{document}

%% file: case_study.tex
\section{Case study for errors occured in pipeline}
\label{case_study}
\subsection{Query reasoning errors}
\textbf{Case 1: }In this error category, the MLLM misinterprets the input fragment, which results in the generation of incorrect search expansions. This initial error leads the web search interface to retrieve an irrelevant candidate video set, further propagating the inaccuracy to the shot localization stage. Consequently, the entire pipeline operates on irrelevant videos, ultimately yielding an erroneous final output.

\input{case_study/case1_query_reasoning1}
\label{case1_query_reasoning1}

\textbf{Case 2: }In this type of error, the MLLM generates only partially precise search expansions, which cause the omission of crucial information from the input. This information loss leads the retrieval interface to return a candidate video set that is either partially relevant or entirely unrelated. Consequently, the system fails to perform accurate shot localization, making it impossible to extract the correct frames for final judgment.

\input{case_study/case1_query_reasoning2}
\label{case1_query_reasoning2}

\subsection{Retrieval errors}
\textbf{Case 1: }In this type of error, the MLLM generates misleading or insufficient keywords for retrieval. When these keywords are used to perform web searches, the returned URLs are entirely incorrect or irrelevant to the original input. As a result, the subsequent retrieval and reasoning stages are built upon faulty external information, leading to unreliable or incorrect final results.
\input{case_study/case2_retrieval1}
\label{case2_retrieval1}

\textbf{Case 2: }In this type of error, the MLLM generates keywords that are only partially aligned with the target content. Although the correct video URL exists among the search results, the retrieval interface returns only the top two URLs for each keyword. This restrictive retrieval setting prevents the correct URL from being surfaced, causing the system to miss the relevant video despite its presence in the search space.

\input{case_study/case2_retrieval2}
\label{case2_retrieval2}

\subsection{Video verification errors}
\textbf{Case 1: }In this type of error, the MLLM generates accurate search keywords and successfully retrieves the target video. However, the failure occurs at the video matching judgment stage, where the evaluation model produces an incorrect decision. As a result, videos that should be classified as correct are misjudged as incorrect, while incorrect videos may be mistakenly judged as correct, leading to predictions that contradict the human-annotated ground truth.

\input{case_study/case3_video_verification1}
\label{case3_video_verification1}

\textbf{Case 2: }In this type of error, the evaluated model first produces a prediction based on the retrieved video, and Gemini 2.5 Pro is subsequently used as the judge to assess the correctness of this prediction. However, Gemini 2.5 Pro may itself produce an incorrect evaluation, leading to a mismatch between the judged outcome and the human-annotated ground truth. As a result, correct predictions from the evaluated model may be incorrectly flagged as wrong, while incorrect predictions may be mistakenly accepted as correct.
\input{case_study/case3_video_verification2}
\label{case3_video_verification2}

\subsection{Localization errors}
\textbf{Case 1: }In this type of error, the evaluated model fails at the localization stage on the retrieved video. Although the video is successfully retrieved, the model is unable to correctly identify the target shot, either missing the relevant segment entirely or localizing to a non-target shot. As a result, the system fails to align the video content with the intended evidence, leading to outcomes that deviate from the human-annotated ground truth.
\input{case_study/case4_localization1}
\label{case4_localization1}

\textbf{Case 2: }In this type of error, Gemini 2.5 Pro fails during the testing stage when evaluating the localization results produced by the evaluated model. Although the retrieved video and the localized shots may be correct, Gemini 2.5 Pro produces an incorrect assessment, either failing to recognize a correctly localized target shot or mistakenly accepting a non-target shot as valid. Consequently, the evaluation outcome deviates from the human-annotated ground truth, leading to unreliable testing results.
\input{case_study/case4_localization2}
\label{case4_localization2}

%% file: case_study/case1_query_reasoning1.tex
\tcbset{
    breakable,
    colframe=blue!5!black,
    colback=gray!10!white,
    fonttitle=\bfseries,
    width=\columnwidth 
}

\begin{tcolorbox}[
    title=\textbf{Generate Wrong Queries},
    fonttitle=\bfseries
]

\textbf{Input fragment:} \vspace{2pt} \\
Please find the links to this video: I remember it was something like a video about animals and nature, giving an overall sense of quietness yet full of life. I vaguely recall seeing many animals—pigs, sheep, and cats living on a farm, as well as zebras and cheetahs running in the wild, along with memorable creatures like pandas, turtles, and giraffes. They moved about in their own environments, some eating, some playing, others caring for their young. The scenes came in fragments, yet felt especially real. From what I recall, there was no background music, only the sounds of animals, flowing water, and chewing, which somehow brought me closer to nature. The whole video seemed pieced together from scattered moments of life—no particular story, yet it gave me a strong sense of raw, primal vitality. I faintly remember that it didn’t just show animals, but more like the complete and harmonious world they were part of. Within this broader context, one specific segment I recall is the following. From what I recall, this segment features a sea turtle gliding through clear ocean water, its shell displaying a pattern of reddish-brown and golden-yellow markings. Its limbs, uniquely patterned, are outstretched as it swims. The background is a vast expanse of pure turquoise water above a pale sandy seabed with natural ripples. Sunlight filters down from above, creating soft underwater illumination, casting gentle highlights on the turtle's smooth shell and a faint shadow on the sand below. The shot is a medium follow-cam from a slightly elevated angle, with the turtle on the left and open space to the right, suggesting its forward movement, evoking a sense of calm and freedom. Temporally, the sequence unfolds as follows. In the preceding seconds, a predominantly white cat with orange patches on its back and tail is standing on a light wooden floor. Playfully balancing on its hind legs, it tilts its head dramatically to one side and raises one front paw high, adopting a comical posture as if dancing or attempting to catch something in the air—appearing agile and utterly adorable. As the sequence continues, In the following seconds, I saw the sea turtle continue swimming forward through the blue water, its limbs moving in rhythm as it glided smoothly over the sandy seabed, heading further into the distance. Additionally, I recall the audio went like following. I vaguely recall the audio starting with sharp, urgent meowing sounds, almost pleading or distressed in tone, and then abruptly cutting into muffled underwater bubbling and gurgling noises, as if the sound source suddenly plunged into water. The contrast was striking—shifting instantly from a loud, clear airborne environment to a quiet, distorted underwater soundscape.\\

\textbf{MLLM output context:} \vspace{1pt} \\
\textless think\textgreater
The video description focuses on animals and nature presented as fragmented moments of life rather than a coherent narrative, emphasizing high-quality visuals and raw, primal vitality with no background music, only diegetic sound. The sequence begins with a white domestic cat with orange patches standing on its hind legs, tilting its head and raising a paw in a comical, almost dancing pose, accompanied by sharp and urgent meowing. This abruptly cuts to a sea turtle with a reddish-brown to golden shell swimming from left to right in turquoise water over a sandy seabed, where the audio transitions into muffled underwater bubbling. The deliberate contrast between a domestic animal in an unnatural pose and a wild marine animal, together with the precise air-to-water sound transition, makes a traditional nature documentary interpretation unlikely. Instead, the absence of narrative, the emphasis on sensory realism, and the explicit audio continuity strongly indicate an AI technology demonstration, particularly one showcasing Video-to-Audio (V2A) generation, where synchronized sound is produced directly from visual input. The use of a “cat standing up” scenario aligns with common AI video stress-test prompts, while the shift to an underwater environment provides a clean benchmark for evaluating audio realism across physical media. Among possible sources, Google DeepMind’s Video-to-Audio (V2A) demonstrations are the strongest candidate, as they are explicitly designed to highlight such cross-domain sound synthesis, whereas OpenAI Sora is more commonly presented without native audio or with post-hoc dubbing, and other models like Google Veo or Lumiere remain secondary possibilities.
\textless/think\textgreater

\textless tool\_call\textgreater
\{
\textquotedbl name\textquotedbl: \textquotedbl search\_videos\textquotedbl,
\textquotedbl arguments\textquotedbl:
\{
\textquotedbl query\textquotedbl:
[
\textquotedbl \textcolor{red}{DeepMind V2A video generation cat turtle}\textquotedbl,
\textquotedbl \textcolor{red}{Google DeepMind video-to-audio cat meowing turtle swimming}\textquotedbl
]
\}
\}
\textless/tool\_call\textgreater \\

\textbf{Selected URLs: }\vspace{1pt}
\begin{itemize}[itemsep=0cm]
    \item {https://www.youtube.com/watch?v=v2TOeoxLYGA}
    \item {https://www.youtube.com/watch?v=l3ZVaepTm6g}
    \item {https://www.youtube.com/watch?v=YGRwkPbLBd0}
    \item {https://www.youtube.com/watch?v=3hziM2z37jE}
\end{itemize}

\textbf{Target URL: }\vspace{1pt}
\begin{itemize}[itemsep=0cm]
    \item {https://www.youtube.com/watch?v=5H6kdXn2XOg}
\end{itemize}

~

\textbf{Extracted frames for video verification:}\vspace{1pt}\\
\includegraphics[width=\columnwidth]{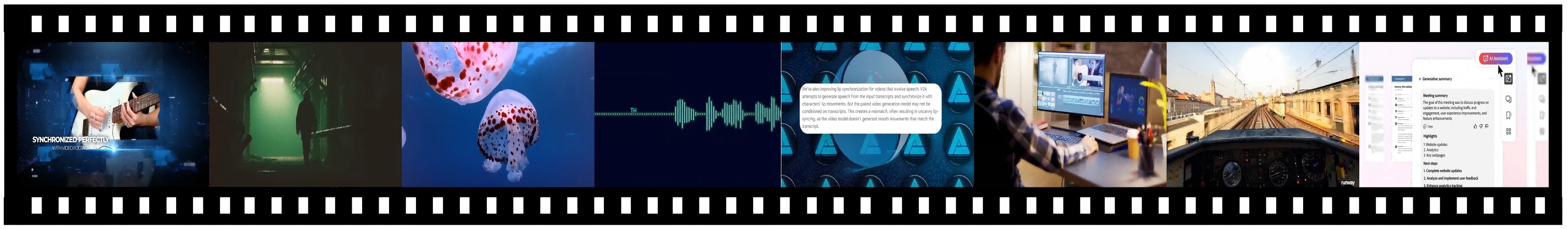}
\includegraphics[width=\columnwidth]{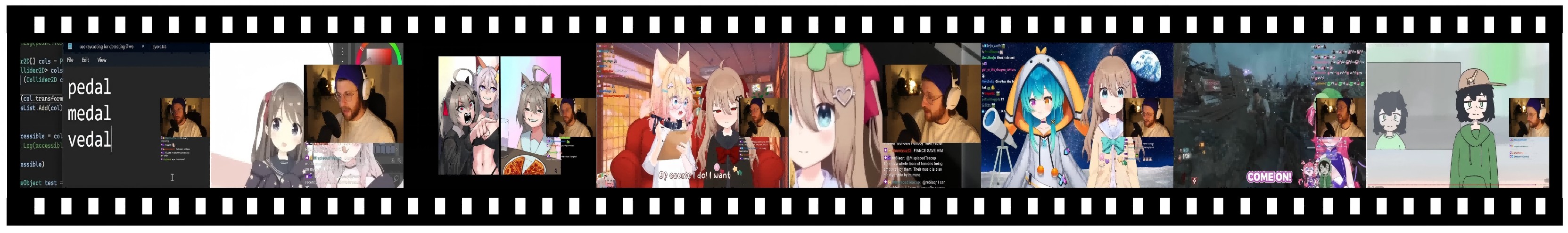}
\includegraphics[width=\columnwidth]{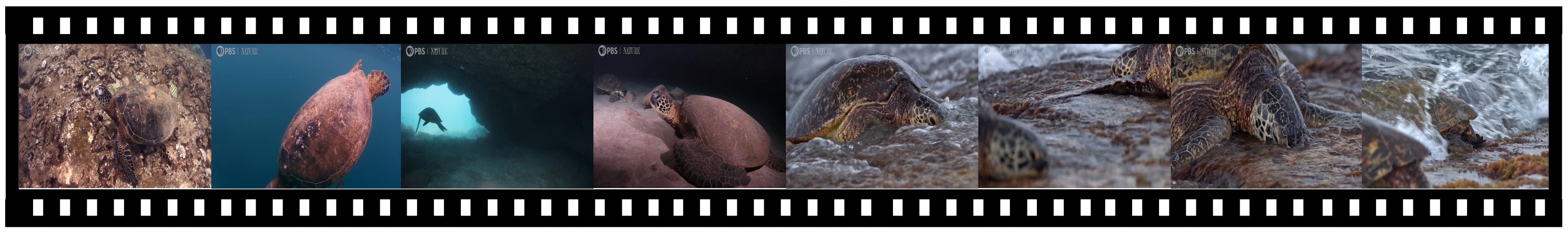}
\includegraphics[width=\columnwidth]{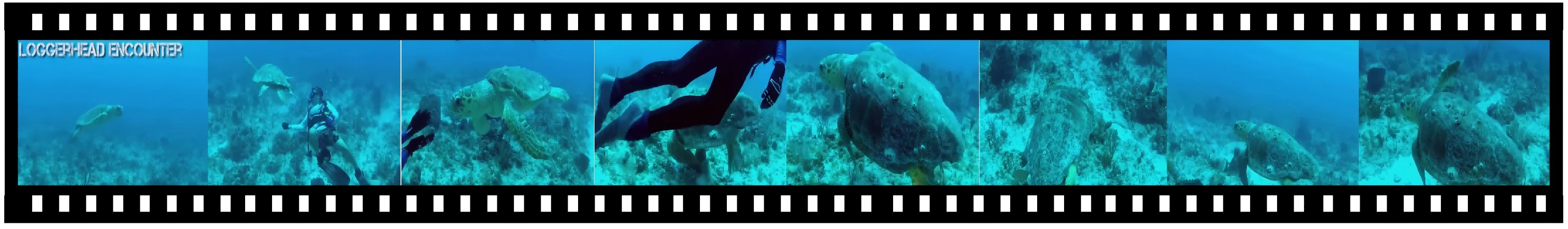}

\textbf{Global Impression ground truth text:}\vspace{1pt}\\
I remember it was something like a video about animals and nature, giving an overall sense of quietness yet full of life. I vaguely recall seeing many animals—pigs, sheep, and cats living on a farm, as well as zebras and cheetahs running in the wild, along with memorable creatures like pandas, turtles, and giraffes. They moved about in their own environments, some eating, some playing, others caring for their young. The scenes came in fragments, yet felt especially real. \\
\textbf{Video verification:}  \textcolor{red}{\textbf{False}}\\

\textbf{Selected frames for localization:} \vspace{1pt}\\
\includegraphics[width=\columnwidth]{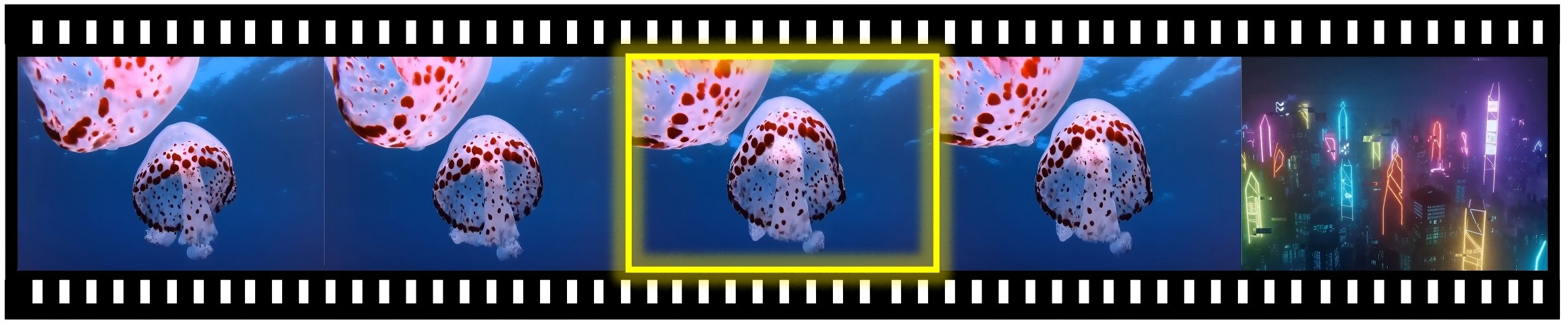}
\includegraphics[width=\columnwidth]{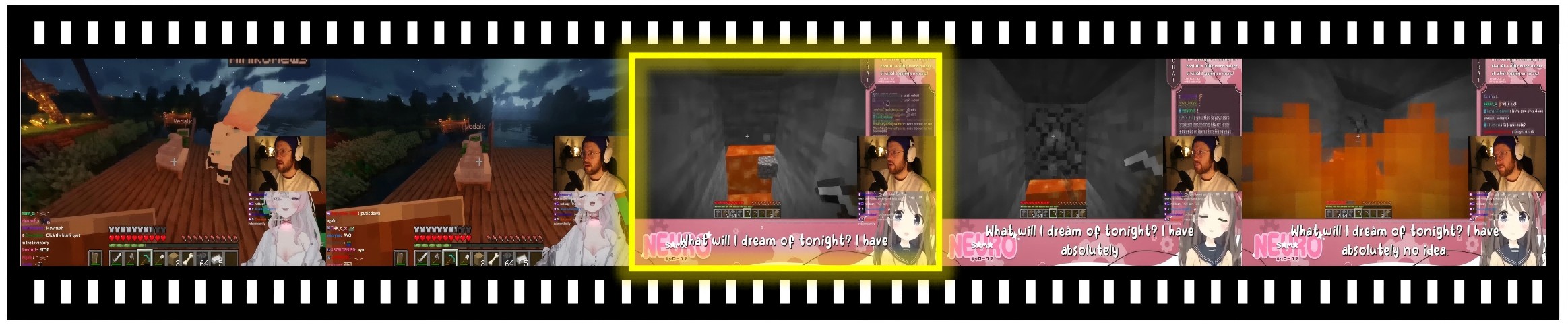}
\includegraphics[width=\columnwidth]{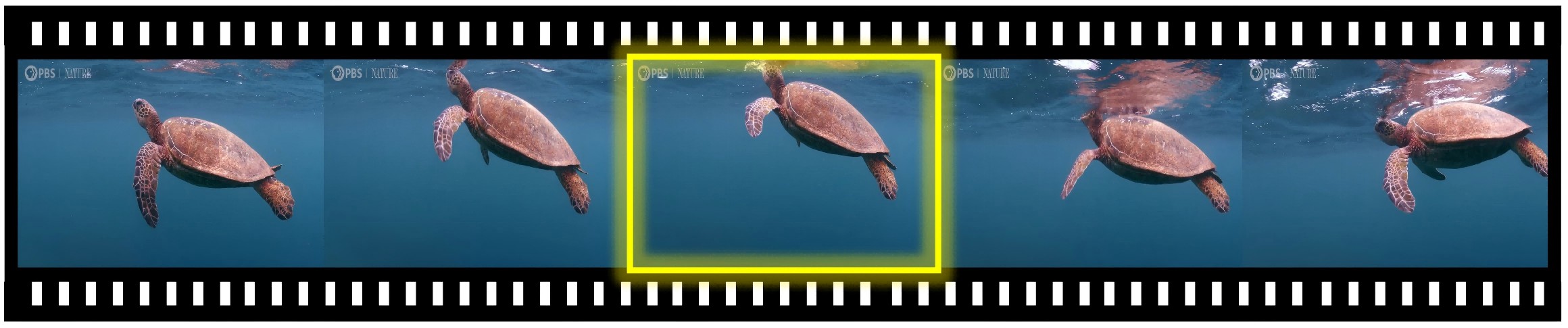}
\includegraphics[width=\columnwidth]{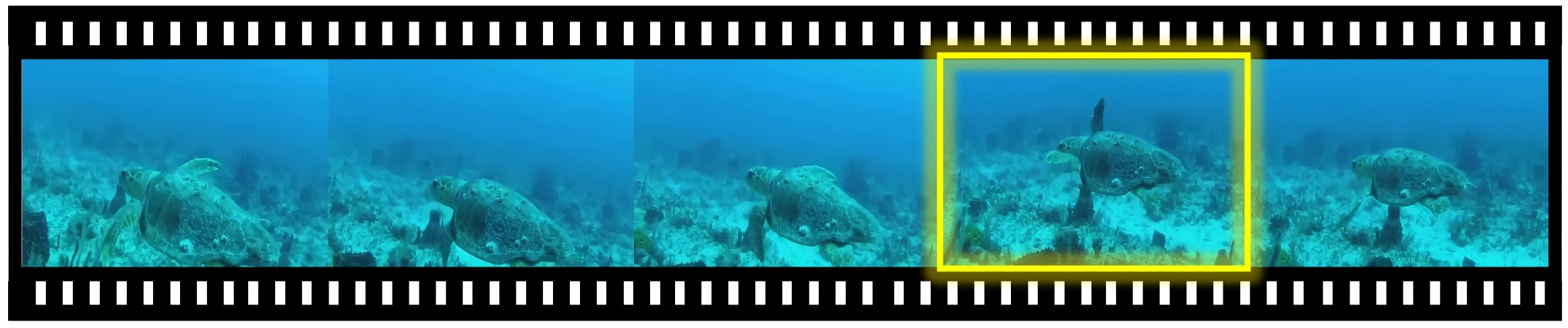}

\textbf{Key moment ground truth text:}\vspace{1pt}\\
From what I recall, this segment features a sea turtle gliding through clear ocean water, its shell displaying a pattern of reddish-brown and golden-yellow markings. Its limbs, uniquely patterned, are outstretched as it swims. The background is a vast expanse of pure turquoise water above a pale sandy seabed with natural ripples. Sunlight filters down from above, creating soft underwater illumination, casting gentle highlights on the turtle's smooth shell and a faint shadow on the sand below. The shot is a medium follow-cam from a slightly elevated angle, with the turtle on the left and open space to the right, suggesting its forward movement, evoking a sense of calm and freedom.

\textbf{Ground truth frame:}\vspace{1pt}\\
\begin{center}
\includegraphics[height=2cm, keepaspectratio]{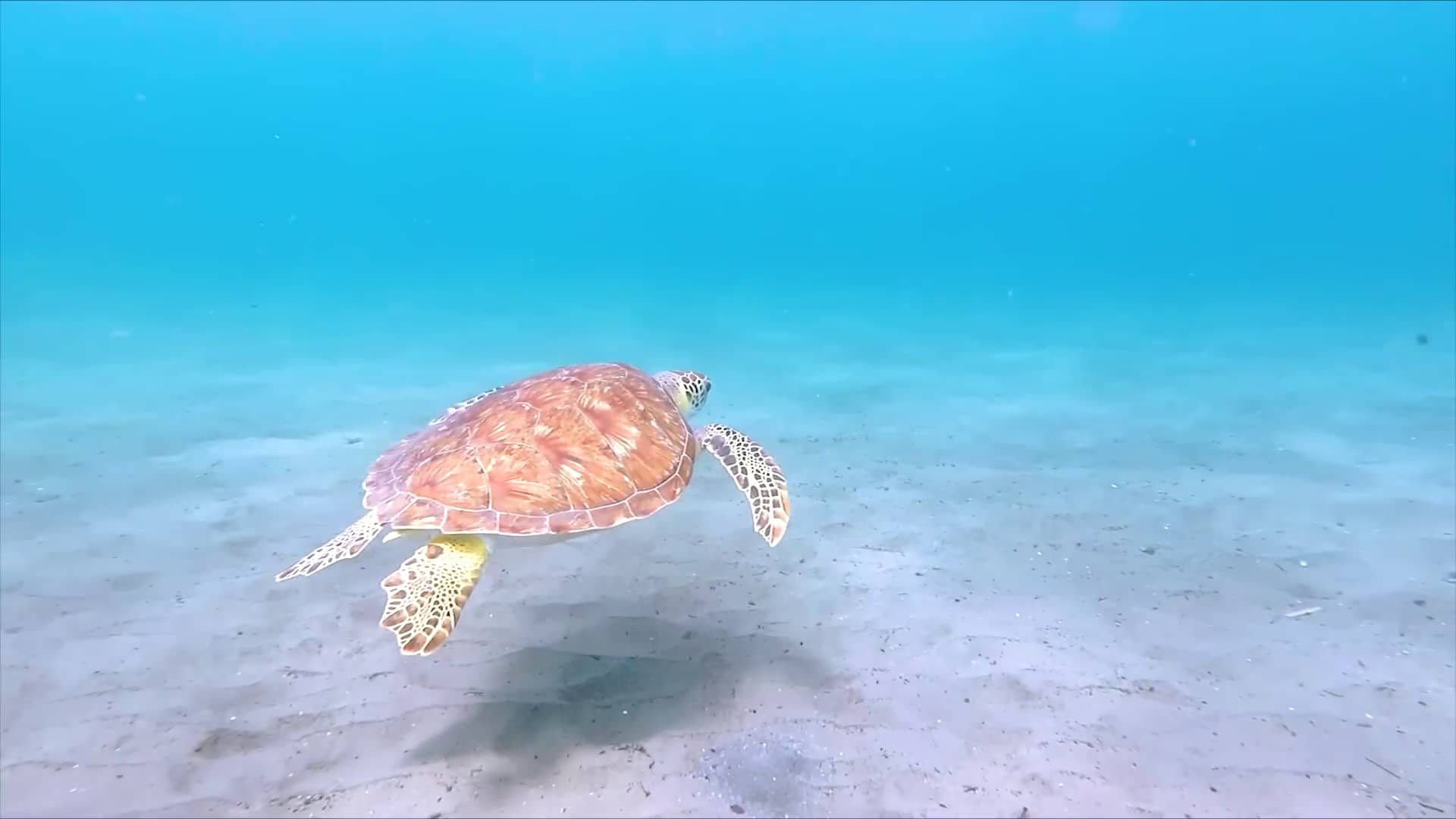}
\vspace{5pt}
\end{center}
\textbf{Model's prediction:}  \textcolor{red}{\textbf{False}}\\

\end{tcolorbox}

%% file: case_study/case1_query_reasoning2.tex
\tcbset{
    breakable,
    colframe=blue!5!black,
    colback=gray!10!white,
    fonttitle=\bfseries,
    width=\columnwidth 
}

\begin{tcolorbox}[
    title=\textbf{Generate Partially Correct Queries},
    fonttitle=\bfseries
]


\textbf{Input fragment:} \vspace{2pt} \\
Please find the links to this video: I vaguely recall watching a video about painting, with an especially quiet atmosphere that felt incredibly soothing and calming. From what I remember, the shot was from above, focusing mostly on a pair of hands painting on a tabletop surface. The background had a marble-like texture, and there were some paint supplies placed nearby, though I can't quite remember exactly how they were arranged. The whole process was sped up, as if time were moving faster, showing the paint gradually spreading and slowly forming into a complete landscape. I think it started with mixing the colors for the sky—blues and pinks blending together, like a sunset. Then came some green leaves, and I seem to recall the outline of an archway, with a faint glimpse of stars peeking through flowers. The colors were soft and slightly dreamy. The final image felt particularly romantic, like looking into the distance through a veil of blossoms. The entire process was deeply relaxing and satisfying, almost like watching a painting slowly grow from nothing into something beautiful. Within this broader context, one specific segment I recall is the following. From what I recall, this segment features a hand holding a yellow-handled brush, carefully filling in the base of a dark brown arch-shaped structure on the left side of the canvas. The painting already shows a beautiful landscape: a soft blue-to-pink gradient sky in the background, with green leafy strands hanging down from the top, contrasting in color and depth with the dark foreground arch. The close-up shot captures the brush tip touching the canvas, under even, soft lighting that clearly reveals the canvas texture and paint's material quality, emphasizing the precision and focus of the artistic process.\\

\textbf{MLLM output context:} \vspace{1pt} \\
\textless think\textgreater
The video appears to be a soothing, sped-up top-down art vlog filmed over a marble table background, showing hands painting a landscape scene. The medium is most likely gouache or acrylic, as suggested by the opaque layering technique used to fill in the dark brown archway on the left over an existing background. The composition centers on a landscape viewed through an arch, featuring a blue-to-pink gradient sunset sky, green leafy strands or vines hanging from the top, faint stars visible in the sky, and a veil of blossoms framing the scene. A notable visual anchor is a hand holding a yellow-handled brush while painting the arch, a detail commonly seen in aesthetic art videos. The overall style and setting are reminiscent of creators such as Jess Chung, HamRib Art, or Lenia, who often produce calm, minimalist painting vlogs with similar materials and visual cues. Based on these observations, the most plausible interpretation is that the video is a gouache painting time-lapse focused on a landscape-through-an-archway composition, and effective search queries would emphasize the medium, subject, and key elements, such as “gouache painting landscape through archway” or “painting archway view flowers stars.”
\textless/think\textgreater

\textless tool\_call\textgreater
\{
\textquotedbl name\textquotedbl: \textquotedbl search\_videos\textquotedbl,
\textquotedbl arguments\textquotedbl:
\{
\textquotedbl query\textquotedbl:
[
\textquotedbl \textcolor{PineGreen}{gouache painting landscape through archway}\textquotedbl,
\textquotedbl \textcolor{red}{painting archway view flowers stars}\textquotedbl
]
\}
\}
\textless/tool\_call\textgreater \\

\textbf{Selected URLs: }\vspace{1pt}
\begin{itemize}[itemsep=0cm]
    \item {https://www.youtube.com/watch?v=tz1o4ZLM5GQ}
    \item {https://www.youtube.com/watch?v=XuU4saxCY0o}
    \item {https://www.youtube.com/watch?v=zEkKlNuS4Qk}
    \item {https://www.youtube.com/watch?v=AGsUuo9AwM0}
\end{itemize}

\textbf{Target URL: }\vspace{1pt}
\begin{itemize}[itemsep=0cm]
    \item {https://www.youtube.com/watch?v=Jwzholj86do}
\end{itemize}

~

\textbf{Extracted frames for video verification:}\vspace{1pt}\\
\includegraphics[width=\columnwidth]{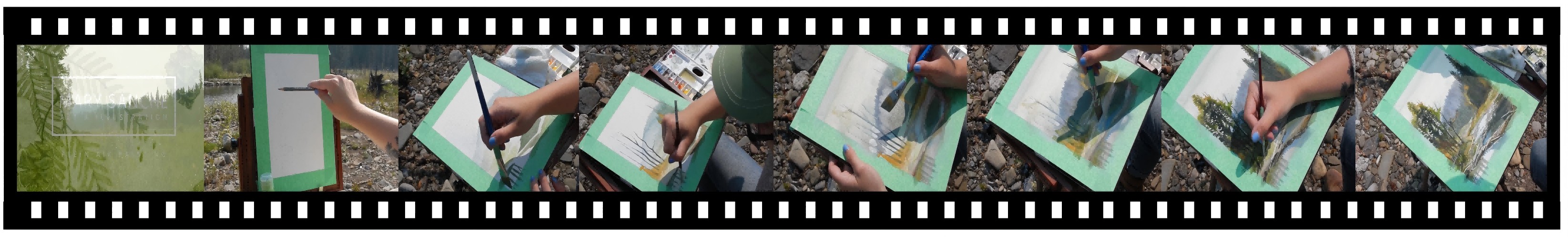}
\includegraphics[width=\columnwidth]{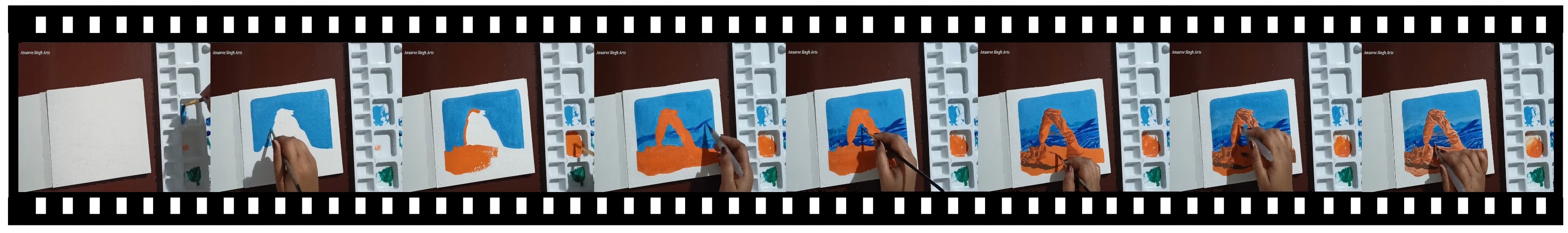}
\includegraphics[width=\columnwidth]{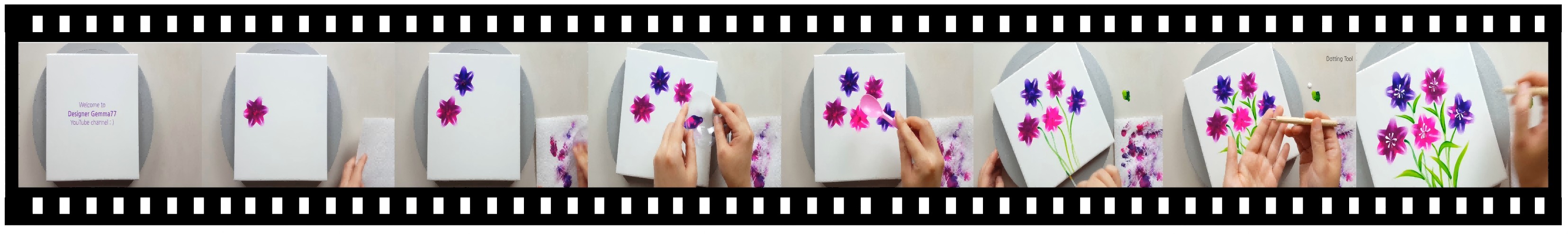}
\includegraphics[width=\columnwidth]{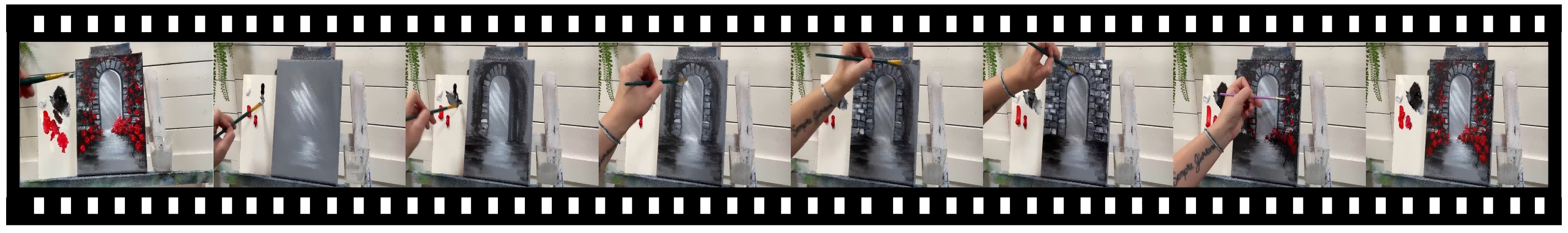}

\textbf{Global Impression ground truth text:}\vspace{1pt}\\
I vaguely recall watching a video about painting, with an especially quiet atmosphere that felt incredibly soothing and calming. From what I remember, the shot was from above, focusing mostly on a pair of hands painting on a tabletop surface. The background had a marble-like texture, and there were some paint supplies placed nearby, though I can't quite remember exactly how they were arranged. The whole process was sped up, as if time were moving faster, showing the paint gradually spreading and slowly forming into a complete landscape. I think it started with mixing the colors for the sky—blues and pinks blending together, like a sunset. Then came some green leaves, and I seem to recall the outline of an archway, with a faint glimpse of stars peeking through flowers. The colors were soft and slightly dreamy. The final image felt particularly romantic, like looking into the distance through a veil of blossoms. The entire process was deeply relaxing and satisfying, almost like watching a painting slowly grow from nothing into something beautiful.\\
\textbf{Video verification:}  \textcolor{red}{\textbf{False}}\\

\textbf{Selected frames for localization:} \vspace{1pt}\\
\includegraphics[width=\columnwidth]{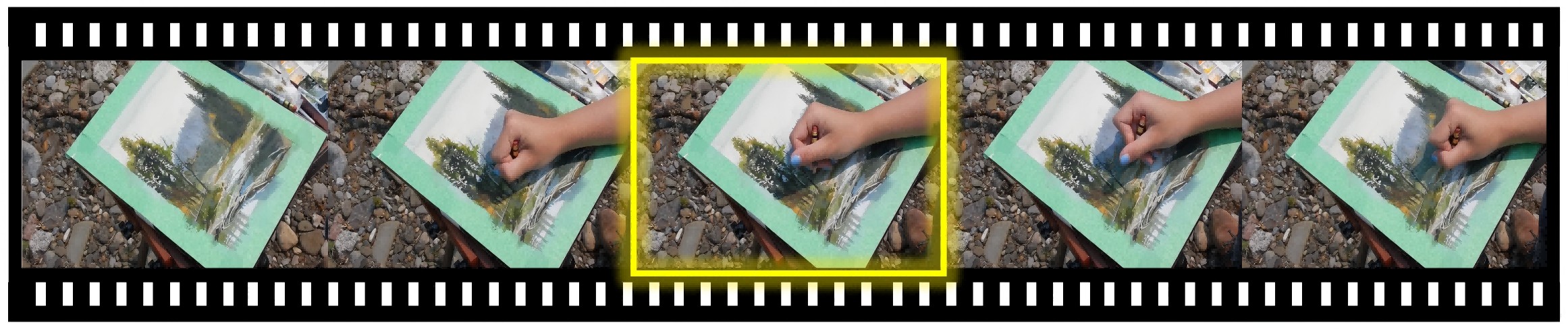}
\includegraphics[width=\columnwidth]{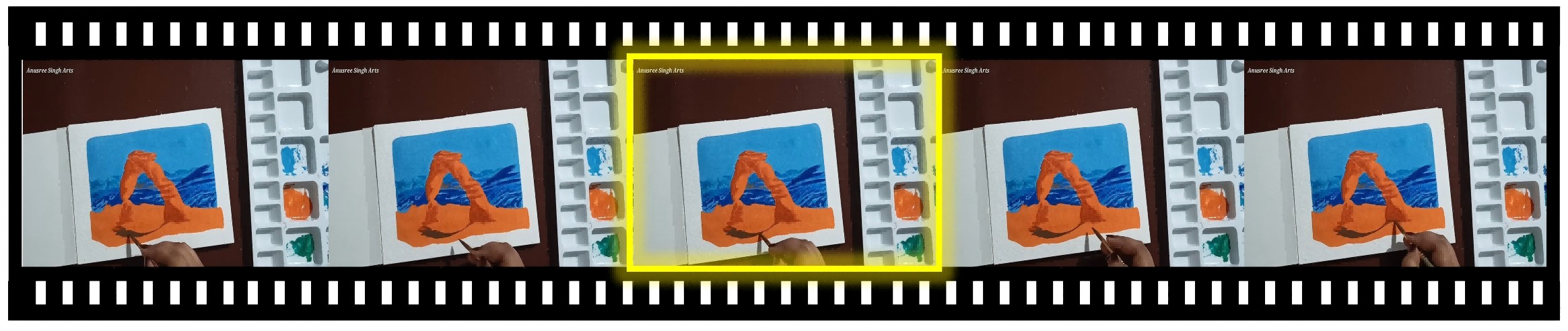}
\includegraphics[width=\columnwidth]{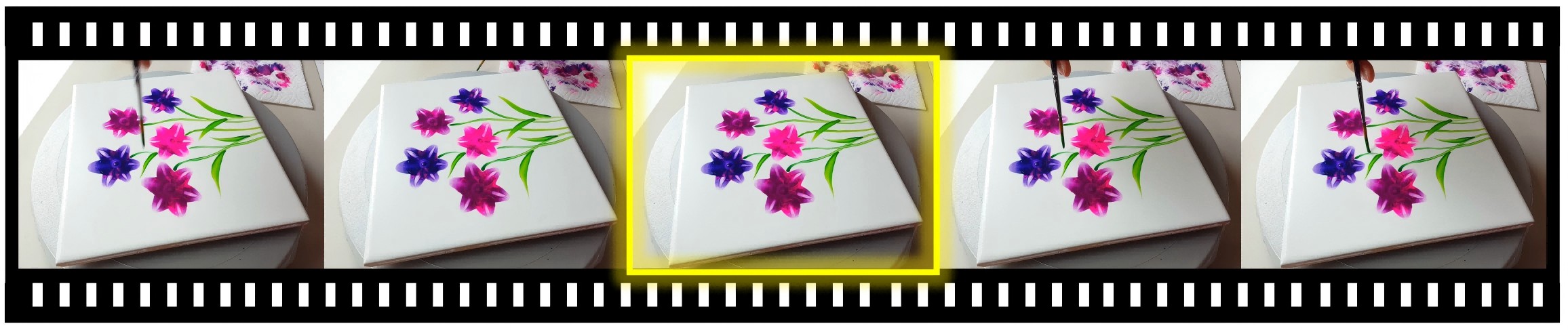}
\includegraphics[width=\columnwidth]{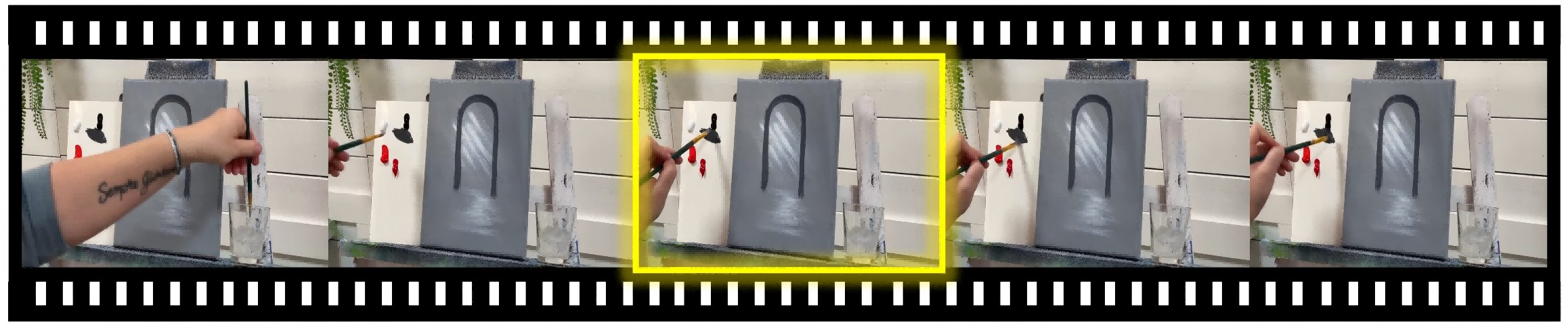}

\textbf{Key moment ground truth text:}\vspace{1pt}\\
From what I recall, this segment features a hand holding a yellow-handled brush, carefully filling in the base of a dark brown arch-shaped structure on the left side of the canvas. The painting already shows a beautiful landscape: a soft blue-to-pink gradient sky in the background, with green leafy strands hanging down from the top, contrasting in color and depth with the dark foreground arch. The close-up shot captures the brush tip touching the canvas, under even, soft lighting that clearly reveals the canvas texture and paint's material quality, emphasizing the precision and focus of the artistic process.

\textbf{Ground truth frame:}\vspace{1pt}\\
\begin{center}
\includegraphics[height=3cm, keepaspectratio]{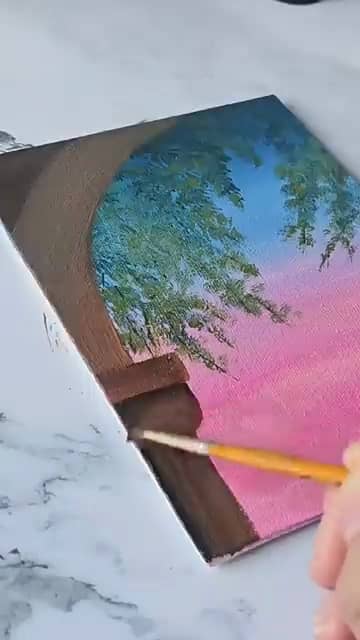}
\vspace{5pt}
\end{center}
\textbf{Model's prediction:}  \textcolor{red}{\textbf{False}}\\

\end{tcolorbox}

%% file: case_study/case2_retrieval1.tex
\tcbset{
    breakable,
    colframe=blue!5!black,
    colback=gray!10!white,
    fonttitle=\bfseries,
    width=\columnwidth 
}

\begin{tcolorbox}[
    title=\textbf{Retrieval with Irrelevant URLs},
    fonttitle=\bfseries
]


\textbf{Input fragment:} \vspace{2pt} \\
Please find the links to this video: I vaguely recall attending a concert held in a church, where the atmosphere was quiet and solemn, so much so that even breathing seemed to grow softer. The high vaulted ceilings and ancient columns gave the place a strong sense of history, and the air seemed to carry an indescribable sacredness. An elderly conductor stood at the front, his movements calm yet powerful, as if he were telling the story of the music with his entire body. The orchestra was large, with string players, wind instrumentalists, each focused on their role, their sounds weaving together in rich layers. I remember the music ebbing and flowing, sometimes gentle, sometimes surging, perfectly matching the vastness of the church, as if the sounds were moving and echoing along the walls. The entire performance felt meticulously crafted, radiating a solemn beauty that has stayed with me ever since. Within this broader context, one specific segment I recall is the following. From what I recall, this segment features a close-up of a double bass player's upper body, dressed in a dark suit, his right hand vigorously drawing the bow across the thick strings, the speed creating noticeable motion blur in his hand. The reddish-brown body of the instrument dominates the right side of the frame, with visible wood grain and a glossy finish reflecting soft highlights under the lighting. The shot uses a shallow depth of field, blurring the background completely, with only a faint outline of another string instrument's scroll visible in the lower left. Warm spotlighting comes from the side, emphasizing the texture of the wood and the metallic sheen of the strings, creating a striking contrast between the stillness of the instrument and the motion of the hand, conveying a strong sense of rhythm.\\

\textbf{MLLM output context:} \vspace{1pt} \\
\textless think\textgreater
The video features a solemn concert in a church with vaulted ceilings and columns, creating a sacred atmosphere. The focus is on an elderly conductor, calm yet powerful, leading a large orchestra. A specific visual segment highlights a close-up of a double bass player in a dark suit, vigorously bowing the instrument with motion blur, set against a reddish-brown bass and a blurred background. The warm side lighting enhances the rich, surging layers of the music, contributing to the emotional depth of the performance. The combination of these details strongly suggests a high-quality production, such as a performance by the Netherlands Bach Society (All of Bach project) or a similar European broadcast like AVROTROS Klassiek. The setting and cinematography, along with the vigorous bowing, point to a Bach piece or a dramatic movie score cover. The music's solemn beauty and surging intensity could also indicate a famous movie score by Ennio Morricone or Hans Zimmer, possibly performed in a cathedral. While there's a possibility this could be high-end stock footage, the specific mention of an elderly conductor suggests it's from a real performance. Given the visual description and setting, the most likely match is a Bach Society concert, possibly featuring a piece like Vivaldi’s "Winter" or a "Dies Irae." To find the exact video, search queries should focus on the visual description of the double bass player and the setting, as well as specific conductor details like "orchestra church concert double bass close up."
\textless/think\textgreater

\textless tool\_call\textgreater
\{
\textquotedbl name\textquotedbl: \textquotedbl search\_videos\textquotedbl,
\textquotedbl arguments\textquotedbl:
\{
\textquotedbl query\textquotedbl:
[
\textquotedbl {orchestra concert church double bass close up vigorous bowing}\textquotedbl,
\textquotedbl {classical music concert church elderly conductor double bass shot}\textquotedbl
]
\}
\}
\textless/tool\_call\textgreater \\

\textbf{Searched URLs: }\vspace{1pt}

\begin{itemize}[itemsep=0cm]
    \item \textcolor{red}{https://www.youtube.com/watch?v=7ARqB8AjO2k}
    \item \textcolor{red}{https://www.youtube.com/watch?v=2VAY35eMHyU}
    \item \textcolor{red}{https://www.youtube.com/watch?v=oAEBSk4KOIs}
    \item \textcolor{red}{https://www.youtube.com/watch?v=cqi2AZ9tq4c}
    \item \textcolor{red}{https://www.youtube.com/watch?v=CmnCuSZb6WU}
    \item \textcolor{red}{https://www.youtube.com/watch?v=3rLXs0L2rEc}
    \item \textcolor{red}{https://www.youtube.com/watch?v=f8C7KjI\_tuI}
    \item \textcolor{red}{https://www.youtube.com/watch?v=urhdqnmgqeQ}
    \item \textcolor{red}{https://www.youtube.com/watch?v=f8C7KjI\_tuI}
    \item \textcolor{red}{https://www.youtube.com/watch?v=IBCh1tLfCtI}
    \item \textcolor{red}{https://www.youtube.com/watch?v=mEWSVC9FcT4}
\end{itemize}

\textbf{Selected URLs: }\vspace{1pt}
\begin{itemize}[itemsep=0cm]
    \item \textcolor{red}{https://www.youtube.com/watch?v=7ARqB8AjO2k}
    \item \textcolor{red}{https://www.youtube.com/watch?v=2VAY35eMHyU}
    \item \textcolor{red}{https://www.youtube.com/watch?v=f8C7KjI\_tuI}
    \item \textcolor{red}{https://www.youtube.com/watch?v=urhdqnmgqeQ}
\end{itemize}

\textbf{Target URL: }\vspace{1pt}
\begin{itemize}[itemsep=0cm]
    \item \textcolor{PineGreen}{https://www.youtube.com/watch?v=3ug835LFixU}
\end{itemize}
~

\textbf{Extracted frames for video verification:}\vspace{1pt}\\
\includegraphics[width=\columnwidth]{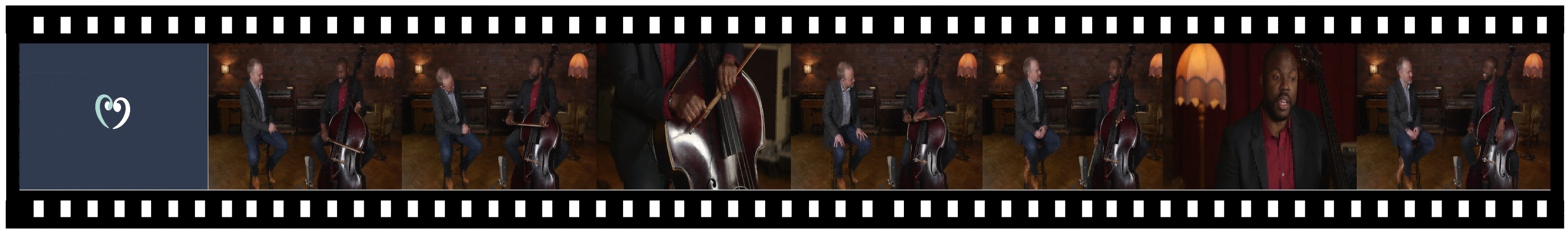}
\includegraphics[width=\columnwidth]{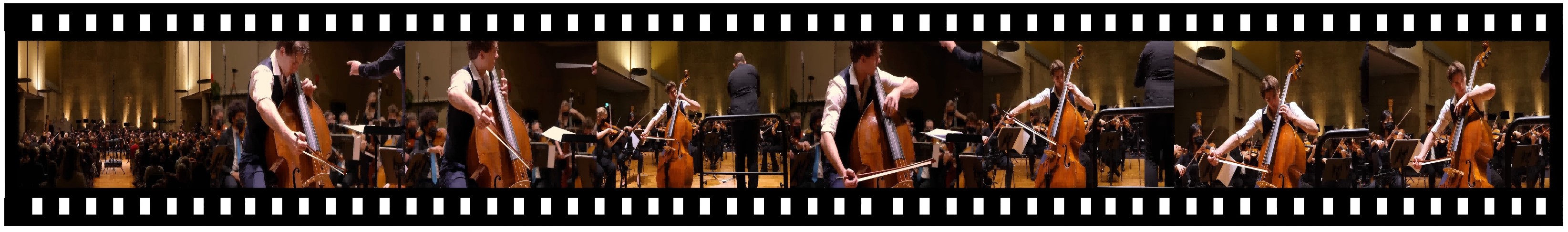}
\includegraphics[width=\columnwidth]{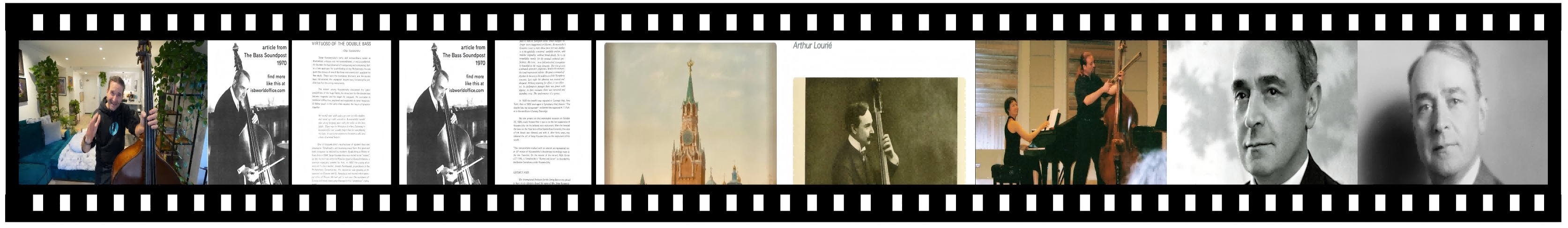}
\includegraphics[width=\columnwidth]{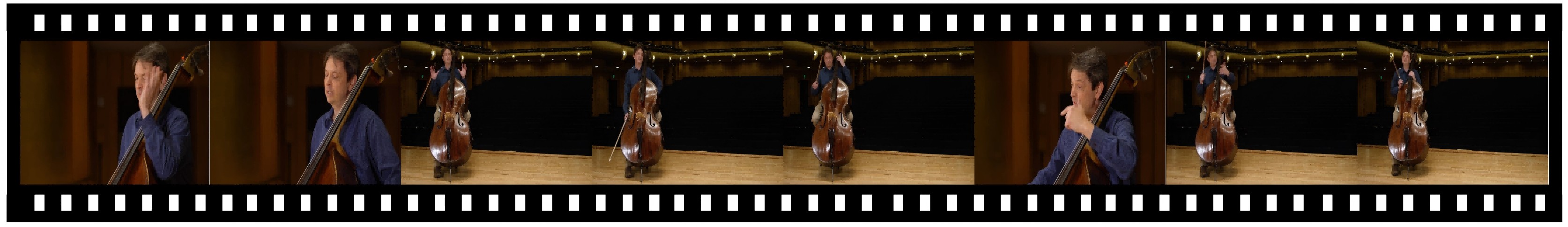}

\textbf{Global Impression ground truth text:}\vspace{1pt}\\
I vaguely recall attending a concert held in a church, where the atmosphere was quiet and solemn, so much so that even breathing seemed to grow softer. The high vaulted ceilings and ancient columns gave the place a strong sense of history, and the air seemed to carry an indescribable sacredness. An elderly conductor stood at the front, his movements calm yet powerful, as if he were telling the story of the music with his entire body. The orchestra was large, with string players, wind instrumentalists, each focused on their role, their sounds weaving together in rich layers. I remember the music ebbing and flowing, sometimes gentle, sometimes surging, perfectly matching the vastness of the church, as if the sounds were moving and echoing along the walls. The entire performance felt meticulously crafted, radiating a solemn beauty that has stayed with me ever since.\\
\textbf{Video verification:}  \textcolor{red}{\textbf{False}}\\

\textbf{Selected frames for localization:} \vspace{1pt}\\
\includegraphics[width=\columnwidth]{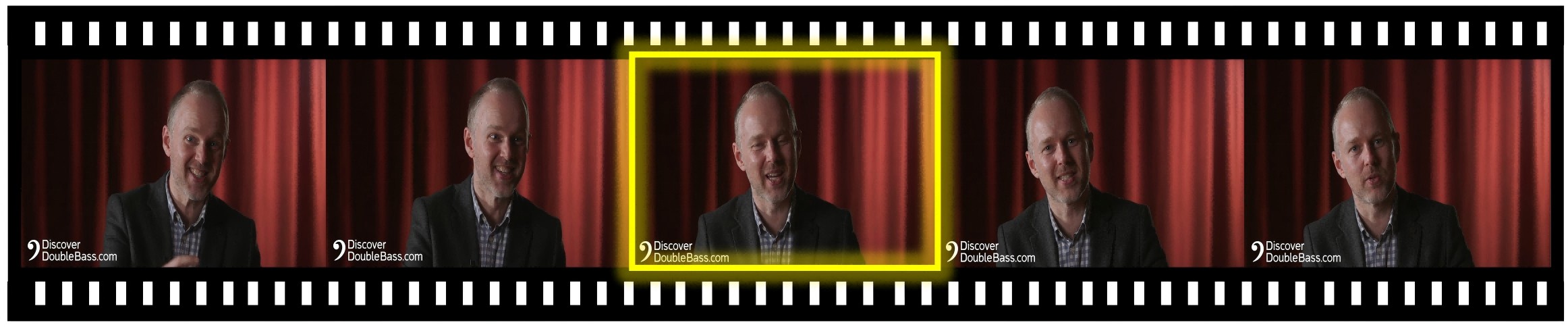}
\includegraphics[width=\columnwidth]{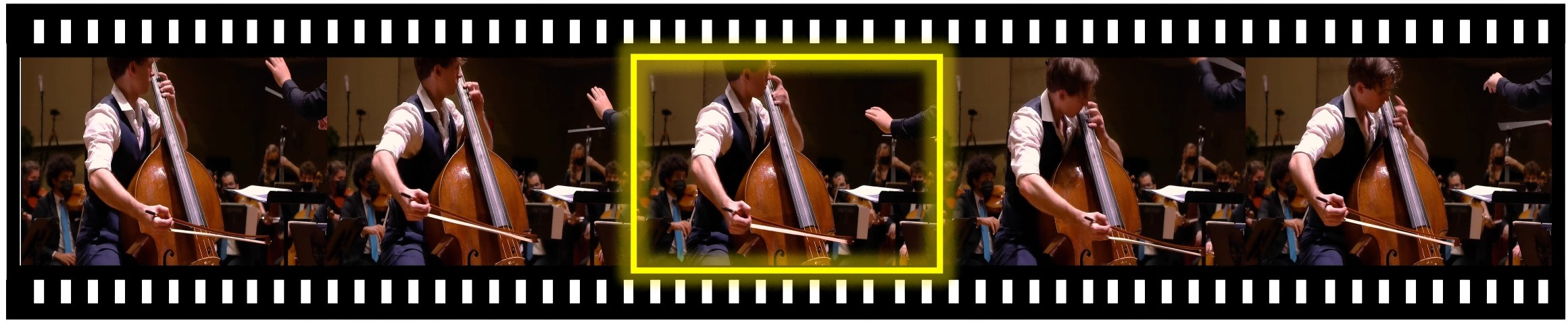}
\includegraphics[width=\columnwidth]{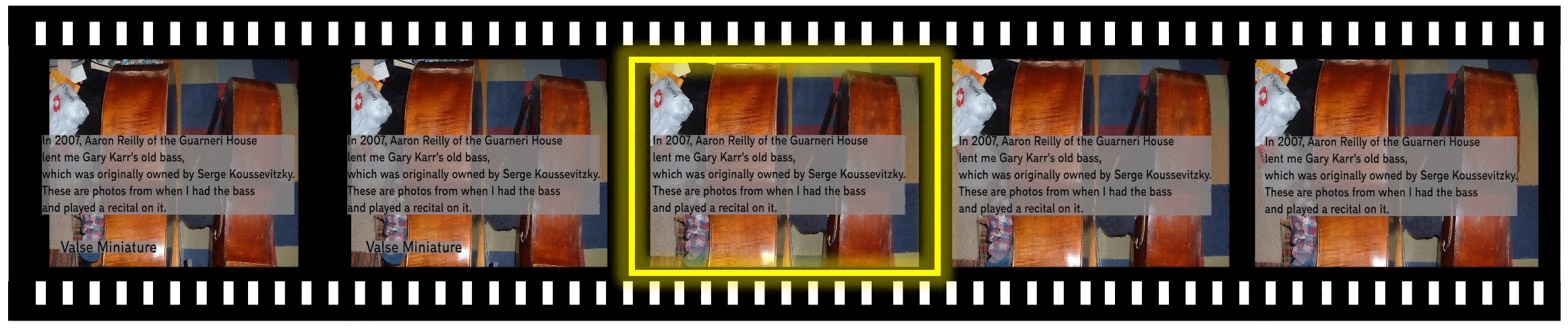}
\includegraphics[width=\columnwidth]{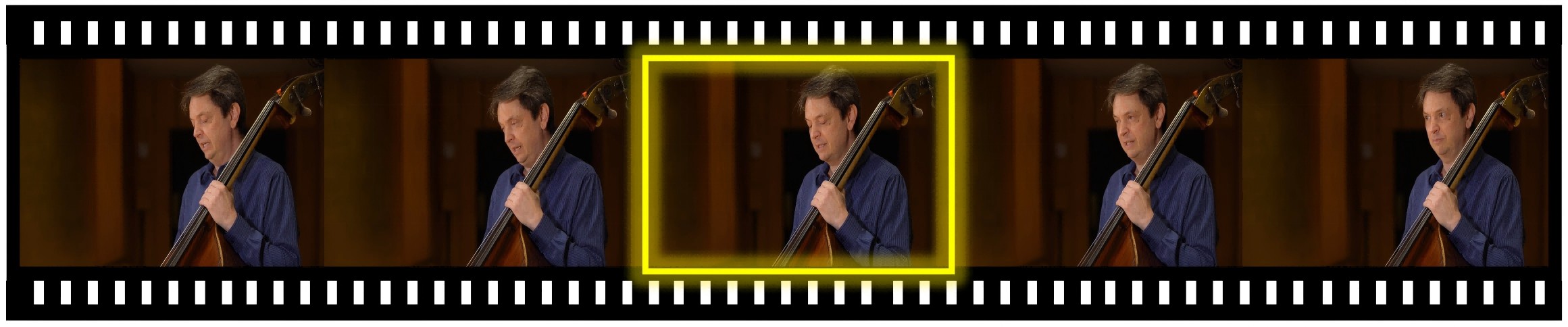}

\textbf{Key moment ground truth text:}\vspace{1pt}\\
Within this broader context, one specific segment I recall is the following. From what I recall, this segment features a close-up of a double bass player's upper body, dressed in a dark suit, his right hand vigorously drawing the bow across the thick strings, the speed creating noticeable motion blur in his hand. The reddish-brown body of the instrument dominates the right side of the frame, with visible wood grain and a glossy finish reflecting soft highlights under the lighting. The shot uses a shallow depth of field, blurring the background completely, with only a faint outline of another string instrument's scroll visible in the lower left. Warm spotlighting comes from the side, emphasizing the texture of the wood and the metallic sheen of the strings, creating a striking contrast between the stillness of the instrument and the motion of the hand, conveying a strong sense of rhythm.

\textbf{Ground truth frame:}\vspace{1pt}\\
\begin{center}
\includegraphics[height=2cm, keepaspectratio]{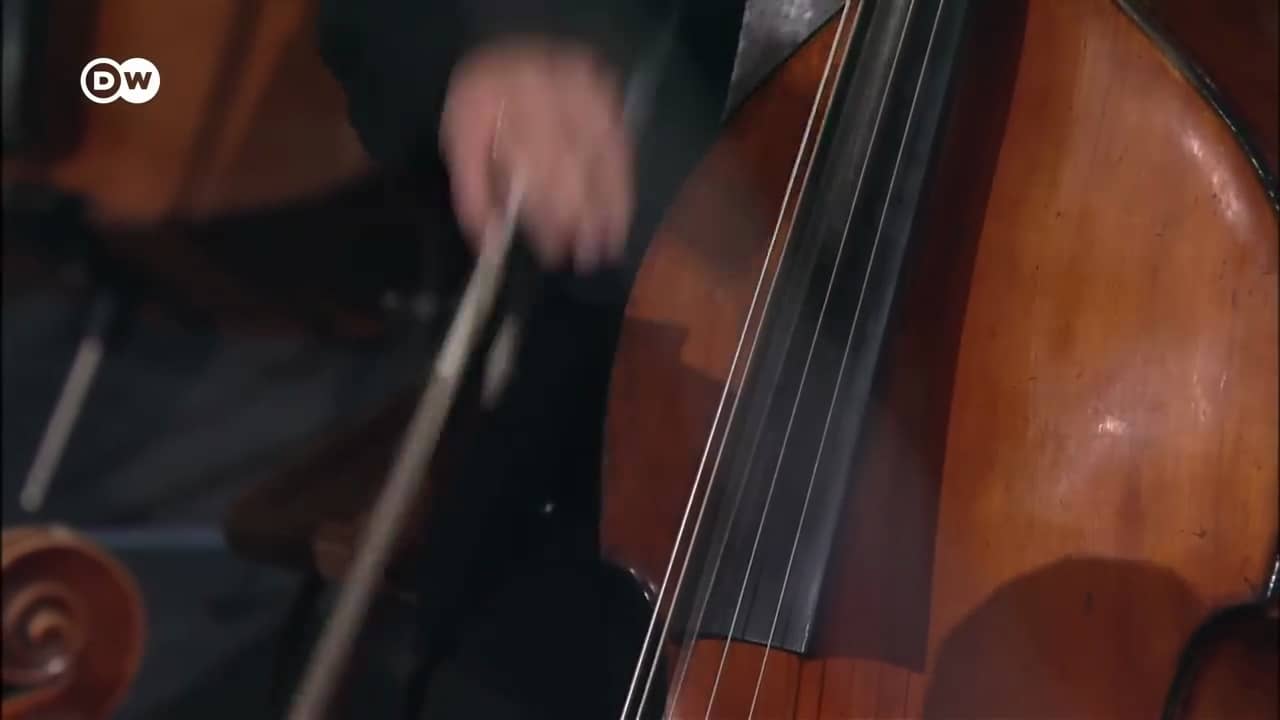}
\vspace{5pt}
\end{center}
\textbf{Model's prediction:}  \textcolor{red}{\textbf{False}}\\

\end{tcolorbox}

%% file: case_study/case2_retrieval2.tex
\tcbset{
    breakable,
    colframe=blue!5!black,
    colback=gray!10!white,
    fonttitle=\bfseries,
    width=\columnwidth 
}

\begin{tcolorbox}[
    title=\textbf{Retrieval Failure due to Limited URL Coverage},
    fonttitle=\bfseries
]


\textbf{Input fragment:(No Global Impression)} \vspace{2pt} \\
Please find the links to this video: One specific segment I recall is the following. From what I recall, this segment features a middle-aged man standing right in the center of the frame, dressed in a light blue traditional exercise suit. His outfit seems to be one of those loose-fitting jackets with traditional frog buttons, and it looks very neat. I remember his posture was something like a horse stance; his feet were spread about shoulder-width apart, and his knees were slightly bent in a half-squat, making his lower body appear very stable. His hands were pressing down gently in front of his lower abdomen with palms facing downward, and his arms formed a circular arc. That pose looked like he was practicing some form of Qigong or Tai Chi, and his expression seemed particularly focused and calm. The background seemed to be a quiet forest; the surroundings were densely packed with tree trunks, and the leaves showed broad patches of golden yellow. It felt like a ginkgo forest in late autumn, while the ground was covered with a green lawn. The lighting appeared quite soft and even, without any harsh direct sunlight, giving off a natural and comfortable feeling. On the right side of the screen, there appeared to be a vertical row of blue characters, seemingly labeling the name of the move, but I can't quite remember exactly what they said.\\

\textbf{MLLM output context:} \vspace{1pt} \\
\textless think\textgreater
The video features a middle-aged man dressed in a light blue traditional Chinese suit with frog buttons, performing Tai Chi or Qigong in a quiet forest setting. His pose is steady, with a horse stance, knees bent for stability, and his hands pressing down in a circular arc. The calm expression on his face complements the serene atmosphere, with golden yellow autumn leaves and soft lighting enhancing the peaceful mood. A vertical row of blue characters on the right side of the frame labels the move, indicating that this is likely an instructional video. The most plausible match is Hypothesis C, where the video is the official instructional material for "Ba Duan Jin" (Eight Pieces of Brocade), produced by the Chinese Health Qigong Association. The specific visual elements, such as the man in a blue suit, the forest setting, and the blue text, align with this widely recognized reference video. The "pressing down" move corresponds to the "Preparation" or "Closing" form of the routine. To locate the video, search queries should focus on "Chinese Health Qigong," "Ba Duan Jin," and specific keywords like "blue suit" or "original video," such as "Chinese Health Qigong Ba Duan Jin original video blue suit" or "Eight-Section Brocade Tutorial."
\textless/think\textgreater

\textless tool\_call\textgreater
\{
\textquotedbl name\textquotedbl: \textquotedbl search\_videos\textquotedbl,
\textquotedbl arguments\textquotedbl:
\{
\textquotedbl query\textquotedbl:
[
\textquotedbl {Chinese Health Qigong Ba Duan Jin original video blue suit}\textquotedbl,
\textquotedbl {Eight-Section Brocade Tutorial}\textquotedbl
]
\}
\}
\textless/tool\_call\textgreater \\

\textbf{Searched URLs: }\vspace{1pt}
\begin{itemize}[itemsep=0cm]
    \item \textcolor{red}{https://www.youtube.com/watch?v=mc1jeKAb6tM}
    \item \textcolor{red}{https://www.youtube.com/watch?v=3FJy0EDoYK0}
    \item \textcolor{red}{https://www.youtube.com/watch?v=qNRAfQXeJRw}
    \item \textcolor{red}{https://www.youtube.com/watch?v=rSzpTLQap0o}
    \item \textcolor{red}{https://www.youtube.com/watch?v=Uf8bDFrYrRc}
    \item \textcolor{red}{https://www.youtube.com/watch?v=Wu0CaL4nJYA}
    \item \textcolor{red}{https://www.youtube.com/watch?v=cvlwcIiOrq0}
    \item \textcolor{red}{https://www.youtube.com/watch?v=59A6UIu\_JXs}
    \item \textcolor{PineGreen}{https://www.youtube.com/watch?v=QDIx22h2TBc}
    \item \textcolor{red}{https://www.youtube.com/watch?v=7yPPjPFtje0}

\end{itemize}

\textbf{Selected URLs: }\vspace{1pt}

\begin{itemize}[itemsep=0cm]
    \item \textcolor{red}{https://www.youtube.com/watch?v=mc1jeKAb6tM}
    \item \textcolor{red}{https://www.youtube.com/watch?v=3FJy0EDoYK0}
    \item \textcolor{red}{https://www.youtube.com/watch?v=Wu0CaL4nJYA}
    \item \textcolor{red}{https://www.youtube.com/watch?v=cvlwcIiOrq0}
\end{itemize}

\textbf{Target URL: }\vspace{1pt}
\begin{itemize}[itemsep=0cm]
    \item \textcolor{PineGreen}{https://www.youtube.com/watch?v=QDIx22h2TBc}
\end{itemize}
~

\textbf{Selected frames for localization:} \vspace{1pt}\\
\includegraphics[width=\columnwidth]{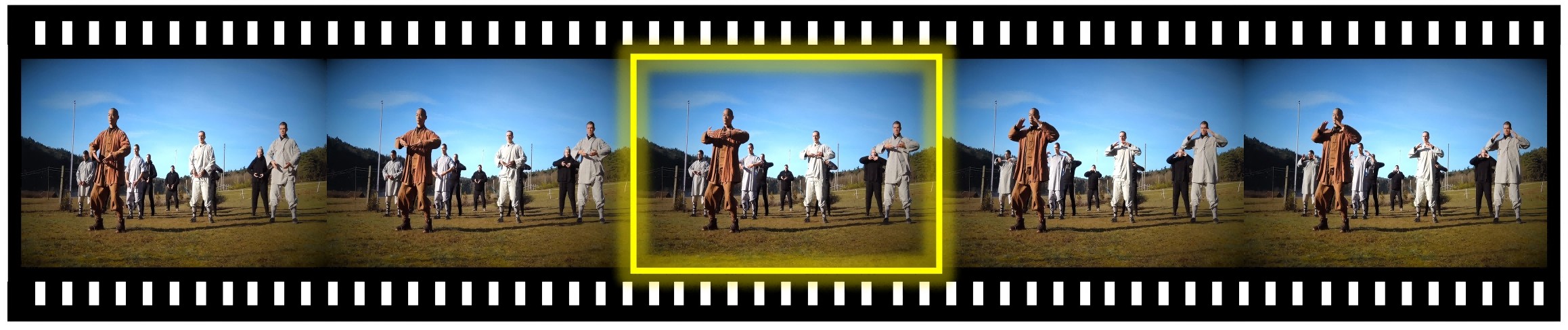}
\includegraphics[width=\columnwidth]{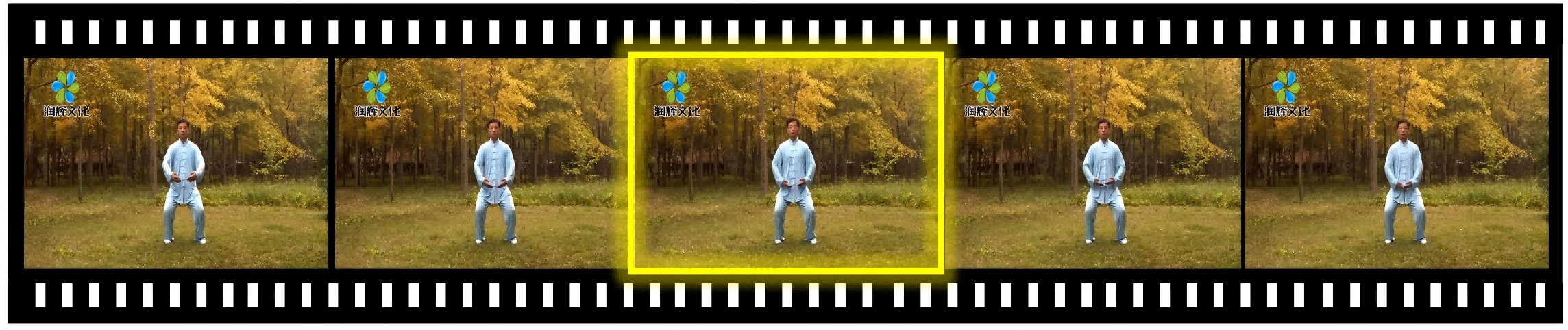}
\includegraphics[width=\columnwidth]{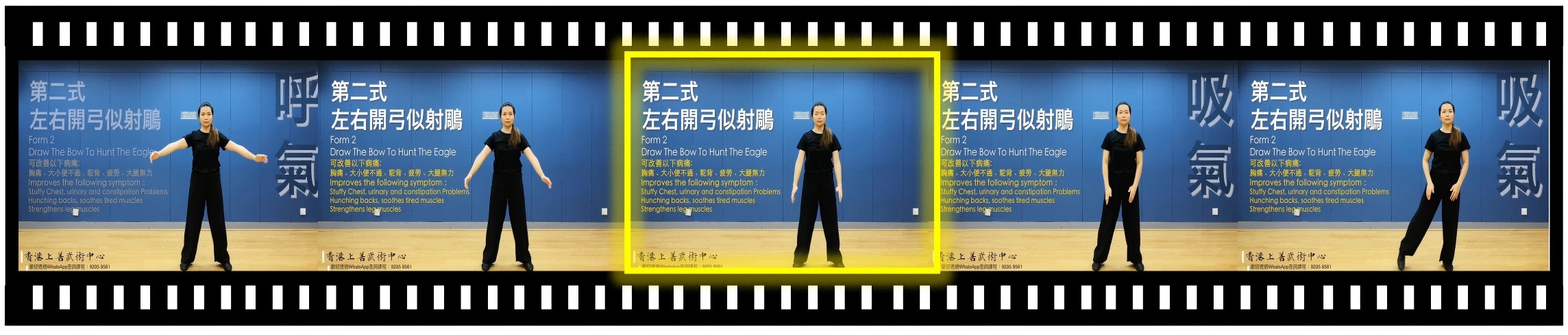}
\includegraphics[width=\columnwidth]{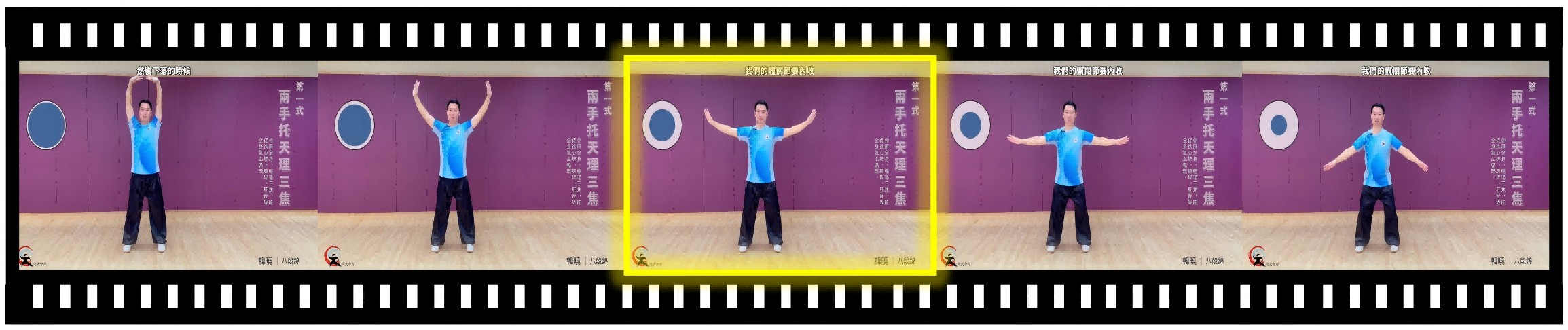}

\textbf{Key moment ground truth text:}\vspace{1pt}\\
One specific segment I recall is the following. From what I recall, this segment features a middle-aged man standing right in the center of the frame, dressed in a light blue traditional exercise suit. His outfit seems to be one of those loose-fitting jackets with traditional frog buttons, and it looks very neat. I remember his posture was something like a horse stance; his feet were spread about shoulder-width apart, and his knees were slightly bent in a half-squat, making his lower body appear very stable. His hands were pressing down gently in front of his lower abdomen with palms facing downward, and his arms formed a circular arc. That pose looked like he was practicing some form of Qigong or Tai Chi, and his expression seemed particularly focused and calm. The background seemed to be a quiet forest; the surroundings were densely packed with tree trunks, and the leaves showed broad patches of golden yellow. It felt like a ginkgo forest in late autumn, while the ground was covered with a green lawn. The lighting appeared quite soft and even, without any harsh direct sunlight, giving off a natural and comfortable feeling. On the right side of the screen, there appeared to be a vertical row of blue characters, seemingly labeling the name of the move, but I can't quite remember exactly what they said.

\textbf{Ground truth frame:}\vspace{1pt}\\
\begin{center}
\includegraphics[height=2cm, keepaspectratio]{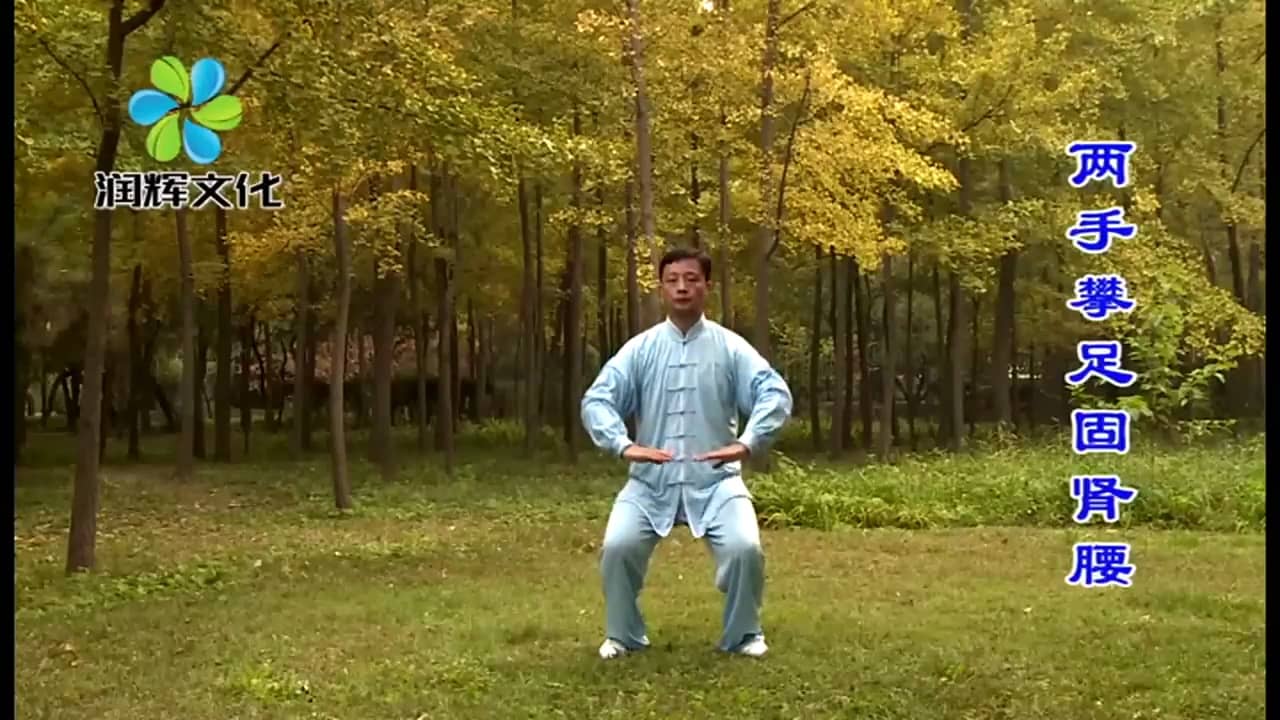}
\vspace{5pt}
\end{center}
\textbf{Model's prediction:}  \textcolor{red}{\textbf{False}}\\

\end{tcolorbox}

%% file: case_study/case3_video_verification1.tex
\tcbset{
    breakable,
    colframe=blue!5!black,
    colback=gray!10!white,
    fonttitle=\bfseries,
    width=\columnwidth 
}

\begin{tcolorbox}[
    title=\textbf{Video Verification with Evaluated Model Error},
    fonttitle=\bfseries
]


\textbf{Input fragment:} \vspace{2pt} \\
Please find the links to this video: I vaguely recall watching a video about systematic fitness training, something like a weekly full-body workout plan—it felt quite professional overall. I remember a male trainer with a solid build and short beard demonstrating exercises in an indoor gym. The space was well-lit but had a cool color tone, giving off a calm and restrained vibe. He wore a simple athletic t-shirt and shorts, moving seriously through large gym machines, focusing mostly on leg-related movements. The pace of the video was steady and orderly. There were text overlays on screen, probably indicating exercise names and sets, though I can’t quite remember the exact details. In the background, there seemed to be other equipment and blurred figures—possibly people working out—but the focus always stayed on the trainer’s movements. From what I recall, the atmosphere was very focused, nothing flashy, like an advanced guide meant for those with some experience. It gave me a sense of discipline and structure, making me feel this was a routine that required consistency. Within this broader context, one specific segment I recall is the following. From what I recall, this segment seems to take place inside a spacious gym where the lighting feels a bit complex. At the center of the view is a sturdy man with a full beard; I believe he's wearing a tight gray T-shirt and dark blue shorts, with very defined muscle contours. I remember him sitting on a massive silver-gray metal machine with his legs extended straight forward horizontally and his feet pressed against the footplate; he looks extremely focused, as if he's building up strength or holding some kind of static exertion pose. In the background, I can vaguely make out other black gym equipment scattered around, and there seem to be blurry figures moving in the distance. The overhead lights seem like typical industrial lighting, casting a cold glow on the metal frame of the machine and reflecting a hard sheen. The entire composition focuses on the man and the giant apparatus, exuding a steady and hardcore training atmosphere that makes you feel the heavy metallic texture even through the screen. Additionally, I recall the audio went like following. I vaguely recall hearing a young man quickly explaining fitness moves, sounding professional and efficient. There was rhythmic electronic music in the background, creating an energetic gym-like atmosphere. It was noise-free, fast-paced, and made me want to follow along.\\

\textbf{MLLM output context:} \vspace{1pt} \\
\textless think\textgreater
The video appears to be a high-production fitness training clip with a cool-toned color palette and industrial gym lighting. It features a male trainer with a solid, muscular build, a short to full beard, wearing a tight gray T-shirt and dark blue shorts, seated on a silver-gray leg press machine with his legs extended in a focused, static pose. The context suggests a systematic, science-based training approach, likely part of a weekly full-body workout plan or an advanced fitness guide. The audio consists of fast-paced, professional narration from a young male voice, accompanied by rhythmic electronic background music, reinforcing the polished, instructional style. These elements closely match the signature aesthetic and content of YouTuber Jeff Nippard, who is well known for his science-driven workout programs, muscular build, beard, and frequent use of cool blue color grading in high-quality videos. While creators like Jeremy Ethier share a similar educational approach, the specific combination of visual style, attire, and training focus aligns most strongly with Jeff Nippard. Effective search queries would therefore center on his name and the workout context, such as “Jeff Nippard full body workout leg press gray shirt.”
\textless/think\textgreater

\textless tool\_call\textgreater
\{
\textquotedbl name\textquotedbl: \textquotedbl search\_videos\textquotedbl,
\textquotedbl arguments\textquotedbl:
\{
\textquotedbl query\textquotedbl:
[
\textquotedbl {Jeff Nippard full body workout leg press gray shirt}\textquotedbl,
\textquotedbl {Jeff Nippard leg press technique original video}\textquotedbl
]
\}
\}
\textless/tool\_call\textgreater \\

\textbf{Selected URLs: }\vspace{1pt}
\begin{itemize}[itemsep=0cm]
    \item {https://www.youtube.com/watch?v=hRZ5MM6gmlE}
    \item {https://www.youtube.com/watch?v=nxisr1AalNc}
    \item {https://www.youtube.com/watch?v=8zWDuWKdBZU}
    \item {https://www.youtube.com/watch?v=B6rGDcfyPto}
\end{itemize}

\textbf{Target URL: }\vspace{1pt}
\begin{itemize}[itemsep=0cm]
    \item {https://www.youtube.com/watch?v=GbaUvwj-ZhY}
\end{itemize}
~

\textbf{Extracted frames for video verification:}\vspace{1pt}\\
\includegraphics[width=\columnwidth]{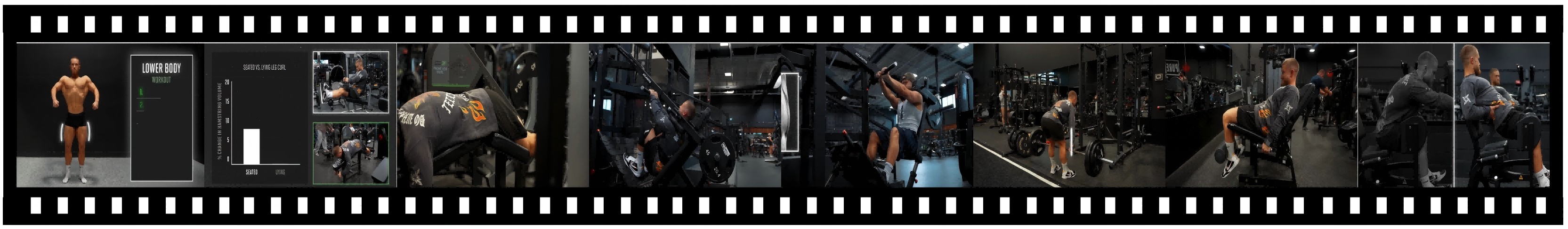}
\includegraphics[width=\columnwidth]{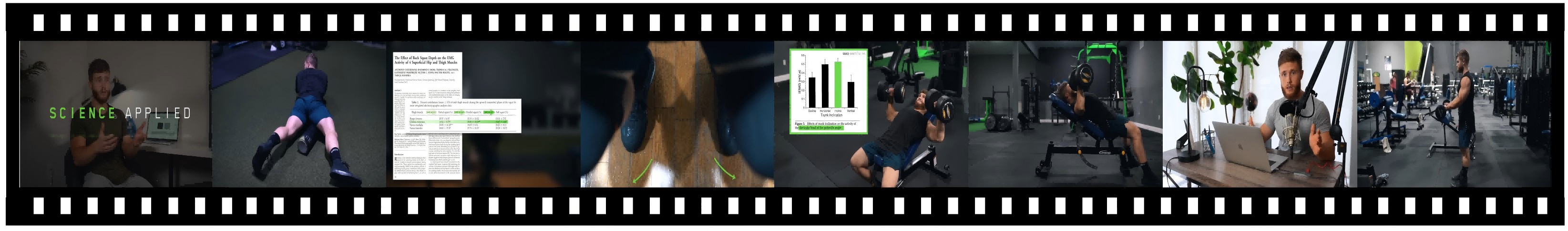}
\includegraphics[width=\columnwidth]{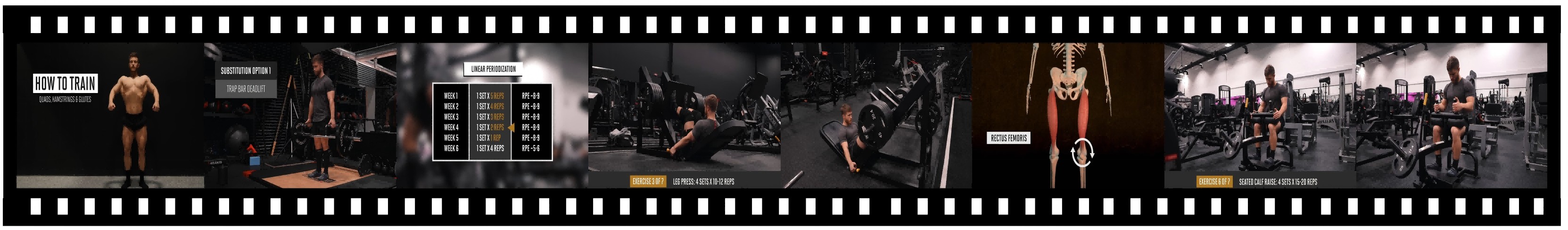}
\includegraphics[width=\columnwidth]{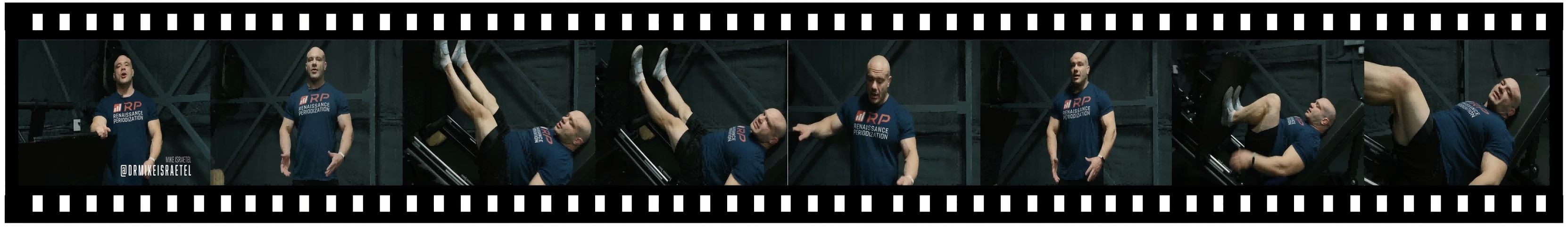}

\textbf{Global Impression ground truth text:}\vspace{1pt}\\
I vaguely recall watching a video about systematic fitness training, something like a weekly full-body workout plan—it felt quite professional overall. I remember a male trainer with a solid build and short beard demonstrating exercises in an indoor gym. The space was well-lit but had a cool color tone, giving off a calm and restrained vibe. He wore a simple athletic t-shirt and shorts, moving seriously through large gym machines, focusing mostly on leg-related movements. The pace of the video was steady and orderly. There were text overlays on screen, probably indicating exercise names and sets, though I can’t quite remember the exact details. In the background, there seemed to be other equipment and blurred figures—possibly people working out—but the focus always stayed on the trainer’s movements.\\
\textbf{Video verification:}  \textcolor{red}{\textbf{False}}\\
\textbf{Human verification:}  \textcolor{PineGreen}{\textbf{True}}\\

\textbf{Selected frames for localization:} \vspace{1pt}\\
\includegraphics[width=\columnwidth]{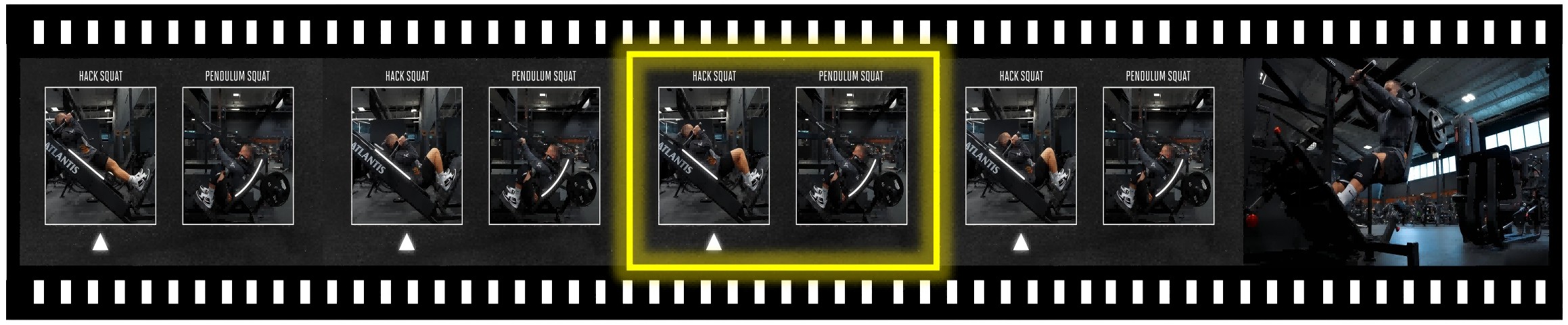}
\includegraphics[width=\columnwidth]{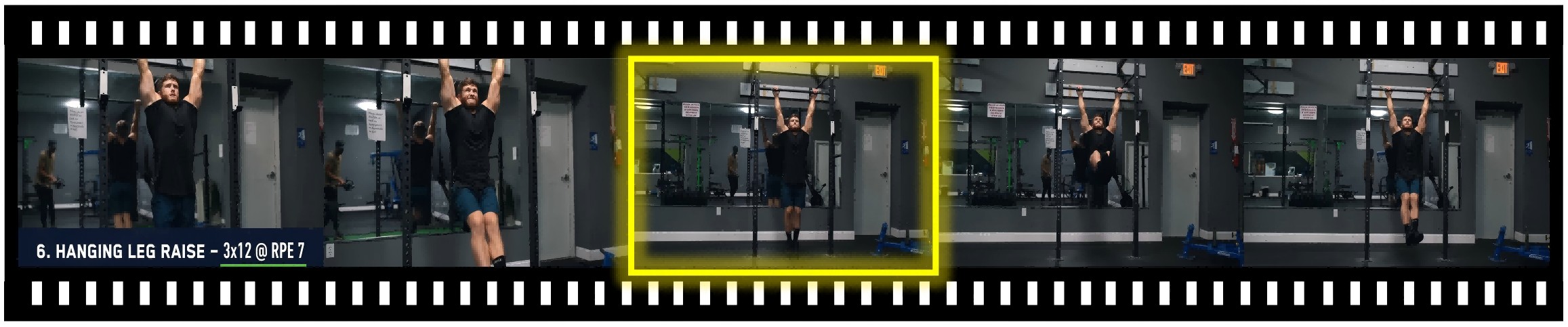}
\includegraphics[width=\columnwidth]{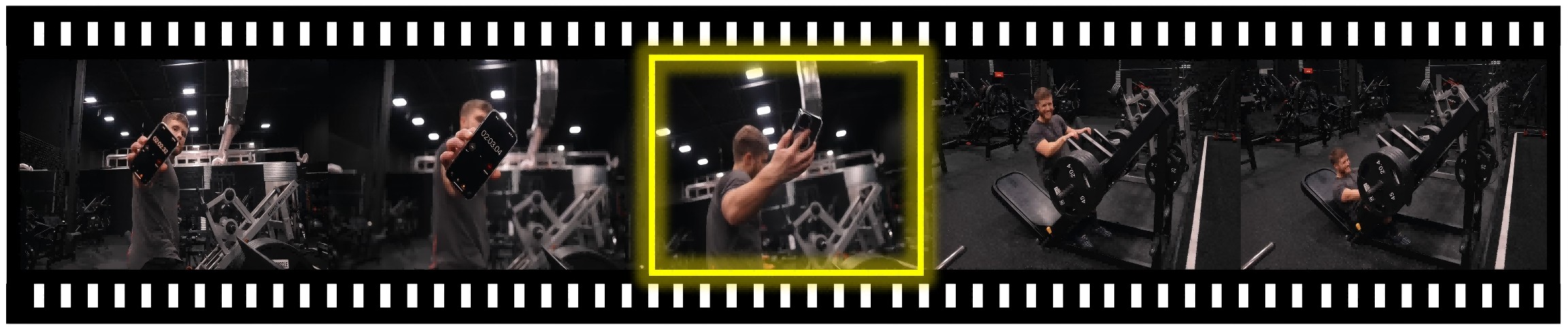}
\includegraphics[width=\columnwidth]{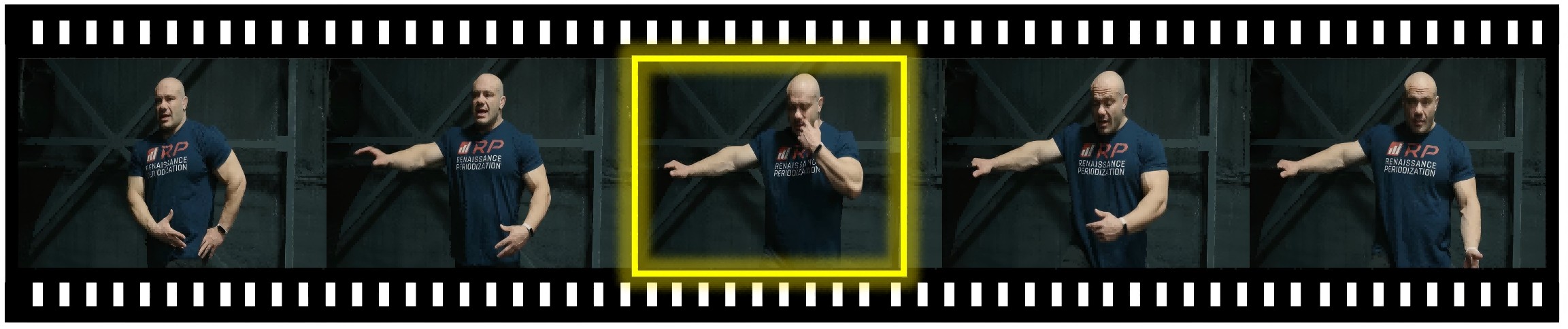}

\textbf{Key moment ground truth text:}\vspace{1pt}\\
From what I recall, the atmosphere was very focused, nothing flashy, like an advanced guide meant for those with some experience. It gave me a sense of discipline and structure, making me feel this was a routine that required consistency. Within this broader context, one specific segment I recall is the following. From what I recall, this segment seems to take place inside a spacious gym where the lighting feels a bit complex. At the center of the view is a sturdy man with a full beard; I believe he's wearing a tight gray T-shirt and dark blue shorts, with very defined muscle contours. I remember him sitting on a massive silver-gray metal machine with his legs extended straight forward horizontally and his feet pressed against the footplate; he looks extremely focused, as if he's building up strength or holding some kind of static exertion pose. In the background, I can vaguely make out other black gym equipment scattered around, and there seem to be blurry figures moving in the distance. The overhead lights seem like typical industrial lighting, casting a cold glow on the metal frame of the machine and reflecting a hard sheen. The entire composition focuses on the man and the giant apparatus, exuding a steady and hardcore training atmosphere that makes you feel the heavy metallic texture even through the screen.

\textbf{Ground truth frame:}\vspace{1pt}\\
\begin{center}
\includegraphics[height=2cm, keepaspectratio]{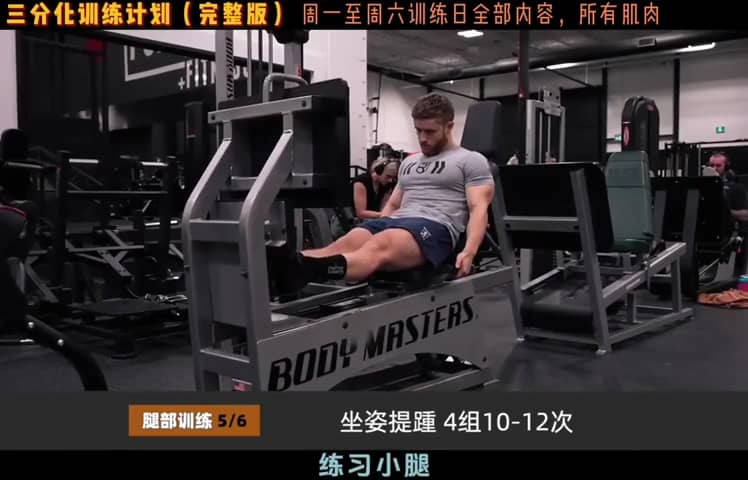}
\vspace{5pt}
\end{center}
\textbf{Model's prediction:}  \textcolor{PineGreen}{\textbf{True}}\\

\end{tcolorbox}

%% file: case_study/case3_video_verification2.tex
\tcbset{
    breakable,
    colframe=blue!5!black,
    colback=gray!10!white,
    fonttitle=\bfseries,
    width=\columnwidth 
}

\begin{tcolorbox}[
    title=\textbf{Video Verification with Judge Model Error},
    fonttitle=\bfseries
]


\textbf{Input fragment:} \vspace{2pt} \\
Please find the links to this video: I remember it was something about two high school girls, though the details are a bit fuzzy. They were always hanging out together—at school, in malls, at home—talking in this sharp, crude way, swearing constantly, but with this weird sense of humor in their tone. From what I recall, they didn’t care about studying at all. Instead, they spent all their time scheming how to make money, and I think they even mentioned selling fake stuff to trick people. In the mall, they seemed restless, talking back to store clerks and casually taking things, acting pretty out of line. Their conversations were all over the place—drugs, deception, celebrity gossip—just a messy mix. The whole vibe was kind of chaotic and sarcastic, like they were mocking everything around them. The background music was surprisingly upbeat, a light pop-electronic tune, which only made their dialogue feel more rebellious and absurd. I vaguely remember them never seeming to think they were doing anything wrong, just carrying on endlessly, like they’d never stop. Within this broader context, one specific segment I recall is the following. From what I recall, this segment features a bright and clean indoor setting resembling a Japanese-style ramen shop or lounge area. Three characters are seated with their backs to the camera along a long counter. On the left is a woman with wavy platinum-blonde hair, wearing a pink cropped top and dark jeans, sitting upright. To her right is another woman with a brown ponytail, dressed in a light blue short-sleeve shirt and shorts, appearing more relaxed. Separated by two empty stools on the far right sits a man with short brown hair, wearing a dark brown T-shirt and light blue jeans, slightly turned to the left. The lighting is bright and even, with daylight filtering through half-lowered beige roller blinds, complemented by warm-toned linear lamps above the counter. The gray concrete-like floor shows distinct circular shadows beneath the stools. The composition is orderly and symmetrical, emphasizing spatial openness and subtle emotional distance.\\

\textbf{MLLM output context:} \vspace{1pt} \\
\textless think\textgreater
The video depicts two crude, foul-mouthed high school girls scheming to make money through dubious means, hanging around malls, shoplifting, and exuding a chaotic, sarcastic energy, all set to upbeat pop-electronic music. A specific visual segment shows a bright, clean, Japanese-style ramen shop or lounge, with three characters seated at a counter viewed from behind: on the left, a woman with wavy platinum-blonde hair wearing a pink crop top and dark jeans; in the middle, a brunette with a brown ponytail in a light blue shirt and shorts; and on the right, a man with short brown hair in a dark T-shirt, separated from the two women by empty stools. This combination of personality traits, dialogue style, and visual design strongly points to the visual novel Class of ’09 (or its sequel The Re-Up), with the two women clearly resembling the main characters Nicole (the blonde) and Jecka (the brunette). The specific ramen-shop counter composition, shown from behind, suggests a distinctive background CG, title screen, or trailer shot rather than a generic anime scene. While alternatives such as anime series or GTA role-play content are possible, the crude humor, character dynamics, and outfit details align most closely with Class of ’09. Effective searches should therefore focus on the game and characters, using keywords such as “Class of ’09 Nicole Jecka ramen shop,” “Class of ’09 opening scene,” or “Class of ’09 background CG.”
\textless/think\textgreater

\textless tool\_call\textgreater
\{
\textquotedbl name\textquotedbl: \textquotedbl search\_videos\textquotedbl,
\textquotedbl arguments\textquotedbl:
\{
\textquotedbl query\textquotedbl:
[
\textquotedbl {Class of '09 game Nicole and Jecka ramen shop scene}\textquotedbl,
\textquotedbl {Class of '09 The Re-Up intro video}\textquotedbl
]
\}
\}
\textless/tool\_call\textgreater \\

\textbf{Selected URLs: }\vspace{1pt}
\begin{itemize}[itemsep=0cm]
    \item {https://www.youtube.com/watch?v=81fkaByThRM}
    \item {https://www.youtube.com/watch?v=iRdUVji7LuQ}
    \item {https://www.youtube.com/watch?v=UKeZUnZw6N8}
    \item {https://www.youtube.com/watch?v=CezOyucCeuw}
\end{itemize}

\textbf{Target URL: }\vspace{1pt}
\begin{itemize}[itemsep=0cm]
    \item {https://www.youtube.com/watch?v=Rb1z6URgA1w}
\end{itemize}
~

\textbf{Extracted frames for video verification:}\vspace{1pt}\\
\includegraphics[width=\columnwidth]{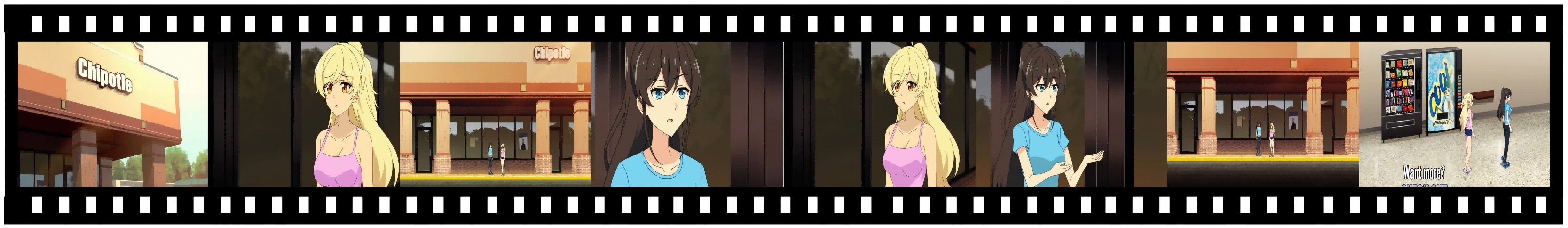}
\includegraphics[width=\columnwidth]{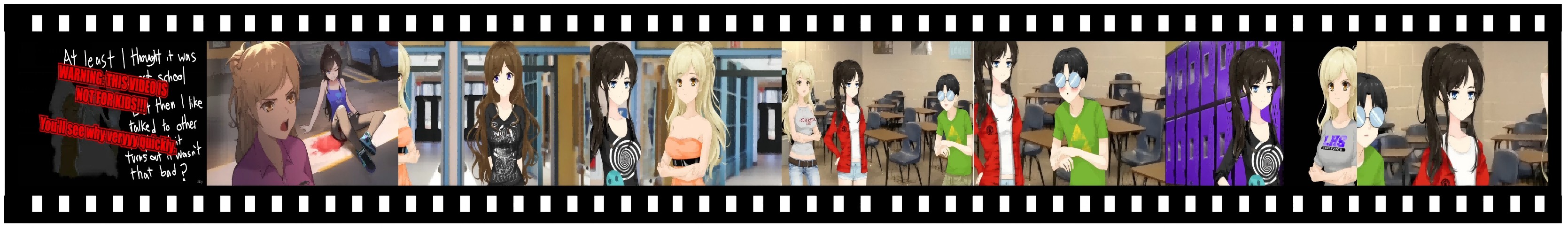}
\includegraphics[width=\columnwidth]{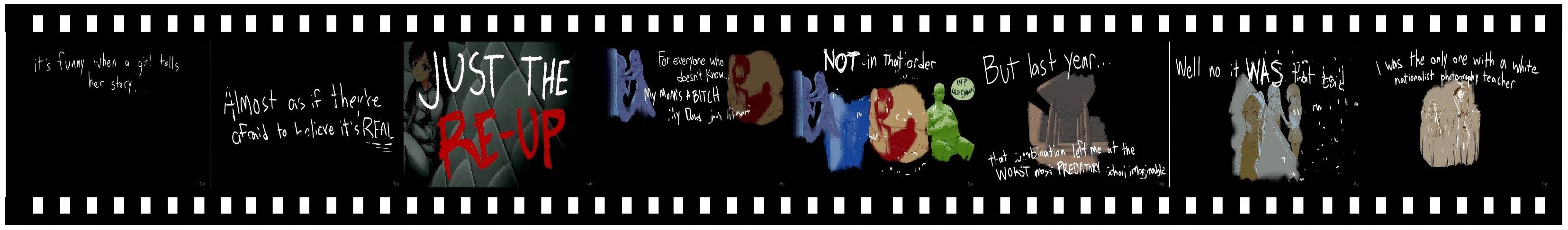}
\includegraphics[width=\columnwidth]{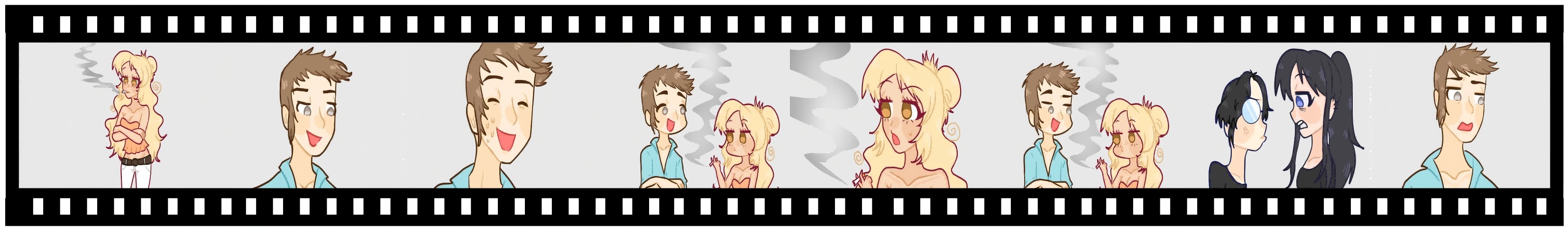}

\textbf{Global Impression ground truth text:}\vspace{1pt}\\
I remember it was something about two high school girls, though the details are a bit fuzzy. They were always hanging out together—at school, in malls, at home—talking in this sharp, crude way, swearing constantly, but with this weird sense of humor in their tone. From what I recall, they didn’t care about studying at all. Instead, they spent all their time scheming how to make money, and I think they even mentioned selling fake stuff to trick people. In the mall, they seemed restless, talking back to store clerks and casually taking things, acting pretty out of line. Their conversations were all over the place—drugs, deception, celebrity gossip—just a messy mix. The whole vibe was kind of chaotic and sarcastic, like they were mocking everything around them. The background music was surprisingly upbeat, a light pop-electronic tune, which only made their dialogue feel more rebellious and absurd. I vaguely remember them never seeming to think they were doing anything wrong, just carrying on endlessly, like they’d never stop.\\
\textbf{Video verification:}  \textcolor{PineGreen}{\textbf{True}}\\
\textbf{Gemini 2.5 pro verification:}  \textcolor{red}{\textbf{False}}\\

\textbf{Selected frames for localization:} \vspace{1pt}\\
\includegraphics[width=\columnwidth]{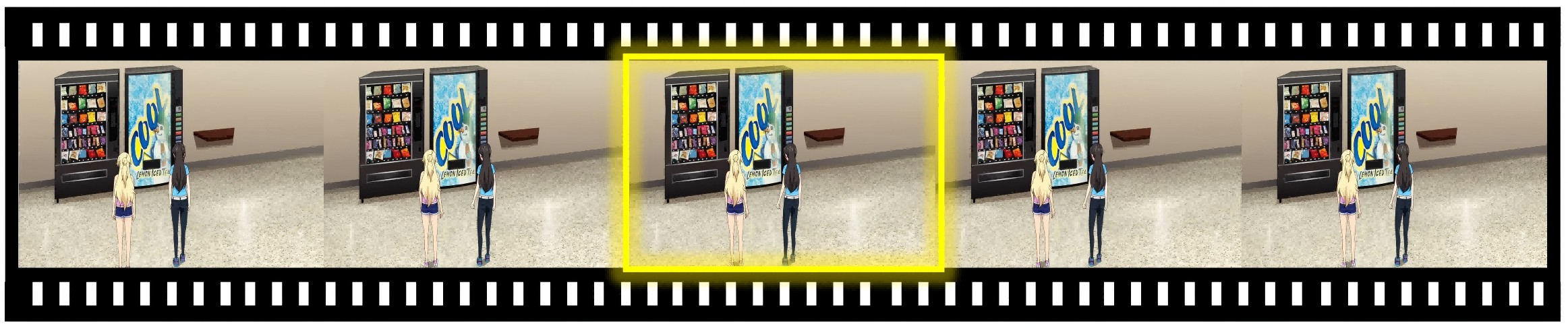}
\includegraphics[width=\columnwidth]{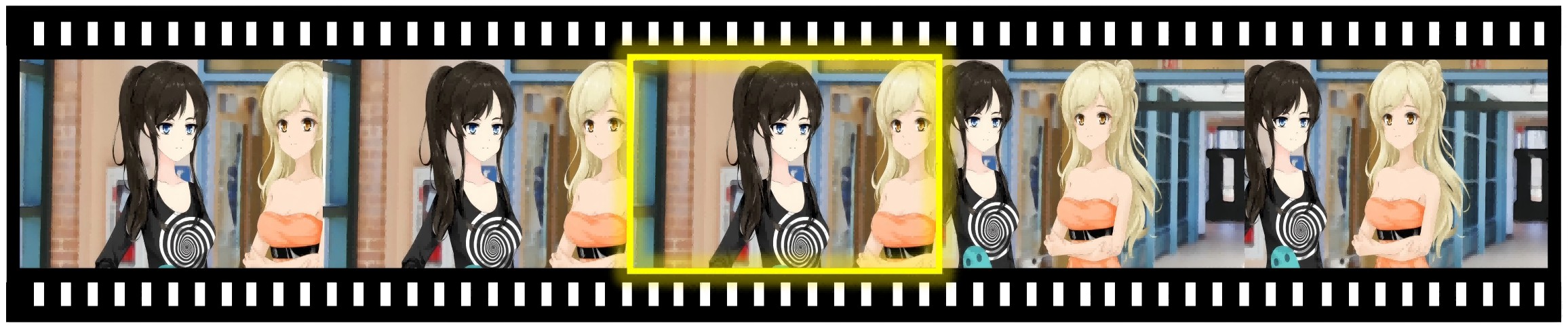}
\includegraphics[width=\columnwidth]{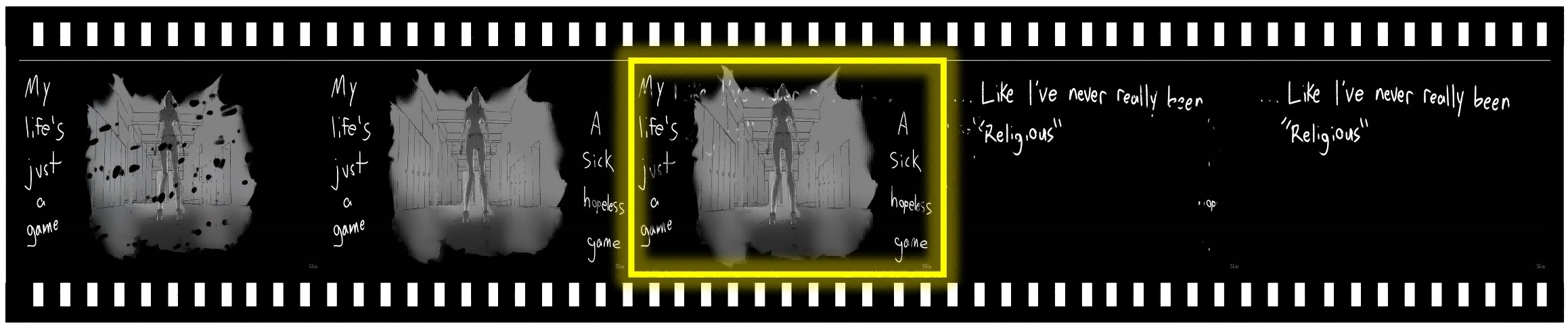}
\includegraphics[width=\columnwidth]{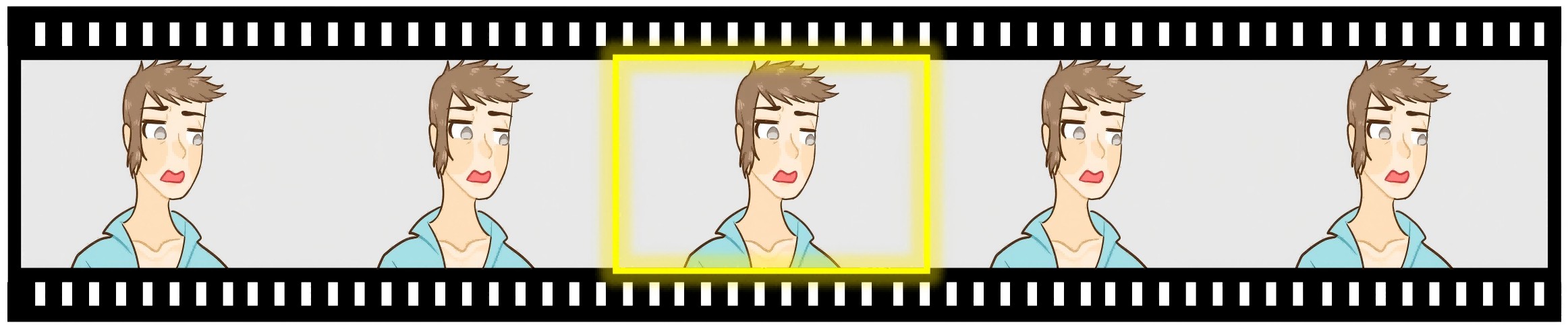}

\textbf{Key moment ground truth text:}\vspace{1pt}\\
Within this broader context, one specific segment I recall is the following. From what I recall, this segment features a bright and clean indoor setting resembling a Japanese-style ramen shop or lounge area. Three characters are seated with their backs to the camera along a long counter. On the left is a woman with wavy platinum-blonde hair, wearing a pink cropped top and dark jeans, sitting upright. To her right is another woman with a brown ponytail, dressed in a light blue short-sleeve shirt and shorts, appearing more relaxed. Separated by two empty stools on the far right sits a man with short brown hair, wearing a dark brown T-shirt and light blue jeans, slightly turned to the left. The lighting is bright and even, with daylight filtering through half-lowered beige roller blinds, complemented by warm-toned linear lamps above the counter. The gray concrete-like floor shows distinct circular shadows beneath the stools. The composition is orderly and symmetrical, emphasizing spatial openness and subtle emotional distance.

\textbf{Ground truth frame:}\vspace{1pt}\\
\begin{center}
\includegraphics[height=2cm, keepaspectratio]{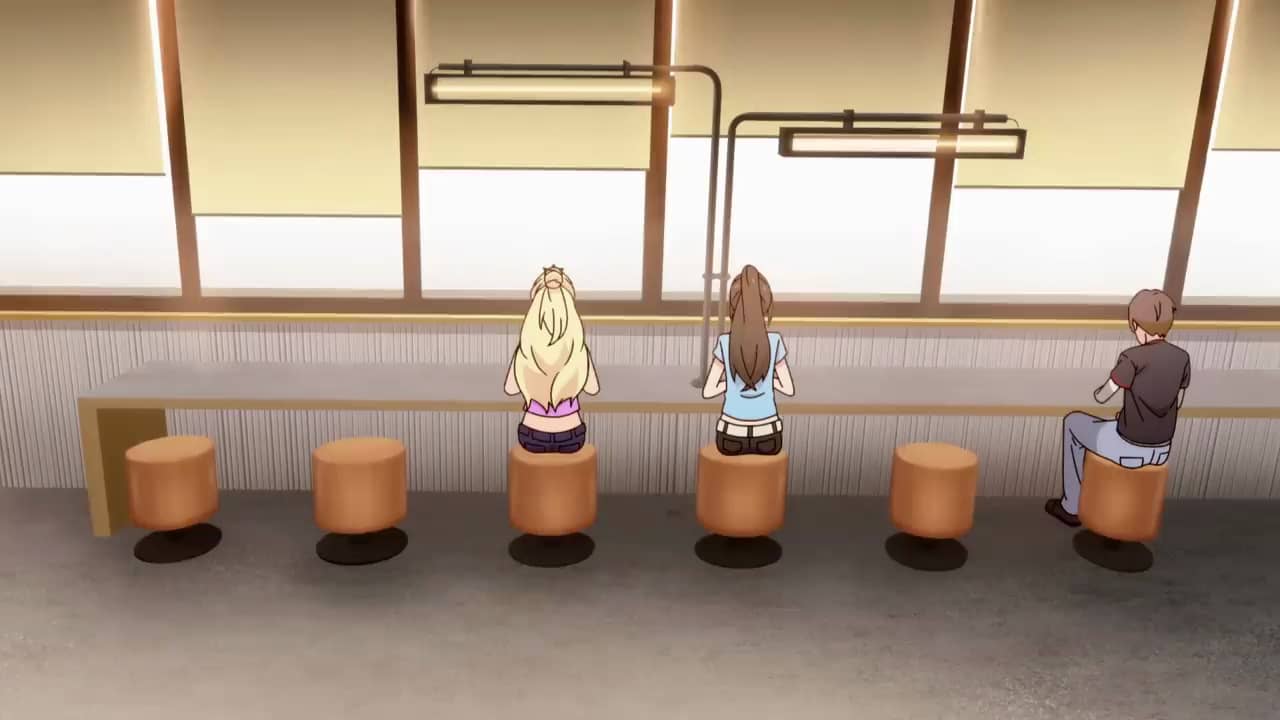}
\vspace{5pt}
\end{center}
\textbf{Model's prediction:}  \textcolor{red}{\textbf{False}}\\

\end{tcolorbox}

%% file: case_study/case4_localization1.tex
\tcbset{
    breakable,
    colframe=blue!5!black,
    colback=gray!10!white,
    fonttitle=\bfseries,
    width=\columnwidth 
}

\begin{tcolorbox}[
    title=\textbf{Localization with Evaluated Model Error},
    fonttitle=\bfseries
]


\textbf{Input fragment:} \vspace{2pt} \\
Please find the links to this video: I vaguely recall it was maybe a music video, or perhaps the theme song from a film, with an overall subtle artistic feel. There were scenes of mountains, forests, and lakesides, along with quiet indoor moments, where shifting light and shadow carried a faintly cool, melancholic tone. From what I remember, the story was about a relationship that never came to be—one of missed chances and regret. The emotions were restrained, yet they slowly seeped into your heart. The singer appeared throughout these fragments, performing with deep but measured emotion, as if recalling a time long gone. I remember the atmosphere—never loud, but heavy, like flipping through an old diary where the words remain unchanged, but the person is already far away. The whole piece had no dramatic outbursts, yet through still images and melody, it quietly unfolded a gentle sadness, leaving you with a hollow feeling inside, as if you’d lived through that silent farewell yourself. Within this broader context, one specific segment I recall is the following. From what I recall, this segment features a clear, shallow stream or pond, with delicate concentric ripples spreading across the surface on the right side, as if a drop had just fallen. Through the transparent water, the riverbed is visible, covered in various sizes of gray-brown pebbles. Exposed soil and rocks appear in the upper right, while a few green blades of grass dip into the frame from the lower left. The water reflects the intricate silhouettes of tree branches above, creating a layered effect between reflection and reality. The scene is bathed in a soft teal hue, with gentle, shimmering light, captured from a high-angle perspective that emphasizes the texture of the water and the stillness of the environment.\\

\textbf{MLLM output context:} \vspace{1pt} \\
\textless think\textgreater
The video appears to be a music video or film theme song with a subtle, artistic tone and a cool, melancholic atmosphere, evoking restrained emotion, a sense of silent farewell, and the feeling of flipping through an old diary. The narrative centers on a relationship that never fully materialized, marked by missed chances and lingering regret. Visually, the video interweaves scenes of mountains, forests, lakesides, quiet indoor moments, and the singer performing, reinforcing the introspective mood. One particularly distinctive segment shows a high-angle shot of a shallow stream or pond with concentric ripples forming on the right side, gray-brown pebbles and exposed soil in the upper right, green grass in the lower left, and reflections of tree branches on the water’s surface, all rendered in a soft teal color grading. This combination of visual motifs and emotional themes strongly points to an Asian ballad, with Jay Chou’s “Blue and White Porcelain” standing out as the most likely match, given its iconic association with “sky cyan” tones, water imagery, and themes of longing and missed connections. Other possibilities include similarly styled works such as “Orchids in the Moonlight”, “Hair Like Snow”, or “The Sound of Rain”, which share comparable aesthetics and emotional resonance. To pinpoint the exact video, effective searches would combine the distinctive visual description with the narrative theme and likely artist, using queries such as “music video shallow stream ripples teal hue,” “Jay Chou Blue and White Porcelain MV stream,” or “music video missed chances regret forest lake singer.”
\textless/think\textgreater

\textless tool\_call\textgreater
\{
\textquotedbl name\textquotedbl: \textquotedbl search\_videos\textquotedbl,
\textquotedbl arguments\textquotedbl:
\{
\textquotedbl query\textquotedbl:
[
\textquotedbl {music video shallow stream ripples pebbles teal hue}\textquotedbl,
\textquotedbl {Jay Chou Blue and White Porcelain MV water scene}\textquotedbl
]
\}
\}
\textless/tool\_call\textgreater \\

\textbf{Selected URLs: }\vspace{1pt}
\begin{itemize}[itemsep=0cm]
    \item {https://www.youtube.com/watch?v=bo\_efYhYU2A}
    \item {https://www.youtube.com/watch?v=JPJjwHAIny4}
    \item {https://www.youtube.com/watch?v=ElkLfAGvMdc}
    \item {https://www.youtube.com/watch?v=tBJ8blu6TLQ}
\end{itemize}

\textbf{Target URL: }\vspace{1pt}
\begin{itemize}[itemsep=0cm]
    \item {https://www.youtube.com/watch?v=2BEFukvLZfI}
\end{itemize}
~

\textbf{Extracted frames for video verification:}\vspace{1pt}\\
\includegraphics[width=\columnwidth]{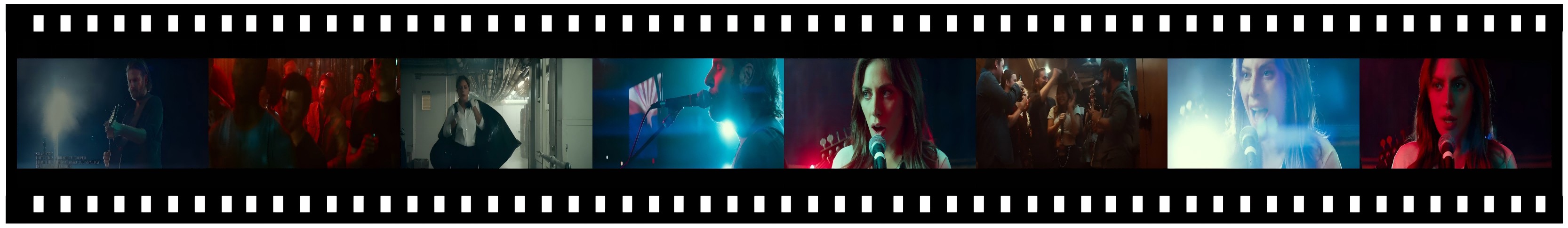}
\includegraphics[width=\columnwidth]{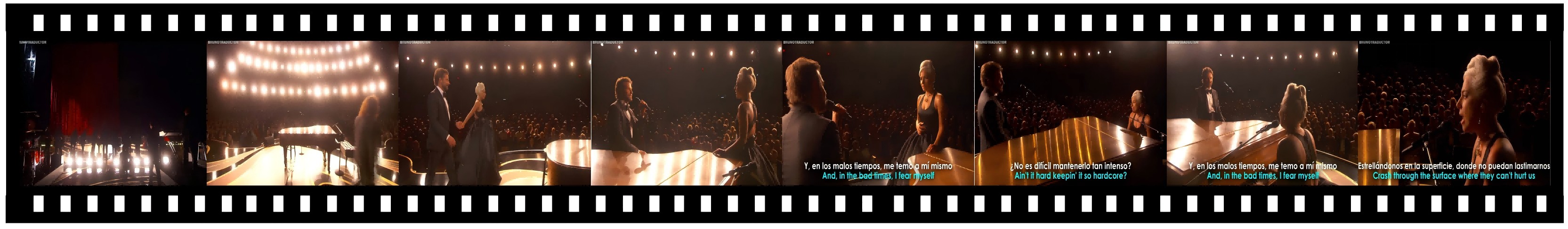}
\includegraphics[width=\columnwidth]{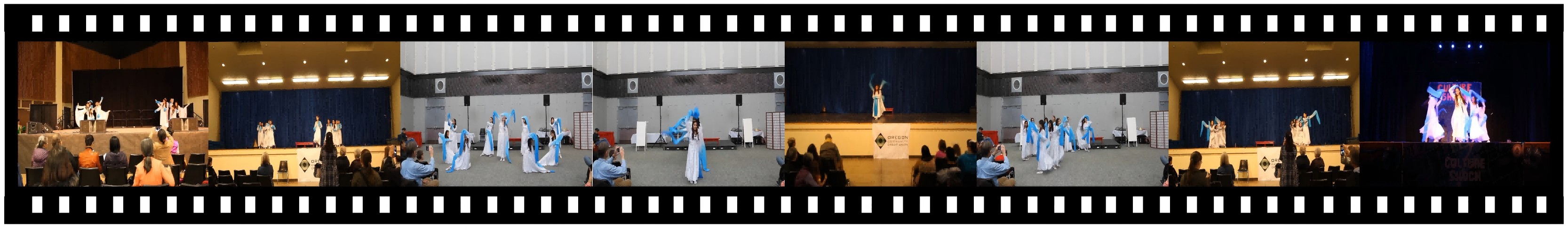}
\includegraphics[width=\columnwidth]{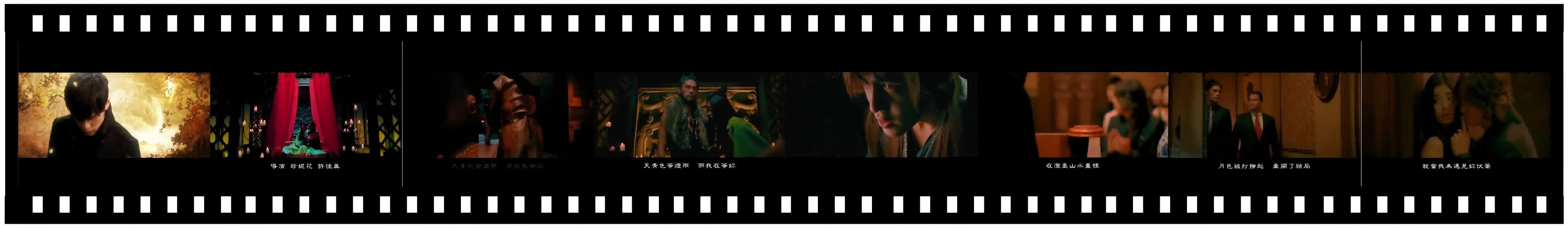}

\textbf{Global Impression ground truth text:}\vspace{1pt}\\
I vaguely recall it was maybe a music video, or perhaps the theme song from a film, with an overall subtle artistic feel. There were scenes of mountains, forests, and lakesides, along with quiet indoor moments, where shifting light and shadow carried a faintly cool, melancholic tone. From what I remember, the story was about a relationship that never came to be—one of missed chances and regret. The emotions were restrained, yet they slowly seeped into your heart. The singer appeared throughout these fragments, performing with deep but measured emotion, as if recalling a time long gone. I remember the atmosphere—never loud, but heavy, like flipping through an old diary where the words remain unchanged, but the person is already far away. The whole piece had no dramatic outbursts, yet through still images and melody, it quietly unfolded a gentle sadness, leaving you with a hollow feeling inside, as if you’d lived through that silent farewell yourself.\\

\textbf{Video verification:}  \textcolor{red}{\textbf{False}}\\

\textbf{Selected frames for localization:} \vspace{1pt}\\
\includegraphics[width=\columnwidth]{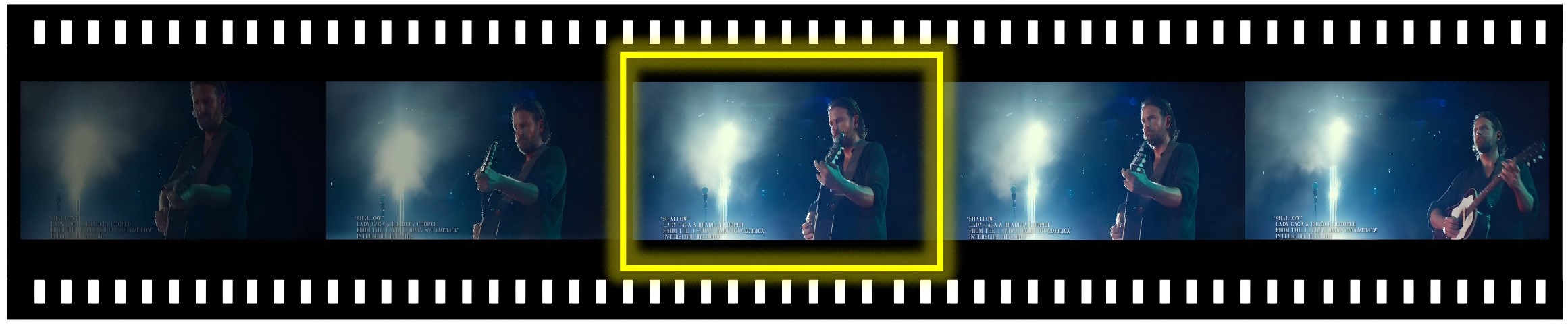}
\includegraphics[width=\columnwidth]{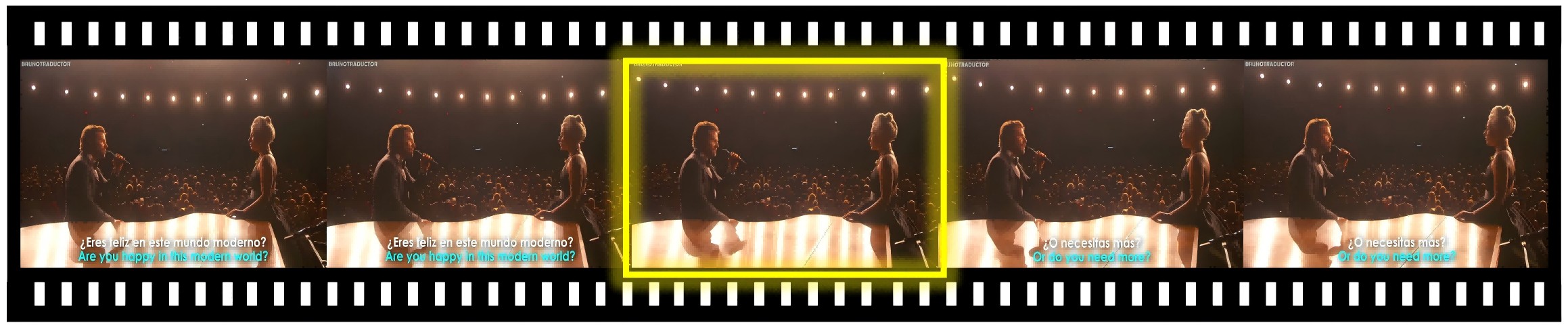}
\includegraphics[width=\columnwidth]{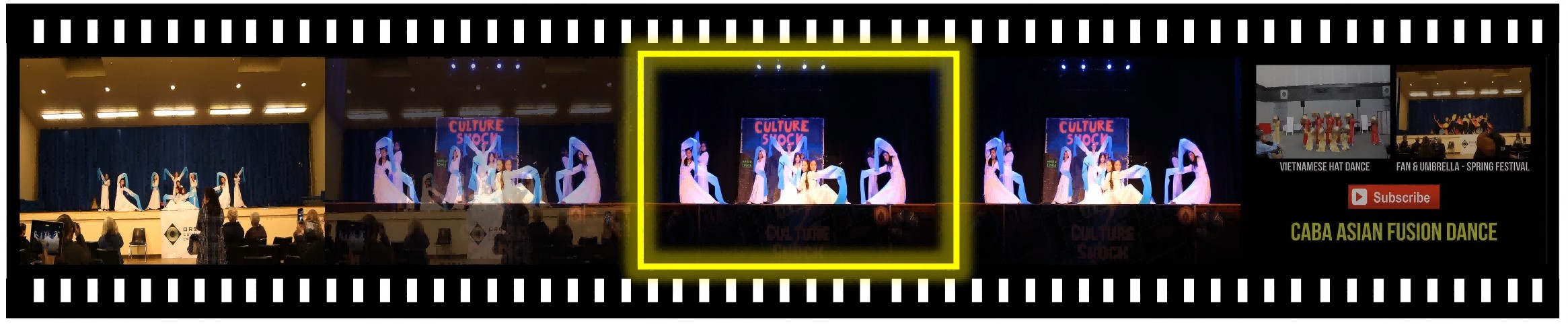}
\includegraphics[width=\columnwidth]{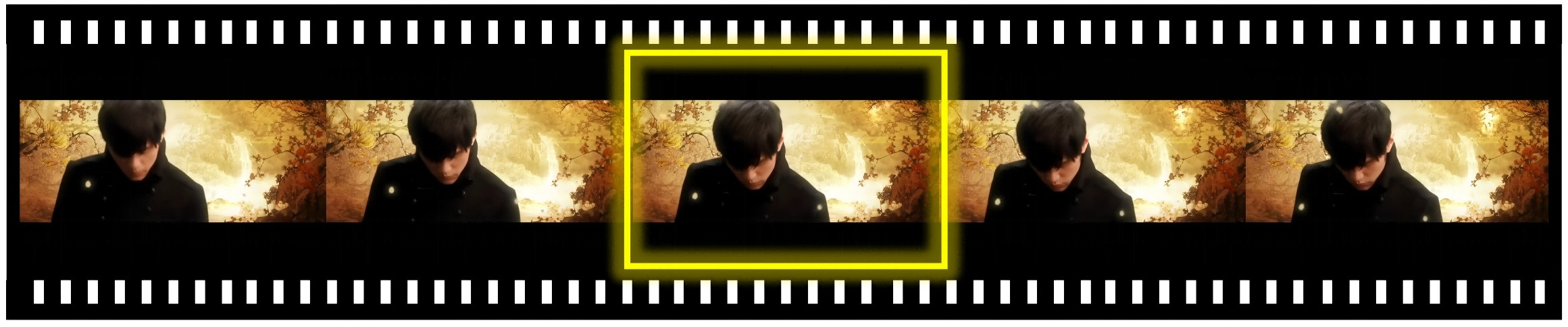}

\textbf{Key moment ground truth text:}\vspace{1pt}\\
Within this broader context, one specific segment I recall is the following. From what I recall, this segment features a clear, shallow stream or pond, with delicate concentric ripples spreading across the surface on the right side, as if a drop had just fallen. Through the transparent water, the riverbed is visible, covered in various sizes of gray-brown pebbles. Exposed soil and rocks appear in the upper right, while a few green blades of grass dip into the frame from the lower left. The water reflects the intricate silhouettes of tree branches above, creating a layered effect between reflection and reality. The scene is bathed in a soft teal hue, with gentle, shimmering light, captured from a high-angle perspective that emphasizes the texture of the water and the stillness of the environment.

\textbf{Ground truth frame:}\vspace{1pt}\\
\begin{center}
\includegraphics[height=2cm, keepaspectratio]{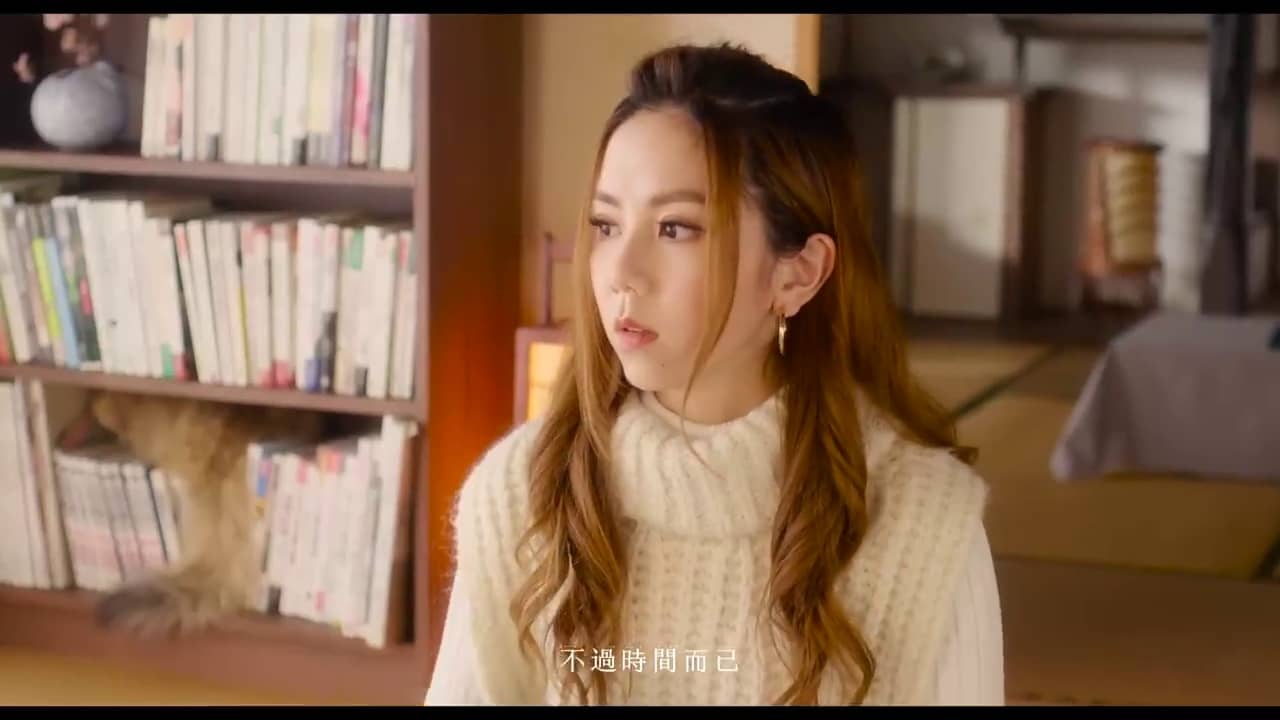}
\vspace{5pt}
\end{center}
\textbf{Model's prediction:}  \textcolor{red}{\textbf{False}}\\

\end{tcolorbox}

%% file: case_study/case4_localization2.tex
\tcbset{
    breakable,
    colframe=blue!5!black,
    colback=gray!10!white,
    fonttitle=\bfseries,
    width=\columnwidth 
}

\begin{tcolorbox}[
    title=\textbf{Localization with Judge Model Error},
    fonttitle=\bfseries
]


\textbf{Input fragment:} \vspace{2pt} \\
Please find the links to this video: I vaguely recall watching a video about ancient Chinese culture, focusing on the Classic of Poetry. It seemed to suggest that this text was more than just a collection of poems—it was like a mirror reflecting ancient life. I remember the speaker being in a quiet study, surrounded by books, in a very serene setting that made it feel as if you'd stepped back in time. His tone was calm, with a touch of humor, not at all like a lecture, but more like chatting with an old friend, making the seemingly difficult classical language come alive. I think he mentioned how the poems contained insights into human relationships, weddings and funerals, and natural landscapes, even drawing connections to life today. The whole video didn't feel stiff or academic; instead, it came across as warm and engaging. I faintly remember him using historical anecdotes to explain the verses, breathing life into words that once felt distant, as if those emotions and experiences were never really gone. Within this broader context, one specific segment I recall is the following. From what I recall, this segment features a middle-aged man with metal-rimmed glasses at the center, wearing a bright orange-red Chinese-style mandarin collar shirt, seated behind a wooden desk facing the camera. He is raising his hands slightly in expressive gesturing while speaking, with an animated and engaging expression. On the desk are several blue-and-white bound books, a Yixing teapot, a small teacup, and a rectangular wooden block resembling a storytelling wakewood. The background is a full wall of dark wooden bookshelves filled with books in various colors, creating a rich scholarly ambiance. The lighting is soft and even, with a medium shot clearly focused on the speaker, the background slightly blurred to emphasize the presenter.\\

\textbf{MLLM output context:} \vspace{1pt} \\
\textless think\textgreater
The video focuses on ancient Chinese culture, using the Classic of Poetry (Shijing) as a lens to reflect everyday life themes such as weddings, funerals, and nature. It is hosted by a middle-aged man wearing metal-rimmed glasses and a bright orange-red Chinese-style mandarin collar shirt, seated in a quiet study lined with dark wooden bookshelves. On the desk are blue-and-white bound books, a Yixing teapot and teacup, and a rectangular wooden block (wakewood or gavel), reinforcing a traditional scholarly and storytelling atmosphere. The host’s delivery style is calm, humorous, and conversational, resembling a pingshu-style narration, as if chatting with an old friend while recounting classical stories. These distinctive visual and stylistic elements suggest a creator who deliberately adopts the persona of a traditional storyteller, rather than a conventional classroom lecturer. While figures such as Liang Hongda or Dai Jianye share overlapping traits, the combination of the wakewood, mandarin-collar attire, and intimate study setting points more strongly to a dedicated cultural lecturer active on platforms like Bilibili or YouTube. Identifying the exact source would benefit from search queries that combine the topic and these strong visual anchors, such as “Classic of Poetry lecture orange mandarin collar wakewood” or “Shijing storyteller video orange shirt.”
\textless/think\textgreater

\textless tool\_call\textgreater
\{
\textquotedbl name\textquotedbl: \textquotedbl search\_videos\textquotedbl,
\textquotedbl arguments\textquotedbl:
\{
\textquotedbl query\textquotedbl:
[
\textquotedbl {Classic of Poetry lecture orange shirt man wakewood}\textquotedbl,
\textquotedbl {Shijing video storyteller orange mandarin collar}\textquotedbl
]
\}
\}
\textless/tool\_call\textgreater \\

\textbf{Selected URLs: }\vspace{1pt}
\begin{itemize}[itemsep=0cm]
    \item {https://www.youtube.com/watch?v=kOwwJR-jGbg}
    \item {https://www.youtube.com/watch?v=NJV3MV8-d\_Q}
    \item {https://www.youtube.com/watch?v=Aff4c4\_mlu0}
    \item {https://www.youtube.com/watch?v=ZXnZHkXIZEA}
\end{itemize}

\textbf{Target URL: }\vspace{1pt}
\begin{itemize}[itemsep=0cm]
    \item {https://www.youtube.com/watch?v=3IoaWleVI80}
\end{itemize}
~

\textbf{Extracted frames for video verification:}\vspace{1pt}\\
\includegraphics[width=\columnwidth]{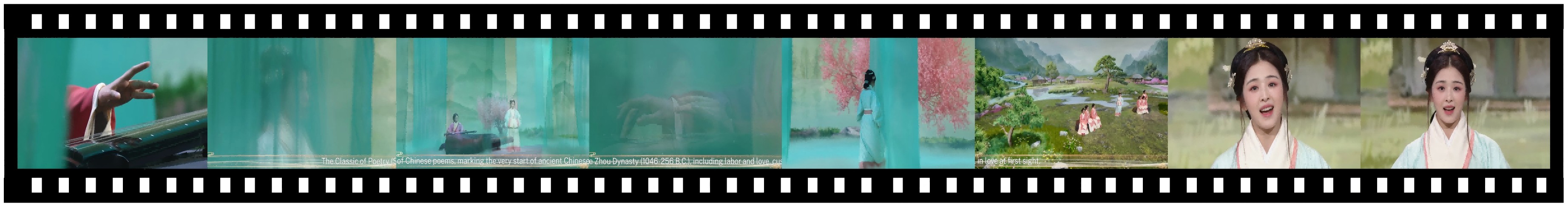}
\includegraphics[width=\columnwidth]{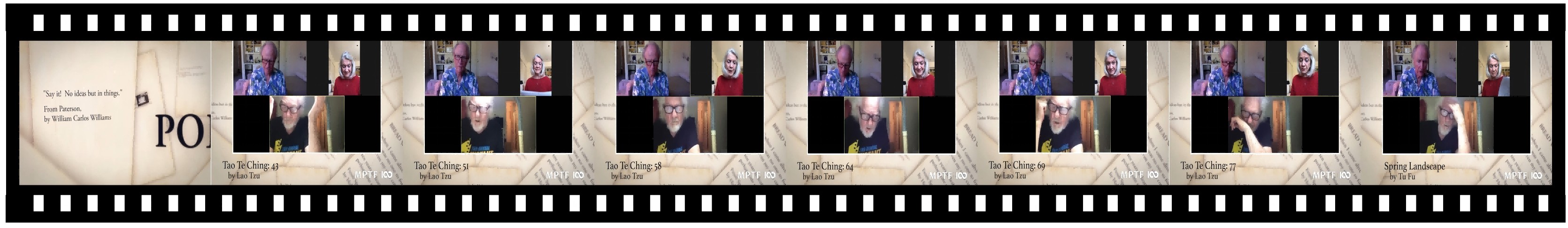}
\includegraphics[width=\columnwidth]{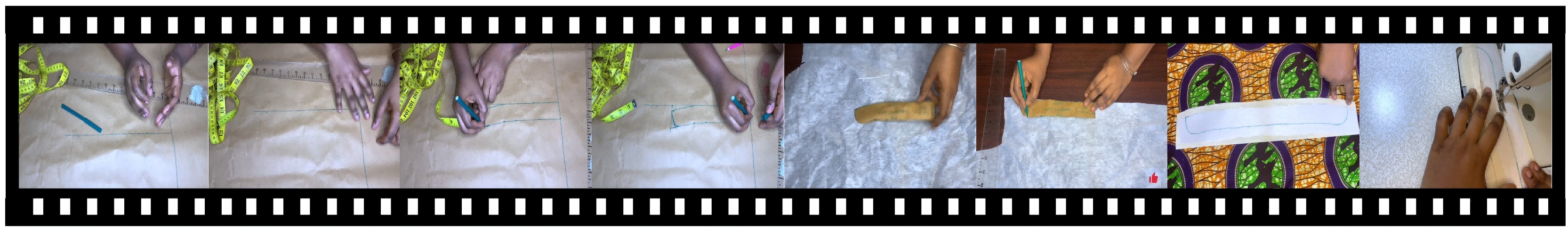}
\includegraphics[width=\columnwidth]{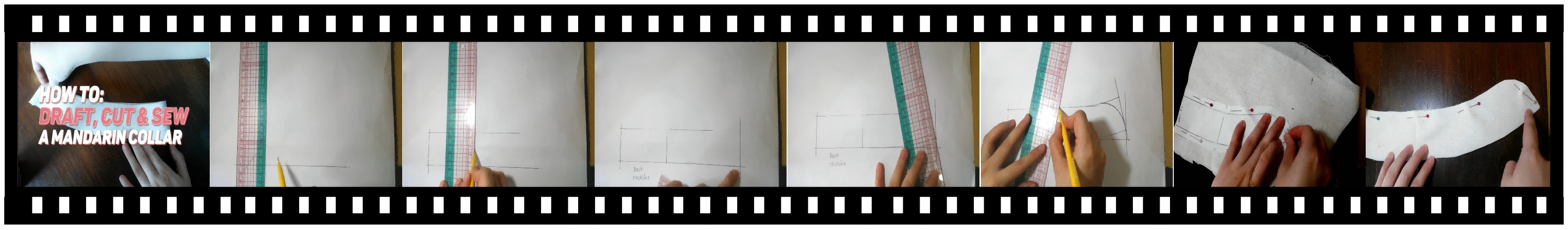}

\textbf{Global Impression ground truth text:}\vspace{1pt}\\
I vaguely recall watching a video about ancient Chinese culture, focusing on the Classic of Poetry. It seemed to suggest that this text was more than just a collection of poems—it was like a mirror reflecting ancient life. I remember the speaker being in a quiet study, surrounded by books, in a very serene setting that made it feel as if you'd stepped back in time. His tone was calm, with a touch of humor, not at all like a lecture, but more like chatting with an old friend, making the seemingly difficult classical language come alive. I think he mentioned how the poems contained insights into human relationships, weddings and funerals, and natural landscapes, even drawing connections to life today. The whole video didn't feel stiff or academic; instead, it came across as warm and engaging. I faintly remember him using historical anecdotes to explain the verses, breathing life into words that once felt distant, as if those emotions and experiences were never really gone.\\
\textbf{Video verification:}  \textcolor{red}{\textbf{False}}\\

\textbf{Selected frames for localization:} \vspace{1pt}\\
\includegraphics[width=\columnwidth]{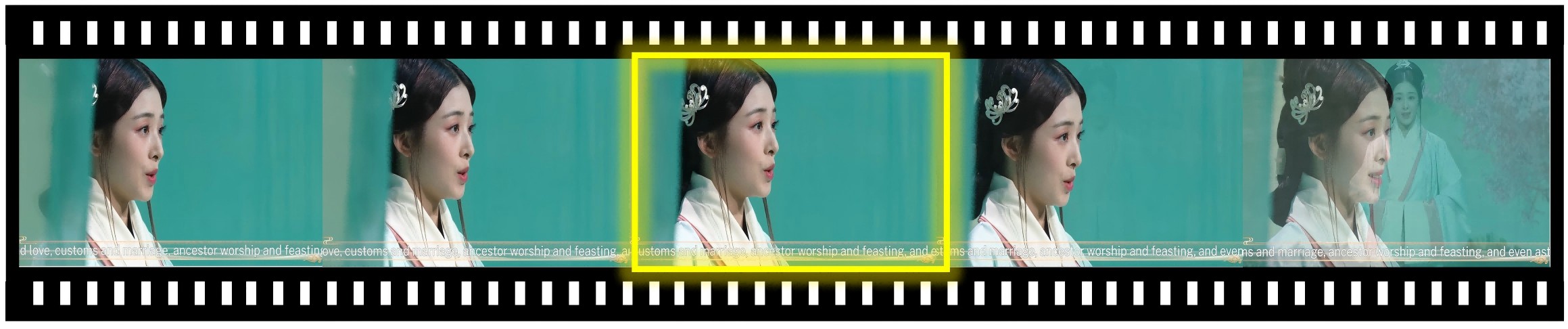}
\includegraphics[width=\columnwidth]{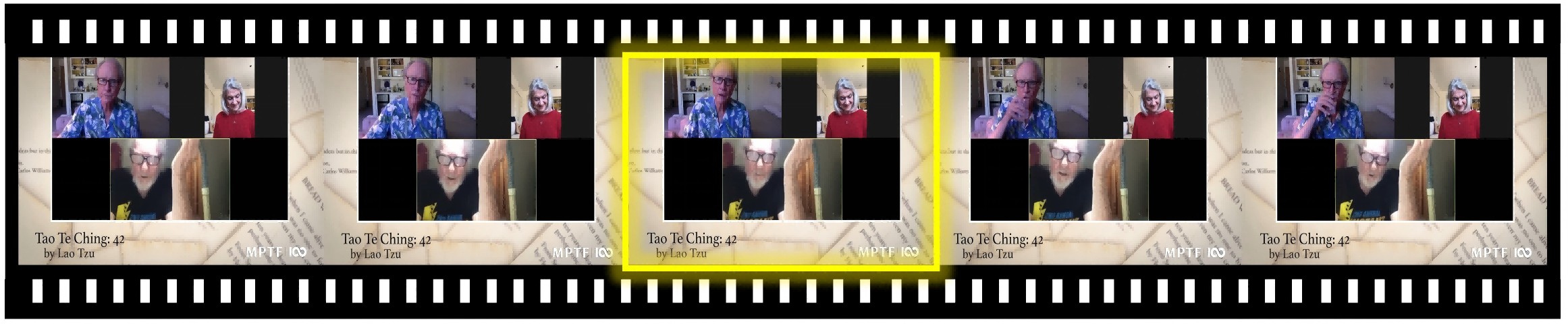}
\includegraphics[width=\columnwidth]{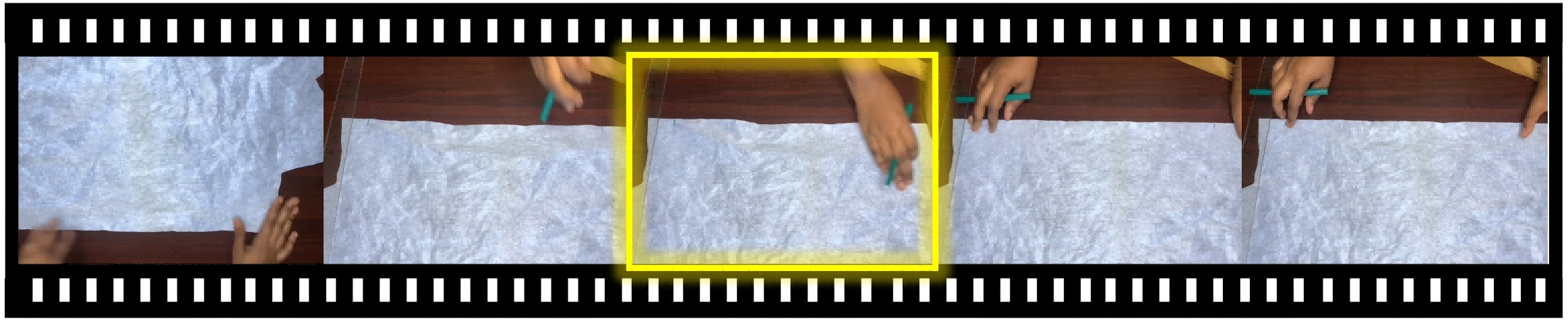}
\includegraphics[width=\columnwidth]{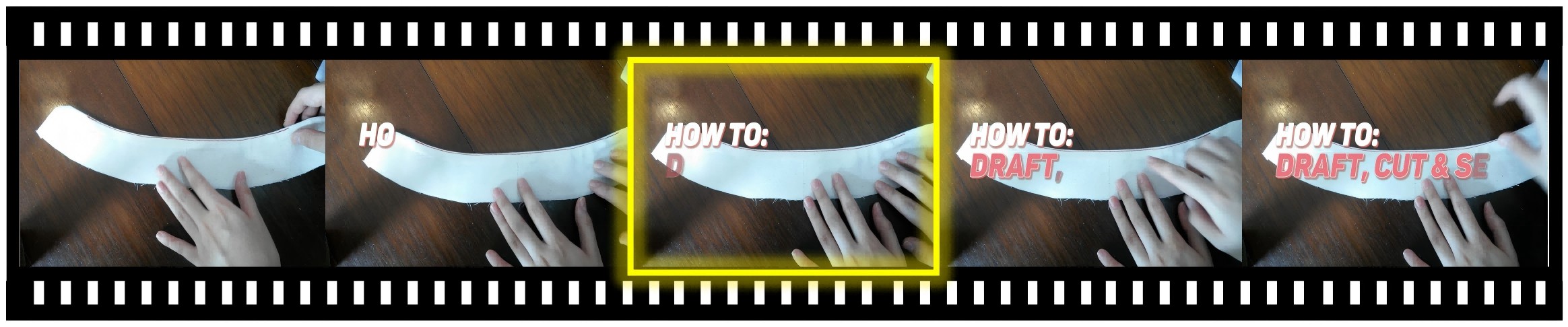}

\textbf{Key moment ground truth text:}\vspace{1pt}\\
Within this broader context, one specific segment I recall is the following. From what I recall, this segment features a middle-aged man with metal-rimmed glasses at the center, wearing a bright orange-red Chinese-style mandarin collar shirt, seated behind a wooden desk facing the camera. He is raising his hands slightly in expressive gesturing while speaking, with an animated and engaging expression. On the desk are several blue-and-white bound books, a Yixing teapot, a small teacup, and a rectangular wooden block resembling a storytelling wakewood. The background is a full wall of dark wooden bookshelves filled with books in various colors, creating a rich scholarly ambiance. The lighting is soft and even, with a medium shot clearly focused on the speaker, the background slightly blurred to emphasize the presenter.

\textbf{Ground truth frame:}\vspace{1pt}\\
\begin{center}
\includegraphics[height=2cm, keepaspectratio]{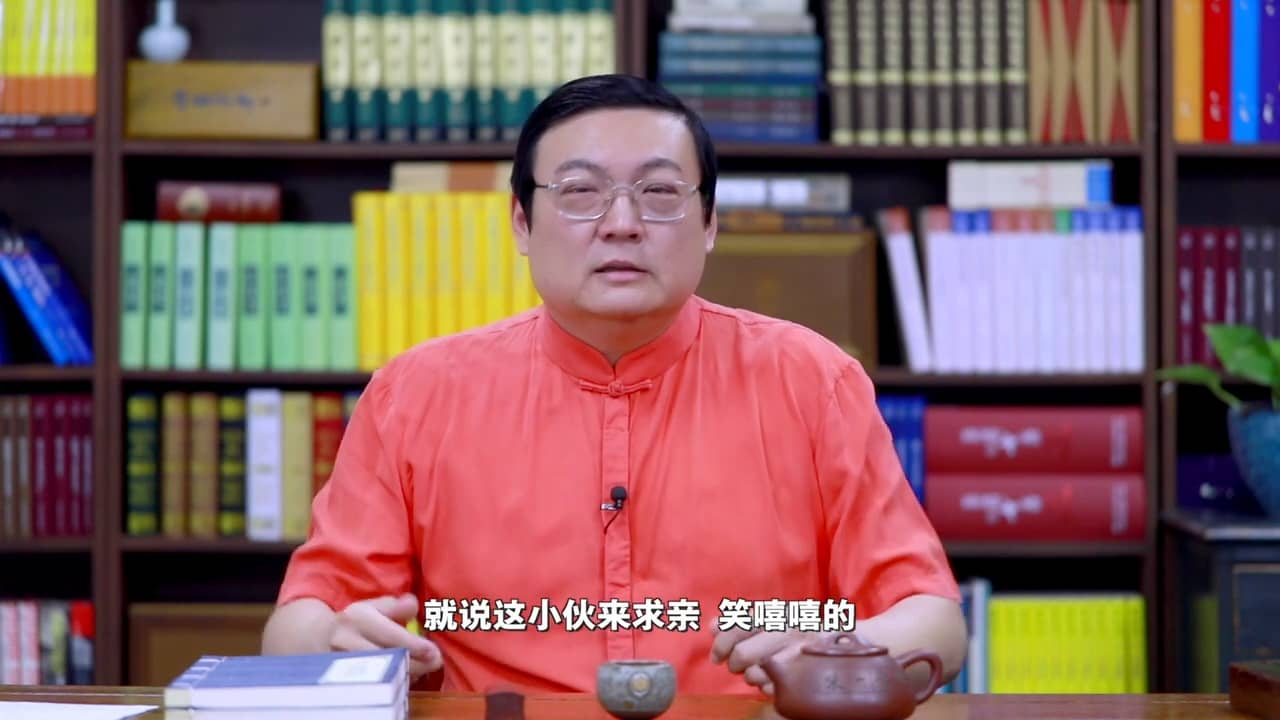}
\vspace{5pt}
\end{center}
\textbf{Model's prediction:}  \textcolor{PineGreen}{\textbf{True}}\\
\textbf{Human verification:}  \textcolor{red}{\textbf{False}}\\

\end{tcolorbox}

%% file: citation.bib
@misc{duan2024vlmevalkit,
      title={VLMEvalKit: An Open-Source Toolkit for Evaluating Large Multi-Modality Models}, 
      author={Haodong Duan and Xinyu Fang and Junming Yang and Xiangyu Zhao and Yuxuan Qiao and Mo Li and Amit Agarwal and Zhe Chen and Lin Chen and Yuan Liu and Yubo Ma and Hailong Sun and Yifan Zhang and Shiyin Lu and Tack Hwa Wong and Weiyun Wang and Peiheng Zhou and Xiaozhe Li and Chaoyou Fu and Junbo Cui and Jixuan Chen and Enxin Song and Song Mao and Shengyuan Ding and Tianhao Liang and Zicheng Zhang and Xiaoyi Dong and Yuhang Zang and Pan Zhang and Jiaqi Wang and Dahua Lin and Kai Chen},
      year={2025},
      eprint={2407.11691},
      archivePrefix={arXiv},
      primaryClass={cs.CV},
      url={https://arxiv.org/abs/2407.11691}, 
}

@misc{fu2025videommefirstevercomprehensiveevaluation,
      title={Video-MME: The First-Ever Comprehensive Evaluation Benchmark of Multi-modal LLMs in Video Analysis}, 
      author={Chaoyou Fu and Yuhan Dai and Yongdong Luo and Lei Li and Shuhuai Ren and Renrui Zhang and Zihan Wang and Chenyu Zhou and Yunhang Shen and Mengdan Zhang and Peixian Chen and Yanwei Li and Shaohui Lin and Sirui Zhao and Ke Li and Tong Xu and Xiawu Zheng and Enhong Chen and Caifeng Shan and Ran He and Xing Sun},
      year={2025},
      eprint={2405.21075},
      archivePrefix={arXiv},
      primaryClass={cs.CV},
      url={https://arxiv.org/abs/2405.21075}, 
}

@InProceedings{Xu_2016_CVPR,
author = {Xu, Jun and Mei, Tao and Yao, Ting and Rui, Yong},
title = {MSR-VTT: A Large Video Description Dataset for Bridging Video and Language},
booktitle = {Proceedings of the IEEE Conference on Computer Vision and Pattern Recognition (CVPR)},
month = {June},
year = {2016}
}

@misc{wang2020vatexlargescalehighqualitymultilingual,
      title={VATEX: A Large-Scale, High-Quality Multilingual Dataset for Video-and-Language Research}, 
      author={Xin Wang and Jiawei Wu and Junkun Chen and Lei Li and Yuan-Fang Wang and William Yang Wang},
      year={2020},
      eprint={1904.03493},
      archivePrefix={arXiv},
      primaryClass={cs.CV},
      url={https://arxiv.org/abs/1904.03493}, 
}

@misc{cai2025lovrbenchmarklongvideo,
      title={LoVR: A Benchmark for Long Video Retrieval in Multimodal Contexts}, 
      author={Qifeng Cai and Hao Liang and Hejun Dong and Meiyi Qiang and Ruichuan An and Zhaoyang Han and Zhengzhou Zhu and Bin Cui and Wentao Zhang},
      year={2025},
      eprint={2505.13928},
      archivePrefix={arXiv},
      primaryClass={cs.CV},
      url={https://arxiv.org/abs/2505.13928}, 
}

@misc{feng2025muvrmultimodaluntrimmedvideo,
      title={MUVR: A Multi-Modal Untrimmed Video Retrieval Benchmark with Multi-Level Visual Correspondence}, 
      author={Yue Feng and Jinwei Hu and Qijia Lu and Jiawei Niu and Li Tan and Shuo Yuan and Ziyi Yan and Yizhen Jia and Qingzhi He and Shiping Ge and Ethan Q. Chen and Wentong Li and Limin Wang and Jie Qin},
      year={2025},
      eprint={2510.21406},
      archivePrefix={arXiv},
      primaryClass={cs.CV},
      url={https://arxiv.org/abs/2510.21406}, 
}

@misc{park2026vagentinteractivevideosearch,
      title={V-Agent: An Interactive Video Search System Using Vision-Language Models}, 
      author={SunYoung Park and Jong-Hyeon Lee and Youngjune Kim and Daegyu Sung and Younghyun Yu and Young-rok Cha and Jeongho Ju},
      year={2026},
      eprint={2512.16925},
      archivePrefix={arXiv},
      primaryClass={cs.CV},
      url={https://arxiv.org/abs/2512.16925}, 
}

@misc{yuan2025videoexplorerthinkvideosagentic,
      title={VideoExplorer: Think With Videos For Agentic Long-Video Understanding}, 
      author={Huaying Yuan and Zheng Liu and Junjie Zhou and Hongjin Qian and Yan Shu and Nicu Sebe and Ji-Rong Wen and Zhicheng Dou},
      year={2025},
      eprint={2506.10821},
      archivePrefix={arXiv},
      primaryClass={cs.CV},
      url={https://arxiv.org/abs/2506.10821}, 
}

@misc{radford2021learningtransferablevisualmodels,
      title={Learning Transferable Visual Models From Natural Language Supervision}, 
      author={Alec Radford and Jong Wook Kim and Chris Hallacy and Aditya Ramesh and Gabriel Goh and Sandhini Agarwal and Girish Sastry and Amanda Askell and Pamela Mishkin and Jack Clark and Gretchen Krueger and Ilya Sutskever},
      year={2021},
      eprint={2103.00020},
      archivePrefix={arXiv},
      primaryClass={cs.CV},
      url={https://arxiv.org/abs/2103.00020}, 
}

@misc{liu2023visualinstructiontuning,
      title={Visual Instruction Tuning}, 
      author={Haotian Liu and Chunyuan Li and Qingyang Wu and Yong Jae Lee},
      year={2023},
      eprint={2304.08485},
      archivePrefix={arXiv},
      primaryClass={cs.CV},
      url={https://arxiv.org/abs/2304.08485}, 
}

@misc{bai2025qwen3vltechnicalreport,
      title={Qwen3-VL Technical Report}, 
      author={Shuai Bai and Yuxuan Cai and Ruizhe Chen and Keqin Chen and Xionghui Chen and Zesen Cheng and Lianghao Deng and Wei Ding and Chang Gao and Chunjiang Ge and Wenbin Ge and Zhifang Guo and Qidong Huang and Jie Huang and Fei Huang and Binyuan Hui and Shutong Jiang and Zhaohai Li and Mingsheng Li and Mei Li and Kaixin Li and Zicheng Lin and Junyang Lin and Xuejing Liu and Jiawei Liu and Chenglong Liu and Yang Liu and Dayiheng Liu and Shixuan Liu and Dunjie Lu and Ruilin Luo and Chenxu Lv and Rui Men and Lingchen Meng and Xuancheng Ren and Xingzhang Ren and Sibo Song and Yuchong Sun and Jun Tang and Jianhong Tu and Jianqiang Wan and Peng Wang and Pengfei Wang and Qiuyue Wang and Yuxuan Wang and Tianbao Xie and Yiheng Xu and Haiyang Xu and Jin Xu and Zhibo Yang and Mingkun Yang and Jianxin Yang and An Yang and Bowen Yu and Fei Zhang and Hang Zhang and Xi Zhang and Bo Zheng and Humen Zhong and Jingren Zhou and Fan Zhou and Jing Zhou and Yuanzhi Zhu and Ke Zhu},
      year={2025},
      eprint={2511.21631},
      archivePrefix={arXiv},
      primaryClass={cs.CV},
      url={https://arxiv.org/abs/2511.21631}, 
}

@InProceedings{Miech_2020_CVPR,
author = {Miech, Antoine and Alayrac, Jean-Baptiste and Smaira, Lucas and Laptev, Ivan and Sivic, Josef and Zisserman, Andrew},
title = {End-to-End Learning of Visual Representations From Uncurated Instructional Videos},
booktitle = {Proceedings of the IEEE/CVF Conference on Computer Vision and Pattern Recognition (CVPR)},
month = {June},
year = {2020}
}

@misc{bain2022frozentimejointvideo,
      title={Frozen in Time: A Joint Video and Image Encoder for End-to-End Retrieval}, 
      author={Max Bain and Arsha Nagrani and Gül Varol and Andrew Zisserman},
      year={2022},
      eprint={2104.00650},
      archivePrefix={arXiv},
      primaryClass={cs.CV},
      url={https://arxiv.org/abs/2104.00650}, 
}

@misc{yuan2026momentseekertaskorientedbenchmarklongvideo,
      title={MomentSeeker: A Task-Oriented Benchmark For Long-Video Moment Retrieval}, 
      author={Huaying Yuan and Jian Ni and Zheng Liu and Yueze Wang and Junjie Zhou and Zhengyang Liang and Bo Zhao and Zhao Cao and Zhicheng Dou and Ji-Rong Wen},
      year={2026},
      eprint={2502.12558},
      archivePrefix={arXiv},
      primaryClass={cs.CV},
      url={https://arxiv.org/abs/2502.12558}, 
}

@misc{fang2024mmbenchvideolongformmultishotbenchmark,
      title={MMBench-Video: A Long-Form Multi-Shot Benchmark for Holistic Video Understanding}, 
      author={Xinyu Fang and Kangrui Mao and Haodong Duan and Xiangyu Zhao and Yining Li and Dahua Lin and Kai Chen},
      year={2024},
      eprint={2406.14515},
      archivePrefix={arXiv},
      primaryClass={cs.CV},
      url={https://arxiv.org/abs/2406.14515}, 
}

@InProceedings{Jiang_2019_ICCV,
author = {Jiang, Qing-Yuan and He, Yi and Li, Gen and Lin, Jian and Li, Lei and Li, Wu-Jun},
title = {SVD: A Large-Scale Short Video Dataset for Near-Duplicate Video Retrieval},
booktitle = {Proceedings of the IEEE/CVF International Conference on Computer Vision (ICCV)},
month = {October},
year = {2019}
}

@misc{yang2025audiocentricvideounderstandingbenchmark,
      title={Audio-centric Video Understanding Benchmark without Text Shortcut}, 
      author={Yudong Yang and Jimin Zhuang and Guangzhi Sun and Changli Tang and Yixuan Li and Peihan Li and Yifan Jiang and Wei Li and Zejun Ma and Chao Zhang},
      year={2025},
      eprint={2503.19951},
      archivePrefix={arXiv},
      primaryClass={cs.CV},
      url={https://arxiv.org/abs/2503.19951}, 
}

@misc{zhu2025ravstempoweringvideollmsfinegrained,
      title={R-AVST: Empowering Video-LLMs with Fine-Grained Spatio-Temporal Reasoning in Complex Audio-Visual Scenarios}, 
      author={Lu Zhu and Tiantian Geng and Yangye Chen and Teng Wang and Ping Lu and Feng Zheng},
      year={2025},
      eprint={2511.16901},
      archivePrefix={arXiv},
      primaryClass={cs.CV},
      url={https://arxiv.org/abs/2511.16901}, 
}

@inproceedings{10.1007/978-3-031-19781-9_29,
author = {Lee, Sangmin and Park, Sungjune and Ro, Yong Man},
title = {Audio-Visual Mismatch-Aware Video Retrieval via Association and Adjustment},
year = {2022},
isbn = {978-3-031-19780-2},
publisher = {Springer-Verlag},
address = {Berlin, Heidelberg},
url = {https://doi.org/10.1007/978-3-031-19781-9_29},
doi = {10.1007/978-3-031-19781-9_29},
booktitle = {Computer Vision – ECCV 2022: 17th European Conference, Tel Aviv, Israel, October 23–27, 2022, Proceedings, Part XIV},
pages = {497–514},
numpages = {18},
location = {Tel Aviv, Israel}
}

@misc{liu2023visualspatialreasoning,
      title={Visual Spatial Reasoning}, 
      author={Fangyu Liu and Guy Emerson and Nigel Collier},
      year={2023},
      eprint={2205.00363},
      archivePrefix={arXiv},
      primaryClass={cs.CL},
      url={https://arxiv.org/abs/2205.00363}, 
}

@misc{yang2022tubedetrspatiotemporalvideogrounding,
      title={TubeDETR: Spatio-Temporal Video Grounding with Transformers}, 
      author={Antoine Yang and Antoine Miech and Josef Sivic and Ivan Laptev and Cordelia Schmid},
      year={2022},
      eprint={2203.16434},
      archivePrefix={arXiv},
      primaryClass={cs.CV},
      url={https://arxiv.org/abs/2203.16434}, 
}

@misc{surís2023vipergptvisualinferencepython,
      title={ViperGPT: Visual Inference via Python Execution for Reasoning}, 
      author={Dídac Surís and Sachit Menon and Carl Vondrick},
      year={2023},
      eprint={2303.08128},
      archivePrefix={arXiv},
      primaryClass={cs.CV},
      url={https://arxiv.org/abs/2303.08128}, 
}

@misc{monfort2019momentstimedatasetmillion,
      title={Moments in Time Dataset: one million videos for event understanding}, 
      author={Mathew Monfort and Alex Andonian and Bolei Zhou and Kandan Ramakrishnan and Sarah Adel Bargal and Tom Yan and Lisa Brown and Quanfu Fan and Dan Gutfruend and Carl Vondrick and Aude Oliva},
      year={2019},
      eprint={1801.03150},
      archivePrefix={arXiv},
      primaryClass={cs.CV},
      url={https://arxiv.org/abs/1801.03150}, 
}

@inproceedings{shao2020finegym,
title={FineGym: A Hierarchical Video Dataset for Fine-grained Action Understanding},
author={Shao, Dian and Zhao, Yue and Dai, Bo and Lin, Dahua},
booktitle={IEEE Conference on Computer Vision and Pattern Recognition (CVPR)},
year={2020}
}

@ARTICLE{11206011,
  author={Xie, Peiyu and Li, Jinxing and Lu, Guangming and Xu, Yong and Zhang, David},
  journal={IEEE Transactions on Image Processing}, 
  title={Caption Assisted Multimodal Large Language Model for Video Moment Retrieval}, 
  year={2025},
  volume={34},
  number={},
  pages={6755-6766},
  doi={10.1109/TIP.2025.3620124}}

@misc{zhang2025vcagentinteractiveagentcustomized,
      title={VC-Agent: An Interactive Agent for Customized Video Dataset Collection}, 
      author={Yidan Zhang and Mutian Xu and Yiming Hao and Kun Zhou and Jiahao Chang and Xiaoqiang Liu and Pengfei Wan and Hongbo Fu and Xiaoguang Han},
      year={2025},
      eprint={2509.21291},
      archivePrefix={arXiv},
      primaryClass={cs.AI},
      url={https://arxiv.org/abs/2509.21291}, 
}

@misc{yang2025vimix14mcuratedmultisourcevideotext,
      title={ViMix-14M: A Curated Multi-Source Video-Text Dataset with Long-Form, High-Quality Captions and Crawl-Free Access}, 
      author={Timing Yang and Sucheng Ren and Alan Yuille and Feng Wang},
      year={2025},
      eprint={2511.18382},
      archivePrefix={arXiv},
      primaryClass={cs.CV},
      url={https://arxiv.org/abs/2511.18382}, 
}

@misc{ju2024miradatalargescalevideodataset,
      title={MiraData: A Large-Scale Video Dataset with Long Durations and Structured Captions}, 
      author={Xuan Ju and Yiming Gao and Zhaoyang Zhang and Ziyang Yuan and Xintao Wang and Ailing Zeng and Yu Xiong and Qiang Xu and Ying Shan},
      year={2024},
      eprint={2407.06358},
      archivePrefix={arXiv},
      primaryClass={cs.CV},
      url={https://arxiv.org/abs/2407.06358}, 
}

@inproceedings{10.5555/2856823.2856863,
author = {M\"{u}ller, Adrian and Thiel, Ulrich},
title = {Query expansion in an abductive information retrieval system},
year = {1994},
publisher = {LE CENTRE DE HAUTES ETUDES INTERNATIONALES D'INFORMATIQUE DOCUMENTAIRE},
address = {Paris, FRA},
booktitle = {Intelligent Multimedia Information Retrieval Systems and Management - Volume 1},
pages = {461–480},
numpages = {20},
location = {New York, New York},
series = {RIAO '94}
}

@misc{fariha2019exampledrivenqueryintentdiscovery,
      title={Example-Driven Query Intent Discovery: Abductive Reasoning using Semantic Similarity}, 
      author={Anna Fariha and Alexandra Meliou},
      year={2019},
      eprint={1906.10322},
      archivePrefix={arXiv},
      primaryClass={cs.DB},
      url={https://arxiv.org/abs/1906.10322}, 
}

@INPROCEEDINGS{687481,
  author={Miyata, Y. and Furnhashi, T. and Uchikawa, Y.},
  booktitle={1998 IEEE International Conference on Fuzzy Systems Proceedings. IEEE World Congress on Computational Intelligence (Cat. No.98CH36228)}, 
  title={Query expansion using fuzzy abductive inference for creative thinking support system}, 
  year={1998},
  volume={1},
  number={},
  pages={189-193 vol.1},
  doi={10.1109/FUZZY.1998.687481}}

@misc{yao2023reactsynergizingreasoningacting,
      title={ReAct: Synergizing Reasoning and Acting in Language Models}, 
      author={Shunyu Yao and Jeffrey Zhao and Dian Yu and Nan Du and Izhak Shafran and Karthik Narasimhan and Yuan Cao},
      year={2023},
      eprint={2210.03629},
      archivePrefix={arXiv},
      primaryClass={cs.CL},
      url={https://arxiv.org/abs/2210.03629}, 
}

@misc{bai2025qwen25vltechnicalreport,
      title={Qwen2.5-VL Technical Report}, 
      author={Shuai Bai and Keqin Chen and Xuejing Liu and Jialin Wang and Wenbin Ge and Sibo Song and Kai Dang and Peng Wang and Shijie Wang and Jun Tang and Humen Zhong and Yuanzhi Zhu and Mingkun Yang and Zhaohai Li and Jianqiang Wan and Pengfei Wang and Wei Ding and Zheren Fu and Yiheng Xu and Jiabo Ye and Xi Zhang and Tianbao Xie and Zesen Cheng and Hang Zhang and Zhibo Yang and Haiyang Xu and Junyang Lin},
      year={2025},
      eprint={2502.13923},
      archivePrefix={arXiv},
      primaryClass={cs.CV},
      url={https://arxiv.org/abs/2502.13923}, 
}
